\documentclass[accepted]{uai2026}

\usepackage[american]{babel}
\usepackage{natbib}
\usepackage{mathtools}
\usepackage{booktabs}
\usepackage{tikz}
\usepackage{graphicx}
\usepackage{subcaption}
\usepackage{amsmath}
\usepackage{amssymb}
\usepackage{amsthm}
\usepackage{bm}
\usepackage{bbm}
\usepackage{dsfont}
\usepackage{multirow}
\usepackage[capitalize,noabbrev]{cleveref}
\usepackage{algorithm}
\usepackage{algorithmic}
\theoremstyle{plain}
\newtheorem{theorem}{Theorem}[section]
\newtheorem{proposition}[theorem]{Proposition}
\newtheorem{lemma}[theorem]{Lemma}
\newtheorem{corollary}[theorem]{Corollary}
\theoremstyle{definition}
\newtheorem{definition}[theorem]{Definition}

\theoremstyle{remark}
\newtheorem{remark}[theorem]{Remark}

\makeatletter
\renewenvironment{proof}[1][Proof]{
  \par\pushQED{\qed}
  \normalfont \topsep6\p@\@plus6\p@\relax
  \trivlist
  \item[\hskip\labelsep\bfseries\upshape #1\@addpunct{.}]%
}{%
  \popQED\endtrivlist\@endpefalse
}
\makeatother
\newcommand\shortsection[1]{\par\noindent\textbf{#1.}}
\newcommand\shortersection[1]{\par\noindent\emph{#1.}}

\usepackage{amsmath,amsfonts,bm}

\def\eqref#1{equation~\ref{#1}}
\def\Eqref#1{Equation~\ref{#1}}

\def\1{\bm{1}}

\DeclareMathAlphabet{\mathsfit}{\encodingdefault}{\sfdefault}{m}{sl}
\SetMathAlphabet{\mathsfit}{bold}{\encodingdefault}{\sfdefault}{bx}{n}

\newcommand{\Cov}{\mathrm{Cov}}

\DeclareMathOperator*{\argmax}{arg\,max}

\title{How Learning Dynamics Drive Adversarially Robust Generalization?}

\author[1]{\href{mailto:yuelin.xu@cispa.de}{Yuelin Xu}}
\author[1]{\href{mailto:xiao.zhang@cispa.de}{Xiao Zhang}}
\affil[1]{
    CISPA Helmholtz Center for Information Security\\
    Saarbrücken, Germany
}

\begin{document}

\maketitle 

\begin{abstract}
Despite being widely adopted as a canonical framework for learning robust models, adversarial training suffers from robust overfitting. Existing empirical and theoretical explorations fail to provide a satisfactory mechanistic interpretation of the phenomenon. By modeling adversarial training with momentum SGD as a discrete-time dynamical system, we propose a PAC-Bayesian analytical framework that proves time-resolved robust generalization bounds. Specifically, our framework tracks the closed-form evolution of the posterior mean and covariance under both stationary and non-stationary transient regimes, connecting the model's robust generalization performance to learning rate, local loss geometry, and mini-batch stochastic gradients. By estimating the key quantities associated with the bound, we illustrate the underlying mechanism of robust overfitting. Our framework also shows how adversarial weight perturbation reduces robust generalization gaps by suppressing dominant loss-curvature modes, while suggesting that excessive penalization can be sub-optimal for optimization.
\end{abstract}

\section{Introduction}

Adversarial training~\citep{madry2017towards}, which uses projected gradient descent (PGD) to solve a minimax optimization objective, remains the most widely adopted algorithmic framework for training neural networks to be resilient to inputs crafted with small perturbations intended to fool the model~\citep{szegedy2013intriguing}.
Despite its prevalence, adversarial training exhibits a striking failure mode: robust test accuracy can deteriorate late in training---often immediately after a learning rate decay---even as the robust training loss continues to decrease. 
This phenomenon, known as \textit{robust overfitting}~\citep{rice2020overfitting}, suggests that fitting the training objective does not necessarily translate into improved robust generalization on unseen data. 
To address the issue, various techniques have been proposed~\citep{zhang2019theoretically,wu2020adversarial,carmon2019unlabeled}, yet a satisfying mechanistic interpretation of the phenomenon is lacking.
While many heuristic robustness measures have been proposed, no single one can serve as a universal, reliable indicator of the model's robust generalization capability~\citep{kim2023fantastic}.

The surprising robust overfitting phenomenon and its complexity have motivated theoretical studies~\citep{xiao2022stability,wang2023improving,xiao2023pac,fu2023theoretical,liu2024impact,walter2025flatness,tian2025algorithmic} to analyze the factors that influence adversarially robust generalization by adapting classical frameworks such as PAC-Bayes and algorithmic stability. 
Nevertheless, the worst-case robustness guarantees derived from these studies are often loose and rely on strong assumptions that may not reflect the actual behavior of algorithms used in practice. 
Moreover, these bounds are usually \textit{static}, designed to certify a predictor at a fixed model checkpoint, but are insufficient to characterize the \textit{time-varying dynamics} of adversarial training, which is essential to fully understand robust overfitting.

In this work, we argue that robust overfitting is driven by a transient imbalance between loss curvature and stochastic noise as the model navigates the adversarial loss landscape through stochastic gradient descent (SGD). 
To uncover the underlying mechanism, we adopt the PAC-Bayesian framework and model the SGD dynamics of adversarial training algorithms as a discrete-time dynamical system (Section \ref{sec:link to loss and posterior geometry}). By treating the iterative parameter distribution as an implicit posterior, we derive closed-form solutions that characterize how its mean and covariance evolve with learning rate, local curvature, and gradient-noise dynamics, which yield schedule-aware, time-varying robust generalization bounds tailored for different training stages of interest (Section \ref{sec:posterior dynamics analysis}).
Based on an efficient spectral estimation protocol, we empirically estimate the quantities identified by the theory by tracking the per-epoch Hessian and gradient-noise dynamics across different learning algorithms, thereby characterizing the driving factors associated with adversarially robust generalization (Section \ref{sec:experiments}).
Together, our analyses support a \textit{time-varying mechanistic interpretation} of robust overfitting: a sharp learning-rate decay leads to posterior contraction along sharp directions, benefiting robust optimization while lowering the curvature-weighted variance during the initial transient phase (corresponding to the early jump of training and testing robust accuracy); however, the Hessian eigenvalues keep increasing as adversarial training prolongs, which eventually boosts the curvature-weighted variance despite the contracted posterior and adversely affects the model's robust generalization (corresponding to the gradual degradation of robust test accuracy in later epochs).

Our main contributions are summarized as follows:
\begin{enumerate}[itemsep=0.1cm,parsep=0.1cm]
    \item[\textbullet] By modeling momentum SGD as a dynamical system, we prove PAC-Bayesian robust generalization bounds for different learning stages of adversarial training. As a byproduct, our method can explicitly track the time-varying evolution of the posterior mean and covariance.
    \item[\textbullet] Through empirical estimation of the spectral quantities reflected in the bound, we provide a time-resolved diagnostic account of robust overfitting, illustrating how learning-rate decay, progressive sharpening, and posterior contraction jointly shape robust generalization.
    \item[\textbullet] We conduct controlled experiments to analyze the diagnostic patterns across various learning algorithms, suggesting that adversarial weight perturbation~\citep{wu2020adversarial} may overly penalize Hessian eigenvalues. 
\end{enumerate}

\section{Related Work}

\shortsection{Robust Overfitting}
Since the initial discovery of adversarial examples~\citep{szegedy2013intriguing}, numerous methods have been proposed to improve model robustness~\citep{goodfellow2014explaining,papernot2016distillation,wong2018provable,cohen2019certified}. Among them, \textit{adversarial training} (AT), a minimax optimization framework that employs \textit{projected gradient descent} (PGD), is most popular~\citep{madry2017towards}. However, AT exhibits a recurring failure mode: robust test accuracy can drop late in training---often right after learning rate decay---even as the robust training loss continues to decrease, a phenomenon termed \textit{robust overfitting}~\citep{rice2020overfitting}.
Empirical remedies span three main categories: objective-based methods like TRADES~\citep{zhang2019theoretically,chen2020robust} redesign the training optimization objective, geometry-aware approaches like \textit{adversarial weight perturbation} (AWP)~\citep{liu2020loss,wu2020adversarial,foret2020sharpness} penalize the model's sharpness for finding flatter solutions, and
augmentation-based schemes~\citep{carmon2019unlabeled,gowal2021improving} incorporate extra data in adversarial training.
Despite improvements, the above techniques are all empirical, which lacks a unified, mechanistic interpretation of the robust overfitting phenomenon.

\shortsection{Adversarially Robust Generalization}
Existing literature has analyzed robust generalization using various theoretical tools. A line of work adapted the classical PAC-Bayes framework~\citep{mcallester1999pac,neyshabur2017pac,bartlett2021deep} to derive posterior-based robust generalization bounds~\citep{Mustafa2023NonvacuousPB,mustafa2024non,xiao2023pac,wang2023improving}, often relating the KL-divergence complexity term to flatness-based measures. 
Another line of research studied adversarially robust generalization from the perspective of algorithmic stability~\citep{xing2021algorithmic,xiao2022stability,tian2025algorithmic}, which captures a learning algorithm's insensitivity to small perturbations in the training dataset.
Additionally, \citet{fu2023theoretical} extended neural tangent kernel (NTK) analysis to AT, deriving closed-form dynamics under squared loss for neural networks with infinite width, while \citet{liu2024impact} focused on characterizing the impact of individual training instances on robust overfitting. 
Despite rigorous proofs, most theories rely on assumptions that deviate from practice. Even when assumptions are acceptable, the resulting bounds are often loose and cannot tightly explain the empirical robust overfitting curves. 
These observations highlight a critical limitation: existing studies focus on \textit{static} worst-case guarantees; however, robust generalization requires understanding the model's internal dynamics, suggesting the need for better analytical tools to characterize the actual adversarial training dynamics.

\shortsection{SGD Dynamics}
Recently, a line of literature has proposed studying standard deep learning generalization by modeling SGD as a discrete-time stochastic dynamical system~\citep{liu2021noise,ziyin2021strength,ziyin2024parameter}. In particular, they showed how the learning rate, loss curvature, and gradient noise structure jointly shape parameter fluctuations, which can be used to derive an exact analytical form of the stationary distribution and explain how SGD escapes from sharp minima.
Compared with prior literature that uses continuous-time SDEs or Langevin approximations~\citep{mandt2017stochastic,li2017stochastic}, this line of work accounts for the learning step size and minibatch stochastic noise, providing more realistic predictions for SGD dynamics and implicit regularization.
We leverage recent advances in modeling SGD dynamics to characterize \textit{time-resolved} PAC-Bayesian robust generalization bounds for adversarial training, offering a principled framework for understanding robust overfitting.

\section{Preliminaries}
\label{sec:preliminaries}

\shortsection{Adversarial Risk}
We work with the following theoretical definition of adversarial risk, which often serves as the basis for evaluating an ML model's robustness against worst-case input perturbations~\citep{madry2017towards,rice2020overfitting}.

\begin{definition}[Adversarial risk]
\label{def:adversarial risk}
Let $\mathcal{X}\subseteq\mathbb{R}^d$ be the input space, $\mathcal{Y}$ be the set of class labels, and $\mathcal{D}$ be a probability distribution over $\mathcal{X}\times\mathcal{Y}$. For any model
$f_{\bm{w}}:\mathcal{X}\rightarrow\mathcal{Y}$ with parameters $\bm{w}\in\mathbb{R}^m$,
the \textit{adversarial risk} of $f_{\bm{w}}$ over $\mathcal{D}$ against $L_p$ perturbations bounded by $\epsilon\geq 0$ is defined as:
\begin{align}
\label{eq:def adversarial risk}
    \mathcal{R}_{\epsilon}(\bm{w}) = \mathrm{Pr}_{(\bm{x}, y) \sim \mathcal{D}} \big[ \exists\: \bm{x}'\in\mathcal{B}_{\epsilon}(\bm{x}), f_{\bm{w}}(\bm{x}') \neq y \big ], 
\end{align}
where $\mathcal{B}_{\epsilon}(\bm{x}) = \{\bm{x} + \bm{\delta} \: | \: \bm{\delta} \in \mathbb{R}^d, \|\bm{\delta}\|_p \leq \epsilon\}$ stands for the perturbation ball centered at $\bm{x}$ with radius $\epsilon$ in $L_p$-norm.
\end{definition}

Note that when $\epsilon = 0$, adversarial risk reduces to the standard notion of risk. In line with the existing literature, we focus on the most widely considered $L_p$ perturbations. Correspondingly, the adversarial robustness of an ML model $f_{\bm{w}}$ can be understood as $1-\mathcal{R}_{\epsilon}(\bm{w})$.
According to Equation \ref{eq:def adversarial risk}, one can show that adversarial risk is mathematically equivalent to $\mathbb{E}_{(\bm{x},y)\sim\mathcal{D}}
[\max_{\|\bm{\delta}\|_p\le \epsilon}
\mathds{1}\{f_{\bm{w}}(\bm{x}+\bm{\delta})\ne y\}]$.

\shortsection{Adversarial Training}
To effectively train robust models with low adversarial risk, adversarial training is the most popular framework~\citep{goodfellow2014explaining,madry2017towards}. Given a set of examples $\mathcal{S}$ i.i.d. sampled from the underlying distribution $\mathcal{D}$, adversarial training is designed to minimize the following \textit{empirical adversarial loss}:
\begin{align}
\label{eq:def empirical adv loss}
    \hat{\mathcal{L}}_{\epsilon}(\bm{w}) = \frac{1}{|\mathcal{S}|} \sum_{(\bm{x},y)\in\mathcal{S}} \bigg[ \max_{\|\bm\delta\|_p \leq \epsilon} \ell(\bm{w}, \bm{x} + \bm{\delta}, y) \bigg],
\end{align}
where $\ell(\bm{w}, \bm{x} + \bm\delta, y)$ stands for surrogate loss, such as hinge or cross-entropy loss, measured at perturbed inputs. Compared with the non-differentiable objective of adversarial risk (Equation \ref{eq:def adversarial risk}), using a surrogate loss is more amenable to optimization.  
Correspondingly, we define the \textit{expected adversarial loss} over the entire data distribution $\mathcal{D}$ as:
\begin{align}
\label{eq:def expectedadv loss}
    \mathcal{L}_{\epsilon}(\bm{w}) = \mathbb{E}_{(\bm{x},y)\sim\mathcal{D}} \bigg[ \max_{\|\bm\delta\|_p \leq \epsilon} \ell(\bm{w}, \bm{x} + \bm{\delta}, y) \bigg].
\end{align}
In the existing literature, PGD is commonly used to optimize the inner maximization problem~\citep{madry2017towards}, while SGD with momentum~\citep{sutskever2013importance} is often adopted to optimize the model parameters.
Formally, the iterative model updates can be cast as: for $t=0,1,\ldots,$
\begin{align}
\label{eq:sgd_update_momentum_prelim}
    \bm{v}_{t+1} = \kappa \bm{v}_{t} + \nabla \hat{\mathcal{L}}_\epsilon(\bm{w}_{t}) + \bm{\xi}_{t}, \:
    \bm{w}_{t+1} = \bm{w}_{t} - \eta \bm{v}_{t+1}, 
\end{align}
where $\eta>0$ denotes the SGD learning rate, $\kappa\in[0,1)$ is the momentum hyperparameter, $\bm{v}_t$ denotes the velocity vector (accumulated momentum), and $\bm{\xi}_t$ stands for the difference at iterate $t$ between the stochastic gradient with respect to a random mini-batch and the full gradient $\nabla \hat{\mathcal{L}}_\epsilon(\bm{w}_t)$. 
At the initial state, $\bm{w}_0$ corresponds to the initial model parameters, usually according to some random distribution, and $\bm{v}_0 = \bm{0}$.

\shortsection{PAC-Bayesian Robust Generalization Bound} 
The PAC-Bayesian framework has been pivotal in analyzing the generalization of ML models~\citep{mcallester1999pac,neyshabur2017pac,dziugaite2017computing,xiao2023pac,alquier2024user}.
The following theorem, proven in Appendix \ref{append:proof of theorem advbound}, illustrates how to bound the expected adversarial loss over any posterior using the PAC-Bayesian framework.

\begin{theorem}[PAC-Bayesian robust generalization bound for bounded losses]
\label{thm:generic advbound}
Let $\mathcal{S}$ be a set of examples i.i.d. sampled from $\mathcal{D}$ and $\mathcal{W}\subseteq\mathbb{R}^m$ be the space of model parameters. Suppose $\mathcal{P}$ is a data-independent prior over $\mathcal{W}$, and $\mathcal{Q}$ is a posterior distribution supported on $\mathcal{W}$.
For any $\beta > 0$ and $\alpha \in (0,1)$, with probability at least $1-\alpha$, we have
\begin{equation}
\begin{aligned}
\label{eq:PAC robust generalization bound}
        \mathbb{E}_{\bm{w} \sim \mathcal{Q}} &\big[ \mathcal{L}_{\epsilon}(\bm{w}) \big] 
        \leq 
        \mathbb{E}_{\bm{w} \sim \mathcal{Q}} \big[ \hat{\mathcal{L}}_{\epsilon}(\bm{w}) \big]  \\ 
        &\qquad + \frac{1}{\beta} \mathrm{KL}(\mathcal{Q} \,\|\, \mathcal{P}) + \frac{\beta C_\ell^{2}}{8 |\mathcal{S}|} - \frac{1}{\beta} \ln{\alpha},
\end{aligned}
\end{equation}
where $C_\ell$ is a scalar upper bound on the loss function\footnote{For cross-entropy loss, one can derive similar concentration results by applying PAC-Bayes-Chernoff bounds~\citep{casado2024pac} for unbounded losses under light-tail or moment conditions.}.
\end{theorem}

Theorem \ref{thm:generic advbound} suggests that to achieve robust generalization, $\mathcal{Q}$ needs to be selected such that it can fit the empirical data well (low $\mathbb{E}_{\bm{w}\sim\mathcal{Q}} [\hat{\mathcal{L}}_{\epsilon}(\bm{w})]$) and the KL divergence between $\mathcal{Q}$ and the prior $\mathcal{P}$ is small.
To obtain a tighter bound, $\beta$ is usually set to $O\big(\sqrt{|\mathcal{S}|}\big)$ for balancing the terms.
Note that Equation \ref{eq:PAC robust generalization bound} holds for any data-independent prior and posterior distributions; the generality of the PAC-Bayesian bound enables us to later analyze the model's robust generalization performance at different stages during adversarial training.

\section{How Learning Dynamics Shape Robust Generalization?}
\label{sec:pac-bayesian robust generalization}

\subsection{Link to Loss \& Posterior Geometry}
\label{sec:link to loss and posterior geometry}

To further digest Theorem~\ref{thm:generic advbound}, we assume that the prior $\mathcal{P}$ and the posterior $\mathcal{Q}$ follow Gaussian distributions (Lemma \ref{lemma:kl_divergence}), and the adversarial loss $\hat{\mathcal{L}}_{\epsilon}(\bm{w})$ can be locally upper bounded via second-order Taylor expansion (Lemma \ref{lemma:expected empirical adversarial loss}). 
Both assumptions have been widely adopted in the prior PAC-Bayesian literature for closed-form analysis~\citep{germain2009pac,bartlett2021deep,liu2021noise,alquier2024user}.
Based on these assumptions, Equation \ref{eq:PAC robust generalization bound} can then be reduced to a more manageable form (Theorem \ref{coro:compact_pac_bound}).

\begin{lemma}
\label{lemma:kl_divergence}
    Assume the prior $\mathcal{P}$ and the posterior $\mathcal{Q}$ defined in Theorem~\ref{thm:generic advbound} follow Gaussian distributions:
    \begin{align}
    \label{eq:Gaussian prior and posterior}
        \mathcal{P} = \mathcal{N}(\bm{0}, \sigma^2_{\mathcal{P}} \mathbf{I}), \:\: \text{and} \:\: \mathcal{Q} = \mathcal{N}(\bm{\mu}_{\mathcal{Q}}, \mathbf{\Sigma}_{\mathcal{Q}}),
    \end{align}
    where $\sigma_{\mathcal{P}} \in \mathbb{R}_{+}$, $\bm{\mu}_{\mathcal{Q}} \in \mathbb{R}^m$, $\mathbf{\Sigma}_{\mathcal{Q}} \in \mathbb{R}^{m\times m}$ and $\mathbf{\Sigma}_{\mathcal{Q}} \succ 0$. 
    Then, we can derive the following closed-form expression:
    \begin{equation}
    \begin{aligned}
    \label{eq:kl divergence}        \mathrm{KL}\!\left(\mathcal{Q}\,\|\,\mathcal{P}\right) &= \frac{1}{2\sigma^{2}_{\mathcal{P}}}\big(\| \bm{\mu}_{\mathcal{Q}} \|_{2}^{2} + \mathrm{Tr}(\mathbf{\Sigma}_{\mathcal{Q}}) \big) \\
        &\qquad - \frac{1}{2} \ln \det (\mathbf{\Sigma}_{\mathcal{Q}})
        + \frac{m}{2} \ln\sigma^2_{\mathcal{P}}  - \frac{m}{2}.
    \end{aligned}
    \end{equation}
\end{lemma}
We present the proof of Lemma~\ref{lemma:kl_divergence} in Appendix~\ref{append:proof of lemma kl}.
Compared to the spherical Gaussian posteriors adopted in literature~\citep{germain2009pac,jin2022enhancing,alquier2024user},
our choice of a general covariance $\mathbf{\Sigma}_{\mathcal{Q}}$ enables an analytically tractable yet less restrictive bound that captures anisotropic parameter variability.
Note that Lemma \ref{lemma:kl_divergence} and its proof can be extended to a more general scenario where $\mathcal{Q}$ is modeled as a mixture of Gaussians (Corollary \ref{coro:mixture_pac_bound}).

To deal with the expected adversarial loss in Equation \ref{eq:PAC robust generalization bound}, we introduce a local quadratic loss assumption, which connects the term to the gradient, Hessian, and posterior structure.

\begin{lemma}
\label{lemma:expected empirical adversarial loss}
    For a given posterior $\mathcal{Q}$, assume there exists a reference point $\tilde{\bm{w}}\in\mathbb{R}^{m}$ such that, on the local region carrying the mass of $\mathcal{Q}$, the empirical loss $\hat{\mathcal{L}}_{\epsilon}(\bm{w})$ satisfies:
    \begin{equation}
    \begin{aligned}
    \label{eq:quadratic loss}
        \hat{\mathcal{L}}_{\epsilon}(\bm{w}) &\le \hat{\mathcal{L}}_{\epsilon}(\tilde{\bm{w}}) + \big\langle \nabla \hat{\mathcal{L}}_{\epsilon}(\tilde{\bm{w}}), \Delta \bm{w} \big\rangle \\
        & \qquad + \frac{1}{2} {\Delta \bm{w}}^{\top} \hat{\mathbf{H}}_{\epsilon} (\tilde{\bm{w}}) \Delta \bm{w} + r_{\tilde{\bm{w}}}(\bm{w}),
    \end{aligned}
    \end{equation}
    where $\Delta \bm{w} = \bm{w} - \tilde{\bm{w}}$, $\nabla \hat{\mathcal{L}}_{\epsilon}(\tilde{\bm{w}})$ is the gradient of the empirical adversarial loss, $\hat{\mathbf{H}}_{\epsilon} (\tilde{\bm{w}})$ is the Hessian matrix, and $r_{\tilde{\bm{w}}}(\bm{w})$ is the higher-order remainder term. Then we have
    \begin{align}
    \label{eq:expected empirical adversarial loss}
        \nonumber &\mathbb{E}_{\bm{w}\sim\mathcal{Q}}
        \big[\hat{\mathcal{L}}_{\epsilon}(\bm{w})\big]
        \le\; \hat{\mathcal{L}}_{\epsilon}(\tilde{\bm{w}})
        + \big\langle
        \nabla \hat{\mathcal{L}}_{\epsilon}(\tilde{\bm{w}}),
        \Delta \bar{\bm{w}}
        \big\rangle \\
        & \quad + \frac{1}{2} {\Delta \bar{\bm{w}}}^{\top}
        \hat{\mathbf{H}}_{\epsilon}(\tilde{\bm{w}})
        \Delta \bar{\bm{w}} + \frac{1}{2}
        \mathrm{Tr}\big(
        \hat{\mathbf{H}}_{\epsilon}(\tilde{\bm{w}})
        \mathbf{\Sigma}_{\mathcal{Q}}
        \big) + R_{\mathcal{Q}}(\tilde{\bm{w}}),
    \end{align}
    where $\Delta \bar{\bm{w}} = \bm{\mu}_{\mathcal{Q}} - \tilde{\bm{w}}$ is the posterior mean drift from the reference point $\tilde{\bm{w}}$ and $R_{\mathcal{Q}}(\tilde{\bm{w}})=\mathbb{E}_{\bm{w}\sim\mathcal{Q}}[r_{\tilde{\bm{w}}}(\bm{w})]$.
\end{lemma}

Here, the reference point $\tilde{\bm{w}}$ can be understood as the local coordinate anchor for the posterior being analyzed. Once the posterior is fixed, $\tilde{\bm{w}}$ is chosen in the same local basin or short-time window such that $\bm{w}-\tilde{\bm{w}}$ is small for most posterior mass.
Lemma \ref{lemma:expected empirical adversarial loss}, with its detailed proof provided in Appendix~\ref{sec:proof expected empirical adversarial loss}, illustrates that the posterior-averaged empirical adversarial loss is regulated by the local geometry of the loss landscape and the posterior covariance. 
Similar decomposition extends to Gaussian-mixture posteriors by applying the expansion within each component (Corollary~\ref{coro:mixture_pac_bound}).

Compared with the standard loss, the inner maximization operator unique to the adversarial loss (Equation \ref{eq:def empirical adv loss}) complicates the derivation of the Hessian $\hat{\mathbf{H}}_{\epsilon}$.
Adopting a local quadratic loss decomposition will largely simplify our later analysis while capturing the leading variations.
Strictly speaking, $\hat{\mathbf{H}}_{\epsilon}$ may not exist everywhere; however, note that standard neural networks are usually piecewise smooth (e.g., ReLU-based), the adversarial loss can thus be considered a maximum over piecewise-smooth functions and is therefore twice differentiable almost everywhere. This implies that Equation \ref{eq:expected empirical adversarial loss} holds for almost all inputs and parameters.
In our experiments, we adopt the common approach to estimate the Hessian by ignoring the second-order dependence of $\bm{\delta}^*$ on $\bm{w}$ using a Danskin-type argument and employing PGD to approximate the inner maximization~\citep{liu2020loss}.

Equipped with Lemmas \ref{lemma:kl_divergence} and \ref{lemma:expected empirical adversarial loss}, we can now reduce the generic PAC-Bayesian robust generalization bound in Theorem \ref{thm:generic advbound} to the following theorem proven in Appendix~\ref{append:proof_compact_pac_bound}.

\begin{theorem}
\label{coro:compact_pac_bound}
Let $\mathcal{P} = \mathcal{N}(\bm{0}, \sigma^2_{\mathcal{P}} \mathbf{I})$ and $\mathcal{Q}$ be a posterior that satisfies the conditions in Lemmas \ref{lemma:kl_divergence} and \ref{lemma:expected empirical adversarial loss}. For any $\beta > 0$, $\alpha \in (0,1)$, with probability at least $1-\alpha$, we have
\begin{equation}
\begin{aligned}
\label{eq:compact_pac_bound}
    &\mathbb{E}_{\bm{w}\sim\mathcal{Q}}
    \big[\mathcal{L}_{\epsilon}(\bm{w})\big]
    \leq\; \hat{\mathcal{L}}_{\epsilon}(\tilde{\bm{w}}) + \big\langle \nabla \hat{\mathcal{L}}_{\epsilon}(\tilde{\bm{w}}), \Delta_{\mathcal Q} \big\rangle
    \\
    & \:\: + \frac{1}{2}\Delta_{\mathcal Q}^{\top}
    \hat{\mathbf{H}}_{\epsilon}(\tilde{\bm{w}})\Delta_{\mathcal Q} + \frac{1}{2}\mathrm{Tr}\big(
    \hat{\mathbf{H}}_{\epsilon}(\tilde{\bm{w}})
    \mathbf{\Sigma}_{\mathcal{Q}} \big) + R_{\mathcal{Q}}(\tilde{\bm{w}}) \\
    & \:\: + \frac{1}{2\beta \sigma^{2}_{\mathcal{P}}}
    \big(\| \bm{\mu}_{\mathcal{Q}} \|_{2}^{2}
    + \mathrm{Tr}(\mathbf{\Sigma}_{\mathcal{Q}}) \big)  - \frac{1}{2\beta} \ln \det(\mathbf{\Sigma}_{\mathcal{Q}}) + C_{\mathrm{PB}},
\end{aligned}
\end{equation}
where $\Delta_{\mathcal Q}=\bm{\mu}_{\mathcal{Q}}-\tilde{\bm{w}}$, and $C_{\mathrm{PB}}$ collects all the terms that are independent of the posterior $\mathcal{Q}$, formally defined as:
\begin{align*}
    C_{\mathrm{PB}} =
    \frac{\beta C_\ell^2}{8|\mathcal{S}|}
    -\frac{1}{\beta}\ln\alpha
    +\frac{m}{2\beta}\ln\sigma_{\mathcal P}^2
    -\frac{m}{2\beta},
\end{align*}
and $C_{\ell}$ is a constant introduced in Theorem \ref{thm:generic advbound}.
\end{theorem}

Theorem \ref{coro:compact_pac_bound} connects the PAC-Bayesian robust generalization bound to the internal adversarial training dynamics (see Corollary \ref{coro:mixture_pac_bound} for the extension to Gaussian mixtures). Specifically, the expected adversarial loss over the posterior is controlled by three main components: (i) first- and second-order biases, (ii) curvature-weighted variance, and (iii) KL-divergence related terms, capturing the interactions between the local geometry of the loss landscape (gradient and Hessian) and the posterior (mean drift and covariance).
During adversarial training, the local geometry of the loss landscape and posterior can vary drastically, necessitating a more fine-grained temporal analysis of their evolution.

\subsection{Time-Varying Posterior Dynamics}
\label{sec:posterior dynamics analysis}

While Theorem \ref{coro:compact_pac_bound} relates robust generalization bounds to specific local geometry-induced properties, the posterior-related terms remain obscure. 
In this section, we illustrate how to use dynamical system modeling techniques to derive closed-form solutions of the posterior dynamics under both stationary and non-stationary transient regimes.

\shortsection{Stationary Regime}
\phantomsection
\label{sec:stationary}
Let $\mathcal{Q}_{t}$ denote the posterior distribution at timestep $t$ induced by the momentum SGD update rule (Equation~\ref{eq:sgd_update_momentum_prelim}). We say the learning system reaches a \emph{stationary} state at time $t_0$ if $\mathcal{Q}_t=\mathcal{Q}_{t_0}$ for any $t\ge t_0$.
The following lemma, proven in Appendix~\ref{append:proof_cov_param_projection_detailed}, characterizes the state-space representation of the posterior dynamics, as well as the associated mean $\bm{\mu}$ and covariance $\mathbf{\Sigma}$ at stationarity.

\begin{lemma}
\label{lemma:cov_param_projection_detailed}
Suppose an ML system defined by Equation \ref{eq:sgd_update_momentum_prelim} reaches a stationary state at time $t_0$. Denote by the joint state vector $\bm{u}_t = [\bm{w}_t - \tilde{\bm{w}}; \bm{v}_t] \in \mathbb{R}^{2m}$. Under the local quadratic loss assumption as in Lemma \ref{lemma:expected empirical adversarial loss}, we have 
\begin{align}
     \forall t \geq t_0, \:\: \bm{u}_{t+1} = \mathbf{A} \bm{u}_{t} + \mathbf{G} \big[ \nabla \hat{\mathcal{L}}_{\epsilon}(\tilde{\bm{w}}) + \bm{\xi}_{t} \big],
\end{align}
where $\bm{\xi}_t$ is the mini-batch gradient noise at time $t$. Assume the mini-batch noise forms a martingale-difference sequence conditioned on the training set, $\mathbb{E}[\bm{\xi}_t\mid\mathcal{F}_t,\mathcal{S}]=\bm{0}$, where $\mathcal{F}_t$ is the filtration generated by the past training trajectory, and $\mathbf{A}$ and $\mathbf{G}$ are the transition matrix and bias vector of the dynamical system, respectively defined as:
\begin{align}
\label{eq:posterior covariance iterates}
    \mathbf{A} =
    \begin{bmatrix}
    \mathbf{I} - \eta \hat{\mathbf{H}}_{\epsilon}(\tilde{\bm{w}}) & -\eta\kappa\mathbf I \\
    \hat{\mathbf{H}}_{\epsilon}(\tilde{\bm{w}}) & \kappa \mathbf I
    \end{bmatrix}, \:\:
    \mathbf G=\begin{bmatrix}-\eta\mathbf I\\ \mathbf I\end{bmatrix}.
\end{align}
Let $\mathbf{C} = \lim_{t\rightarrow \infty} \mathbb{E}[\bm{\xi}_t \bm{\xi}_t^{\top}\mid\mathcal{S}]$ denote the finite stationary gradient-noise covariance. Assume $\mathbf{H}_{\rm ref}:=\hat{\mathbf{H}}_{\epsilon}(\tilde{\bm{w}})\succ0$, $\mathbf{C}$ commutes with $\mathbf{H}_{\rm ref}$, and every eigenvalue $\lambda_i$ of $\mathbf{H}_{\rm ref}$ satisfies $0<\eta\lambda_i<2(1+\kappa)$, $i=1,2,\ldots,m$. Then, we can derive the stationary posterior mean and covariance:
\begin{align}
    \label{eq:stationary posterior mean} \bm{\mu} &= \tilde{\bm{w}} - \big[ \hat{\mathbf{H}}_{\epsilon}(\tilde{\bm{w}}) \big]^{-1} \nabla \hat{\mathcal{L}}_{\epsilon}(\tilde{\bm{w}}), \\
     \label{eq:stationary posterior covariance} \bm{\Sigma} &= \bigg[ \hat{\mathbf{H}}_{\epsilon}(\tilde{\bm{w}}) \bigg(2 \mathbf{I} - \frac{\eta}{1+\kappa} \hat{\mathbf{H}}_{\epsilon}(\tilde{\bm{w}}) \bigg) \bigg]^{-1} \frac{\eta}{1-\kappa} \mathbf{C}.
\end{align}
\end{lemma}

Lemma~\ref{lemma:cov_param_projection_detailed} reveals two key properties regarding the stationary posterior. First, Equation \ref{eq:stationary posterior mean} indicates that the stationary mean involves a curvature-weighted gradient correction with respect to the reference $\tilde{\bm{w}}$. When translated into first- and second-order bias terms, it yields a Newton-type optimization gain ($- \frac{1}{2} \nabla \hat{\mathcal{L}}_{\epsilon}^{\top} \hat{\mathbf{H}}_{\epsilon}^{-1} \nabla \hat{\mathcal{L}}_{\epsilon}$), acting as a constant force for lowering the empirical adversarial loss within the stationary trajectory.  
Second, Equation~\ref{eq:stationary posterior covariance} makes explicit how learning rate $\eta$, Hessian $\hat{\mathbf{H}}_{\epsilon}$, and noise covariance $\mathbf{C}$ jointly shape the posterior spread: for each eigendirection of $\hat{\mathbf{H}}_{\epsilon}$, larger eigenvalues accelerate the contraction while larger noise variances and learning rates broaden it.
To ensure the covariance is well defined at stationarity, the Hessian eigenvalues need to stay within the range of $0<\lambda_i<2(1+\kappa)/\eta$. This stability condition implicitly imposes a regularization effect on the loss curvature when the learning rate $\eta$ is large (prior to the first learning-rate decay in adversarial training).

\shortsection{Non-Stationary Regime}
\phantomsection
\label{sec:non-stationary}
Although stationary analysis is the primary focus for analyzing SGD dynamics in prior literature~\citep{liu2021noise,ziyin2021strength}, it is insufficient to capture the non-stationary transition during the onset of robust overfitting, where the learning rate sharply decays and the system drifts away from its previous stationary state. 
Note that the transition matrix and noise covariance depend on the local loss geometry with reference to $\tilde{\bm{w}}$, which can vary across time under non-stationary regimes. 
To obtain a tractable approximation, we assume the system remains approximately invariant over short time windows and characterize the posterior dynamics via iterative linearization.

\begin{lemma}
\label{lemma:windowed_mean_covariance_recursion}
Let $\mathcal{Q}_t$ be the posterior induced by the SGD update rule in Equation \ref{eq:sgd_update_momentum_prelim} at timestep $t$. Assume that the learning rate $\eta_t$ and Hessian $\mathbf{H}_t:=\hat{\mathbf{H}}_\epsilon(\tilde{\bm{w}}_t)$ remain fixed over the next $k$ updates. Assume that $\{\bm{\xi}_s\}_{s=t}^{t+k-1}$ is a martingale-difference sequence conditional on $\mathcal{S}$, $\mathbb{E}[\bm{\xi}_s\mid\mathcal{F}_s,\mathcal{S}]=\bm{0}$, and that its conditional covariance is fixed at $\mathbf{C}_t$ within the window, where $\mathbf{C}_t$ is finite and $0<\eta_t\lambda_{t,i}<2(1+\kappa)$ for every analyzed mode. 
Consider the joint state vector $\bm{u}_{t} = [\bm{w}_{t} - \tilde{\bm{w}_t}; \bm{v}_{t}]$ and let $\bm{\Omega}_t = \mathrm{Cov}(\bm{u}_t)$ be its covariance. Then, at timestep $t' = t+k$, the posterior $\mathcal{Q}_{t'}$ satisfies:
\begin{align}
    \label{eq:non-stationary mean iteration} &\mathbb{E} [\bm{u}_{t'}]  
    = \mathbf{A}_t^k \: \mathbb{E}[{\bm{u}}_t]
    + {\sum\nolimits_{j=0}^{k-1}} \mathbf{A}_t^j \mathbf{G} \nabla \hat{\mathcal{L}}_{\epsilon}(\tilde{\bm{w}}_t), \\
    \label{eq:non-stationary cov iteration} &\bm{\Omega}_{t'} = \mathbf{A}_t^k \bm{\Omega}_{t} (\mathbf{A}_t^{k})^{\top} + {\sum\nolimits_{j=0}^{k-1}} \mathbf{A}_t^j \mathbf{G} \mathbf{C}_t \mathbf{G}^{\top}(\mathbf{A}_t^{j})^{\top}.
\end{align}
Here, $\mathbf{A}_t$ denotes transition matrix, $\mathbf{C}_t = \mathbb{E}[\bm{\xi}_s \bm{\xi}_s^{\top}\mid\mathcal{F}_s,\mathcal{S}]$ is the conditional noise covariance, and $\mathbf{G}$ is the bias vector.
\end{lemma}

Lemma \ref{lemma:windowed_mean_covariance_recursion}, proven in Appendix~\ref{append:proof_windowed_mean_covariance_recursion}, illustrates how the posterior evolves within the time window $[t, t']$, where the mean drift and covariance at time $t'$ can be derived by projecting onto the corresponding state space: $\bm{\mu}_{t'} = \mathbf{\Pi} \mathbb{E}[\bm{u}_{t'}]$ and $\mathbf{\Sigma}_{t'} = \mathbf{\Pi}  \mathbf{\Omega}_{t'}  \mathbf{\Pi}^{\top}$, where $\mathbf{\Pi} = [\mathbf{I} \;\; \mathbf{0}]$. 
It is reasonable to assume the transition matrix is time-invariant over a short time period, as the learning rate $\eta$ is often very small when robust overfitting happens for adversarial training algorithms, suggesting that the system is slowly evolving during the transient regime. To approximate the posterior dynamics at an arbitrary timestep, one can apply Lemma \ref{lemma:windowed_mean_covariance_recursion} to a sequence of consecutive time windows to accumulate the mean drift and posterior covariance fluctuation. 

\shortsection{Final Robust Generalization Bounds}
Now that we have derived the closed-form solutions for the stationary posterior and its evolution during the non-stationary transient stage, we can formalize the PAC-Bayesian robust generalization bound into the following theorem, proven in Appendix~\ref{append:proof_transient_bound}.

\begin{theorem}
\label{thm:final robust generalization bound combined}
Consider the dynamical system defined by the momentum SGD update rule in Equation \ref{eq:sgd_update_momentum_prelim}.
Let $\mathcal{Q}_t$ be the posterior at timestep $t$ with mean $\bm{\mu}_t$ and covariance $\mathbf{\Sigma}_t$. Let $\bm{g}_t=\nabla\hat{\mathcal{L}}_\epsilon(\tilde{\bm{w}}_t)$, $\mathbf{H}_t=\hat{\mathbf{H}}_\epsilon(\tilde{\bm{w}}_t)$, and $\Delta_t=\bm{\mu}_t-\tilde{\bm{w}}_t$. Suppose the assumptions in Lemmas \ref{lemma:kl_divergence} and \ref{lemma:expected empirical adversarial loss} are satisfied; in particular, $\sigma_{\mathcal P}>0$ and $\mathbf{\Sigma}_t\succ0$. 
For any $\beta>0$ and $\alpha\in(0,1)$, with probability at least $1-\alpha$, we have
\begin{align}
\label{eq:robust generalization bound combined}
    \nonumber &\mathbb{E}_{\bm{w}\sim\mathcal{Q}_{t}} \big[\mathcal{L}_{\epsilon}(\bm{w})\big]
    \leq \hat{\mathcal{L}}_{\epsilon}(\tilde{\bm{w}}_t)
    + \big\langle \bm{g}_t, \Delta_t \big\rangle \\
    \nonumber & \quad + \frac{1}{2} \Delta_t^{\top} \mathbf{H}_t \Delta_t + \frac{1}{2} \operatorname{Tr}(\mathbf{H}_t\mathbf{\Sigma}_t)
    + \frac{1}{2\beta \sigma_{\mathcal P}^2}\operatorname{Tr}(\mathbf{\Sigma}_t) \\
     & \quad
    + \frac{1}{2\beta \sigma_{\mathcal P}^2} \|\bm{\mu}_t\|_2^2
    - \frac{1}{2\beta} \ln\det(\mathbf{\Sigma}_t) + R_t + C_{\mathrm{PB}},
\end{align}
where $R_t=R_{\mathcal{Q}_t}(\tilde{\bm{w}}_t)$ and $C_{\mathrm{PB}}$ are defined in Lemma~\ref{lemma:expected empirical adversarial loss} and Theorem \ref{coro:compact_pac_bound}, respectively. 
\end{theorem}

\begin{algorithm*}[t]
\caption{Spectral estimation of bound components}
\label{alg:empirical_estimation}
\begin{algorithmic}[1]
\STATE \textbf{Input:} model checkpoint $\bm{w}$, rank $k$, hyperparameters used for PGD attack, number of sampled mini-batches $M$.
\STATE Generate adversarial mini-batches and compute gradients and Hessian--vector products of $\hat{\mathcal{L}}_{\epsilon}$ at $\bm{w}$.
\STATE Estimate top-$k$ Hessian eigenpairs $\{(\lambda_i,\bm{v}_i)\}_{i=1}^k$ by power iteration using Hessian--vector products.
\STATE Let $\mathbf{V}=[\bm{v}_1,\dots,\bm{v}_k]$. For each mini-batch $b\in\{1,\dots,M\}$, compute $\bm{g}_b$ and $\bm{p}_b=\mathbf{V}^\top \bm{g}_b$.
\STATE Compute $\Gamma=\mathrm{Cov}_b(\bm{p}_b)$ and set $\gamma_i=\Gamma_{ii}$.
\STATE Compute $(\sigma_i^2,\Delta \bar{\bm{w}}_i)$ using stationary formulas (Lemma \ref{lemma:cov_param_projection_detailed}) or non-stationary transient recursions (Lemma \ref{lemma:windowed_mean_covariance_recursion}).
\STATE Form the variance/entropy proxies and bias terms by summing contributions over $i\le k$.
\STATE \textbf{Output:} $\{\lambda_i\}$, $\{\gamma_i\}$, $\{\sigma_i^2\}$, and decomposed bound components.
\end{algorithmic}
\end{algorithm*}

Plugging the derivations of posterior mean and covariance in Lemma \ref{lemma:cov_param_projection_detailed} (stationary) or Lemma \ref{lemma:windowed_mean_covariance_recursion} (non-stationary transient) into Theorem \ref{thm:final robust generalization bound combined}, we can analyze each term associated with the PAC-Bayesian robust generalization bounds and how they evolve during adversarial training.
It is worth noting that closed-form bound component derivations make use of the assumption that the Hessian $\mathbf{H}_t$ and the covariance matrices ($\mathbf{C}_t$ and $\mathbf{\Sigma}_t$) satisfy the commutativity condition (see Figure~\ref{fig:commutativity_alignment} in Appendix~\ref{app:comm_align} for supporting experiments), suggesting that they can be simultaneously diagonalized. 
Specifically, we can show that the trace and determinant terms in Equation \ref{eq:robust generalization bound combined} can be computed as:
\begin{align*}
    & \frac{1}{2}\mathrm{Tr}(\mathbf{H}_t \mathbf{\Sigma}_t) = \frac{1}{2}\sum\nolimits_{i=1}^m \lambda_{t,i}\,\sigma_{t, i}^2, \\ 
    & \frac{1}{2\beta\sigma_{\mathcal{P}}^2} \mathrm{Tr}(\mathbf{\Sigma}_t) = \frac{1}{2\beta\sigma_{\mathcal{P}}^2} \sum\nolimits_{i=1}^m \sigma_{t,i}^2, \\
    & -\frac{1}{2\beta} \ln\det(\mathbf{\Sigma}_t) = -\frac{1}{2\beta}\sum\nolimits_{i=1}^m \ln \sigma_{t,i}^2,
\end{align*}
where $\lambda_{t,i}$ and $\sigma_{t,i}^2$ denote the $i$-th eigenvalues of $\mathbf{H}_t$ and $\mathbf{\Sigma}_t$, respectively. 
In our experiments, we first estimate the top-$k$ Hessian eigenvalues for each intermediate model and the corresponding projected noise-covariance eigenvalues, then approximate each relevant term\footnote{We ignore $\hat{\mathcal{L}}_{\epsilon}(\tilde{\bm{w}}_t)$, $R_t$ and $C_{\mathrm{PB}}$ in our later diagnostic analysis, since they are either independent of $\mathcal{Q}$ or have negligible impact on robust generalization gap compared to remaining terms.} in the derived bounds.

\begin{figure*}[t]
    \centering
    \includegraphics[width=\linewidth]{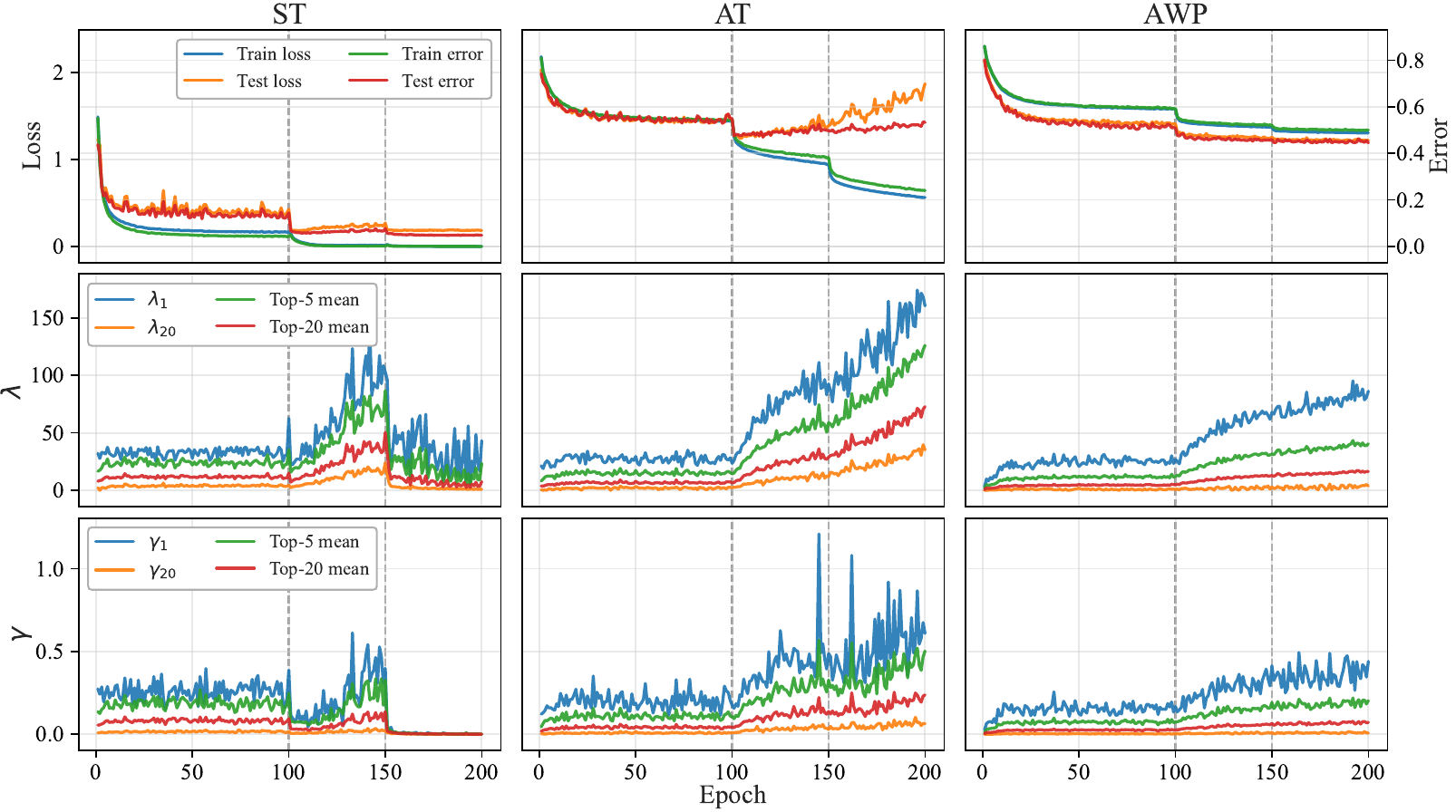}

    \caption{Learning curves across different learning algorithms on CIFAR-10: (first row) training/testing loss and error curves (ST: clean; AT/AWP: robust), (middle row) Hessian eigenvalues, and (last row) projected gradient noise variances.}
    \label{fig:unified_spectra}
\end{figure*}

\begin{figure*}[t]
    \centering
    \includegraphics[width=\linewidth]{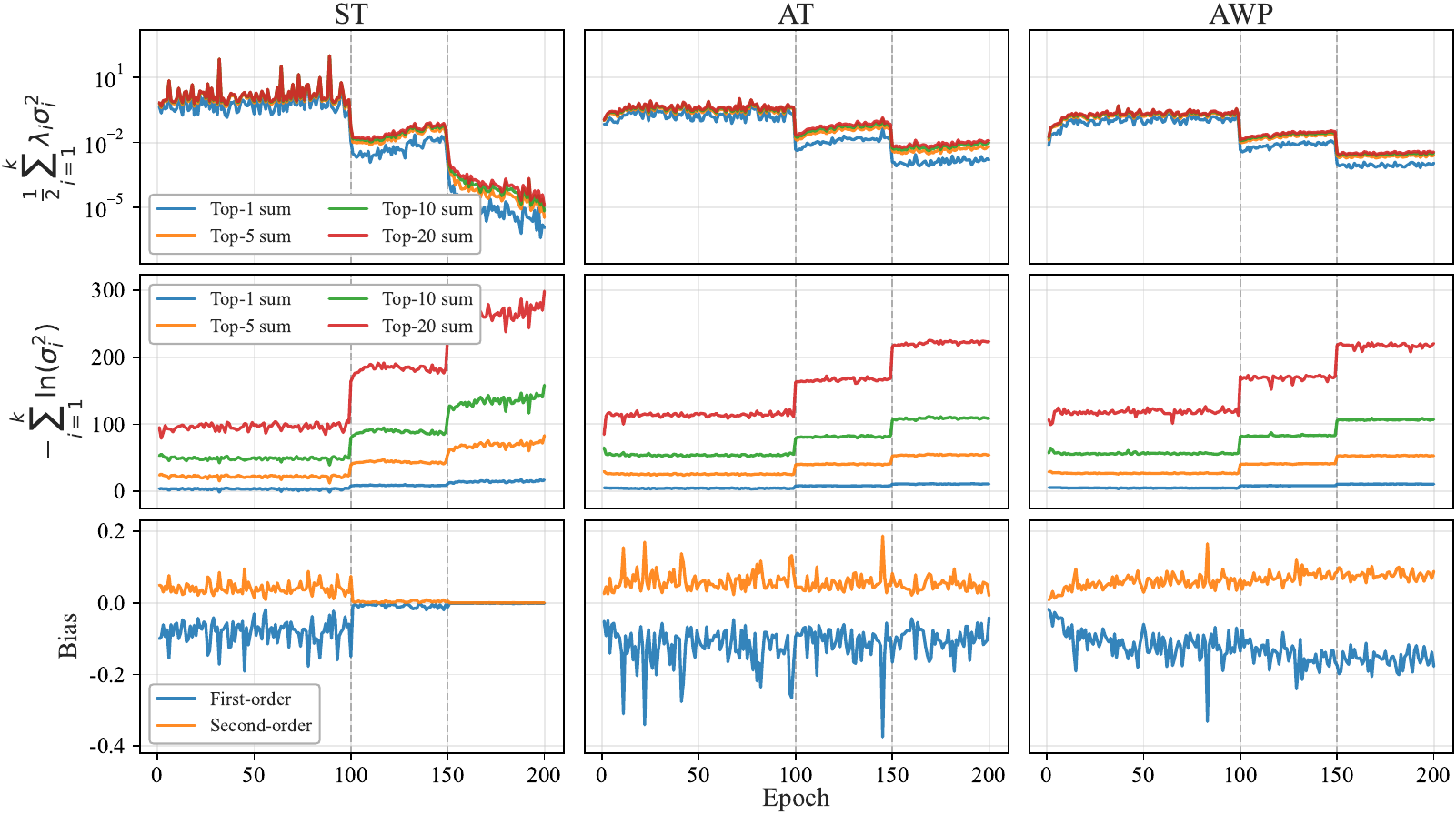}
    \caption{Learning curves of top-$k$ bound diagnostics across different algorithms on CIFAR-10: (top row) curvature-weighted variance, (middle row) entropic KL, and (bottom row) first- and second-order bias terms.}
    \label{fig:unified_bound}
\end{figure*}

\section{Experiments}
\label{sec:experiments}

We focus our main experiments on CIFAR-10 and $L_{\infty}$ perturbations with $\epsilon = 8/255$ to study the robust generalization bounds, where we compare three learning algorithms: standard training (ST), adversarial training (AT)~\citep{madry2017towards}, and adversarial weight perturbation (AWP)~\citep{wu2020adversarial}, all trained on a PreActResNet-18 architecture for $200$ epochs using momentum SGD ($\kappa=0.9$) with batch size $128$ and weight decay $5\times 10^{-4}$.
Aligned with \citet{rice2020overfitting}, we adopt a piecewise scheduler with an initial learning rate of $0.1$, decayed to $0.01$ at epoch $100$ and to $0.001$ at epoch $150$.
For AT and AWP, we evaluate the robust losses and errors with the intermediate models at each epoch, while we record the clean counterparts for ST (see Appendix~\ref{app:exp_settings} for details). 
Appendix~\ref{app:generalizability} reports all the additional experiments, where we vary datasets (CIFAR-100, SVHN, Imagenette-160), algorithms (TRADES, semi-supervised AT), perturbation radii, batch sizes, and learning-rate schedules.

\subsection{Spectral Estimation}
\label{sec:exp_estimation}

A computational challenge is efficiently estimating the statistical quantities, including the loss-geometry-related gradient and Hessian terms, as well as the posterior mean and covariance. 
Since the number of parameters $m$ is huge for modern deep neural networks, computing the full Hessian matrix at a single checkpoint is very expensive. Therefore, we adopt a more efficient \textit{per-epoch spectral estimation} protocol based on power iterations and Hessian-vector products to approximate the top-$k$ eigensubspace with respect to the model checkpoint, to capture the Hessian's important structure while reducing computational cost. Accordingly, other quantities are projected onto the same subspace and then incorporated to obtain an estimate for each decomposed term of the robust generalization bounds in Theorem \ref{thm:final robust generalization bound combined}.
Algorithm \ref{alg:empirical_estimation} demonstrates the adopted estimation procedure.

In particular, at each epoch \(t\), we first estimate the top-\(k\) Hessian eigenpairs
\(\{(\lambda_i(t),v_i(t))\}_{i=1}^k\) with respect to the empirical loss by power iteration using
Hessian-vector products.
Note that for adversarial training algorithms, eigenpairs are computed with respect to the empirical PGD-based adversarial loss, whereas the unperturbed clean loss is used for standard training. 
Next, we approximate the associated gradient-noise covariances by sampling mini-batch gradients $\bm{g}_b(t)$ and projecting them onto the same eigensubspace $\mathbf{V}_t = [\bm{v}_1(t), \ldots, \bm{v}_k(t)]$. 
Eventually, we obtain \(\{\gamma_i(t)\}_{i=1}^k\), where
\(\gamma_i(t)\) is the diagonal entry of the projected mini-batch
gradient-noise covariance in the \(i\)-th Hessian eigendirection.

These estimates $\{\lambda_i(t),\gamma_i(t)\}_{i=1}^k$ will be used to compute the associated posterior variances $\{\sigma_i^2\}_{i=1}^k$, which capture the parameter fluctuations along each top-$k$ eigendirection, through the stationary covariance expression (\Eqref{eq:stationary posterior covariance}) and the post-decay transient recursion (\Eqref{eq:non-stationary cov iteration}). 
We adopt a similar scheme to approximate the posterior mean drift at stationarity (\Eqref{eq:stationary posterior mean}) and the non-stationary variants (\Eqref{eq:non-stationary mean iteration}). Finally, all the above estimates will be translated into Equation~\ref{eq:robust generalization bound combined} to characterize the role of each decomposition in the derived PAC-Bayesian bounds. 

Detailed estimation steps, including batch sampling and power-iteration configurations, and an analysis of its stability with respect to $k$ are provided in Appendix~\ref{app:spectral_details}.
Our estimation protocol is an offline checkpoint diagnostic, not a training-time method: it avoids forming the full Hessian and full noise covariance, and its cost scales with the number of evaluated checkpoints, retained modes, power iterations, and sampled mini-batches (see Appendix~\ref{app:runtime_scalability} for details). 

\subsection{Progressive Curvature Sharpening}
\label{sec:exp_lifecycle}

Figure~\ref{fig:unified_spectra} visualizes the learning curves of statistics, including training and testing error/loss, top-$k$ Hessian eigenvalues $\{\lambda_i\}_{i=1}^k$, and the corresponding projected gradient-noise variances $\{\gamma_i\}_{i=1}^k$, across different algorithms (ST, AT, and AWP).
Note that the $i$-th pair $(\lambda_i, \gamma_i)$ is associated with the $i$-th leading eigenvector of the Hessian matrix at the evaluated epoch. We follow the spectral estimation protocol described in Section \ref{sec:exp_estimation} to obtain these estimates.
To illustrate the dominating trend while keeping presentation cleanliness, we report the following summary spectral statistics: top-1, top-20, the mean of top-5, and the mean of top-20.

Notably, adversarial training depicts a stage-wise curve separated by the learning rate change. 
During the initial phase, both the robust loss and error steadily decrease until reaching a plateau, while the top Hessian eigenvalues and gradient noise stay small. After the learning rate decays by a factor of $10$, both $\lambda_i$ and $\gamma_i$ exhibit a monotonically increasing trend, corresponding to a continual decrease in training robust error/loss but steadily worsening test statistics. The final stage is characterized by the smallest $\eta=0.001$, the top Hessian eigenvalue remaining very large and continuing to increase, and the test robust accuracy dropping further.
These observations align with the local stability condition in Section \ref{sec:posterior dynamics analysis}: top Hessian eigenvalues are capped by $2(1+\kappa)/\eta$,  which is inversely proportional to the learning rate; a large learning rate $\eta$ will restrict the optimization in low-curvature regions, while reducing $\eta$ will relax the constraint, allowing the model to move into regions with higher curvature.

In sharp contrast to AT, the learning curves of ST demonstrate a different trend, especially in the final learning stage. Top Hessian eigenvalues are also progressively sharpened when $\eta=0.01$, indicating the model is gradually fitting more fine-grained features within the training data; however, both $\lambda_i$ and $\gamma_i$ suddenly drop to close to zero and remain very low when $\eta$ drops to $0.001$. We note that the learning curves of Hessian eigenspectrum match the double descent phenomenon~\citep{nakkiran2021deep}, where SGD noise implicitly biases towards flat regions for over-parameterized neural networks.
In addition, AWP penalizes the model's sharpness by perturbing weight parameters, leading to suppressed growth of both $\lambda_i$ and $\gamma_i$ in later training stages. 

In our experiments, we observe that the
top Hessian eigenvalues of adversarially trained models are
much higher than those of standard trained models, especially when $\eta$ is small. This suggests that optimizing adversarial training losses requires exploring high-curvature
regions in the parameter space. The following proposition,
proven in Appendix~\ref{sec:proof proposition adversarial training -> Hessian increase}, provides a theoretical justification for the above statement.

\begin{proposition}
\label{prop:robust optimization & curvature}
Optimizing the adversarial training loss $\hat{\mathcal{L}}_{\epsilon}(\bm{w})$ under $L_2$ or $L_\infty$ norm to sufficiently low requires the top Hessian eigenvalues to be large, especially due to the increased cross-sample alignment of input and parameter gradients with respect to the robust feature subspace.
\end{proposition}

Proposition \ref{prop:robust optimization & curvature} shows that to fit the training data well, robust learning algorithms need to explore high-curvature regions. Unlike ST, it is not possible for adversarial training to learn a model that is “flat” in all directions; a few large Hessian eigenvalues associated with robust features are necessary to ensure the input gradient sensitivity is sufficiently penalized.

\subsection{Posterior Collapse \& Variance Amplification}

Based on the per-epoch spectral estimates, we can now empirically estimate the posterior mean, posterior covariance, and the bound decompositions in Theorem \ref{thm:final robust generalization bound combined}. 
The full bound contains KL terms beyond the log-determinant proxy, including the mean-norm term and the covariance-trace shift $\lambda_{t,i}\mapsto \lambda_{t,i}+1/(\beta\sigma_{\mathcal P}^2)$ (see Appendix~\ref{app:full_kl_diagnostics} for full diagnostics).
Specifically, we approximate the spectral estimates using top-$k$ partial sums with $k\in\{1,5,10,20\}$ for both the curvature-weighted variance term $\frac12\sum_{i=1}^k\lambda_i\sigma_i^2$ and the entropic KL term $-\sum_{i=1}^k\ln(\sigma_i^2)$. 
We use the stationary estimates by default and switch to non-stationary recursion during the transient phase after $\eta$ decays (Appendix \ref{app:spectral_details}).

Figure \ref{fig:unified_bound} visualizes the learning curves across different algorithms.
For AT, we observe that the inferred top-$k$ posterior covariance contracts sharply after $\eta$ decays at epochs $100$ and $150$, reflected by the stark increase in the entropic KL term and the sharp drop in the variance.  This is evident if we derive the closed-form solutions for both terms under the stationary regime and assume commutativity (Lemma \ref{lemma:cov_param_projection_detailed}). Formally, for any $i=1,2,\ldots,k$, we have
\begin{align*}
    \lambda_{i}\,\sigma_{i}^2
    &= \bigg( 2- \frac{\eta\lambda_i}{1+\kappa}\bigg)^{-1}
    \frac{\eta}{1-\kappa} \gamma_i, \\
    \ln (\sigma_{i}^2)
    &= \ln  \Big(\frac{\eta} {1-\kappa}\Big)
    + \ln \bigg( \frac{\gamma_i}{\lambda_i} \bigg)  - \ln \bigg(2- \frac{\eta\lambda_i}{1+\kappa}\bigg).
\end{align*}
Appendix \ref{app:comm_align} provides empirical analysis examining the commutativity assumption, showing that the top Hessian eigendirections largely align with the gradient noise covariance.
When $\eta$ suddenly drops by a factor of $10$, $\lambda_i$ and $\gamma_i$ remain similar, so the immediate change is primarily associated with the drop in $\eta$. Note that the decrease in $\lambda_{i}\,\sigma_{i}^2$ is approximately linear in $\eta$ (dominating factor), whereas the rise in $-\ln (\sigma_{i}^2)$ is logarithmic in $\eta$ (less drastic), which explains the initial drop of the testing curves in Figure \ref{fig:unified_spectra}. 

As adversarial training continues, the curvature-weighted variance gradually increases, which coincides with increases in $\lambda_i$ and $\gamma_i$. In comparison, the entropic KL term remains roughly constant during each stage.
Note that, as illustrated in Figure \ref{fig:unified_spectra}, the top Hessian eigenvalues rise by a factor of $4$ from epoch $100$ to epoch $150$ while noise variances increase by a factor of $2$, jointly boosting the variance at the later stages of adversarial training. The observed trend aligns with the test-robust error curve during the onset of robust overfitting after the learning rate decay, suggesting that it has a dominating influence on robust generalization. Besides, the first- and second-order biases remain at a consistent level throughout training, with only a few large spikes observed at certain epochs. We hypothesize that the spike is due to the (mis-)alignment of input gradients across minibatches.

Compared across the three tested learning algorithms, ST exhibits greater posterior contraction than AT, almost vanishing when $\eta = 0.001$. Such a collapsing posterior leads to a continuous decrease in curvature-weighted variance. The two bias terms approach zero after the learning rate decays. AWP, on the other hand, penalizes loss curvature, which benefits a more controlled variance and explains why it improves robust generalization. Interestingly, the two bias terms diverge gradually, suggesting that AWP may underfit the training objectives due to excessive penalization. These observations align with AWP’s high training loss observed in Figure~\ref{fig:unified_spectra}. Consequently, a promising future direction for our work is to develop a selective penalization scheme for AWP to reduce the training robust loss while preserving its generalization benefits. Combining with the insights from Proposition~\ref{prop:dominant_mode_proxy}, future designs should balance controlling the curvature-weighted variance to avoid robust overfitting while relaxing regularization strength in directions that capture robust features to enhance the fit of training data.

\subsection{Mechanistic Interpretation of Robust Overfitting}

Based on our time-resolved PAC-Bayesian robust generalization bounds and the corresponding empirical diagnostic curves, we can now summarize the underlying mechanisms of robust overfitting.
Prior to the first learning-rate drop in adversarial training, the system reaches a stationary state. Due to the stability condition, Hessian eigenvalues are regularized: $0 < \lambda_i < 2(1+\kappa)/\eta$.
However, as illustrated by Proposition~\ref{prop:robust optimization & curvature}, lowering adversarial loss requires stabilizing input sensitivity, suggesting the need to explore high-curvature regions to further improve optimization.

After $\eta$ sharply decays, the system begins to explore high-curvature directions to further reduce the adversarial training loss. Since the system has only evolved gradually at the initial steps, the Hessian and gradient noise covariance remain similar; however, the posterior will rapidly concentrate due to the sharp learning-rate drop ($\eta \rightarrow \eta/10$), effectively lowering the \textit{curvature-weighted variance} $\frac{1}{2} \sum_{i=1}^m \lambda_{t,i}\,\sigma_{t,i}^2 $. While contracted posterior fluctuation also increases the \textit{entropic KL-penalty} term $- \frac{1}{2\beta} \sum_{i=1}^m \ln \sigma_{t,i}^2$, it cannot offset the linear decrease in variance. Continual training with a small $\eta$ keeps increasing Hessian eigenvalues, eventually boosting the variance term and worsening robust generalization. The above mechanistic interpretation matches the empirical observation of robust overfitting well: test robust error first quickly drops after a learning rate decay, then continues to rise, whereas the training loss keeps decreasing.

\section{Conclusion}  
We derived time-resolved PAC-Bayesian robust generalization bounds by modeling momentum SGD as a discrete-time dynamical system and deriving closed-form posterior evolutions that depend on the learning rate, the loss geometry, and the stochastic gradient noise. Through theoretical analysis, empirical diagnostics, and controlled experiments, our work provides mechanistic insights into how learning dynamics lead to robust overfitting or generalization. Looking forward, important directions include extending our framework to more general settings, such as adaptive optimizers and relaxed assumptions, and leveraging the insights to enhance the training-time fit of AWP while avoiding overfitting.

\section*{Availability}

The code for implementing our spectral estimation algorithm and reproducing our main experimental results is available at: \href{https://github.com/TrustMLRG/RobustGen}{https://github.com/TrustMLRG/RobustGen}.

\bibliography{ref}

\onecolumn
\title{How Learning Dynamics Drive Adversarially Robust Generalization?\\(Supplementary Material)}
\maketitle

\appendix

\section{Proofs of Theoretical Results in Section \ref{sec:link to loss and posterior geometry}}

\subsection{Proof of Lemma~\ref{lemma:kl_divergence}}
\label{append:proof of lemma kl}

\begin{proof}[Proof of Lemma~\ref{lemma:kl_divergence}]
According to the definition of KL divergence, we have
\begin{align}
\label{eq:general KL divergence def}
    \mathrm{KL}\!\left(\mathcal{Q}\,\|\,\mathcal{P}\right)
    = \int q(\bm{w}) \,\ln \frac{q(\bm{w})}{p(\bm{w})}\,\mathrm{d}\bm{w}
    = \mathbb{E}_{\bm{w}\sim\mathcal{Q}}\!\left[\ln q(\bm{w}) - \ln p(\bm{w})\right],
\end{align}
where the density functions of $\mathcal{P}$ and $\mathcal{Q}$ under our assumptions can be written as:
\begin{align*}
    p(\bm{w}) &= (2\pi)^{-\frac{m}{2}}\,\sigma_{\mathcal{P}}^{-m}
    \cdot \exp\!\left(-\frac{1}{2\sigma_{\mathcal{P}}^{2}}\,\bm{w}^{\!\top}\bm{w}\right), \\
    q(\bm{w}) &= (2\pi)^{-\frac{m}{2}}\,\det(\mathbf{\Sigma}_{\mathcal{Q}})^{-\frac{1}{2}} \cdot \exp\!\left(-\frac{1}{2}(\bm{w}-\bm{\mu}_{\mathcal{Q}})^{\!\top}
    \mathbf{\Sigma}_{\mathcal{Q}}^{-1}(\bm{w}-\bm{\mu}_{\mathcal{Q}})\right).
\end{align*}
Taking logs of the above equations, we obtain
\begin{equation}
\label{eq:prior posterior density def}
    \begin{aligned}
        \ln p(\bm{w}) &= -\frac{m}{2}\ln(2\pi) - m\ln\sigma_{\mathcal{P}}
        -\frac{1}{2\sigma_{\mathcal{P}}^{2}}\,\bm{w}^{\!\top}\bm{w}, \\
        \ln q(\bm{w}) &= -\frac{m}{2}\ln(2\pi) - \frac{1}{2}\ln\det(\mathbf{\Sigma}_{\mathcal{Q}})
        -\frac{1}{2}(\bm{w}-\bm{\mu}_{\mathcal{Q}})^{\!\top}\mathbf{\Sigma}_{\mathcal{Q}}^{-1}(\bm{w}-\bm{\mu}_{\mathcal{Q}}).
    \end{aligned}
\end{equation}
Plugging \Eqref{eq:prior posterior density def} into \Eqref{eq:general KL divergence def}, we further obtain
\begin{align*}
\mathrm{KL}\!\left(\mathcal{Q}\,\|\,\mathcal{P}\right)
&= \mathbb{E}_{\bm{w}\sim\mathcal{Q}}\!\Bigg[
-\frac{1}{2}\ln\det(\mathbf{\Sigma}_{\mathcal{Q}})
-\frac{1}{2}(\bm{w}-\bm{\mu}_{\mathcal{Q}})^{\!\top}\mathbf{\Sigma}_{\mathcal{Q}}^{-1}(\bm{w}-\bm{\mu}_{\mathcal{Q}})
+ m\ln\sigma_{\mathcal{P}}
+ \frac{1}{2\sigma_{\mathcal{P}}^{2}}\,\bm{w}^{\!\top}\bm{w}
\Bigg] \\
& = -\frac{1}{2}\ln\det(\mathbf{\Sigma}_{\mathcal{Q}})
-\frac{m}{2}
+ m\ln\sigma_{\mathcal{P}}
+\frac{1}{2\sigma_{\mathcal{P}}^{2}}\!\left(\mathrm{Tr}(\mathbf{\Sigma}_{\mathcal{Q}}) + \|\bm{\mu}_{\mathcal{Q}}\|_{2}^{2}\right),
\end{align*}
where the last equality holds because $\mathcal{Q} = \mathcal{N}(\bm{\mu}_{\mathcal{Q}}, \mathbf{\Sigma}_{\mathcal{Q}})$.
Specifically, for any $\bm{w}\sim\mathcal{Q}$, $\mathbf{\Sigma}^{-1/2}(\bm{w} - \bm{\mu}_{\mathcal{Q}}) \sim \mathcal{N}(\bm{0}, \mathbf{I})$, which suggests that $\mathbb{E}_{\mathcal{Q}}\!\left[(\bm{w}-\bm{\mu}_{\mathcal{Q}})^{\!\top}\mathbf{\Sigma}_{\mathcal{Q}}^{-1}(\bm{w}-\bm{\mu}_{\mathcal{Q}})\right] = m$, and $\mathbb{E}_{\mathcal{Q}}\!\left[\bm{w}^{\!\top}\bm{w}\right]
= \mathbb{E}_{\mathcal{Q}}\!\left[\|\bm{w}-\bm{\mu}_{\mathcal{Q}}\|_2^2\right] + \|\bm{\mu}_{\mathcal{Q}}\|_2^2 = \mathrm{Tr}(\mathbf{\Sigma_{\mathcal{Q}}}) + \|\bm{\mu}_{\mathcal{Q}}\|_2^2$. 
Therefore, we complete the proof of Lemma~\ref{lemma:kl_divergence}.
\end{proof}

\subsection{Proof of Lemma~\ref{lemma:expected empirical adversarial loss}}
\label{sec:proof expected empirical adversarial loss}

\begin{proof}[Proof of Lemma~\ref{lemma:expected empirical adversarial loss}]
Under the local quadratic upper-bound assumption in Lemma~\ref{lemma:expected empirical adversarial loss}, the empirical adversarial loss satisfies the following Taylor upper expansion around the reference point $\tilde{\bm{w}}$:
\begin{equation}
\label{eq:quad-form}
\hat{\mathcal{L}}_{\epsilon}(\bm{w})
\le
\hat{\mathcal{L}}_{\epsilon}(\tilde{\bm{w}})
+ \nabla \hat{\mathcal{L}}_{\epsilon}(\tilde{\bm{w}})^\top (\bm{w} - \tilde{\bm{w}})
+\frac{1}{2}\,(\bm{w}-\tilde{\bm{w}})^{\!\top}\hat{\mathbf{H}}_{\epsilon}(\tilde{\bm{w}})(\bm{w}-\tilde{\bm{w}})
+r_{\tilde{\bm{w}}}(\bm{w}), \text{ for any } \bm{w}\sim\mathcal{Q},
\end{equation}
where $\nabla \hat{\mathcal{L}}_{\epsilon}(\tilde{\bm{w}})$ (resp. $\hat{\mathbf{H}}_{\epsilon}(\tilde{\bm{w}})$) denotes the gradient (resp. Hessian) of the empirical adversarial loss at the reference point.

Taking expectation over $\bm{w}\sim\mathcal{Q}=\mathcal{N}(\bm{\mu}_{\mathcal{Q}},\mathbf{\Sigma}_{\mathcal{Q}})$, we have
\begin{align}
\label{eq:second order taylor expansion Lemma 4.2}
    \mathbb{E}_{\bm{w}\sim\mathcal{Q}}\!\left[\hat{\mathcal{L}}_{\epsilon}(\bm{w})\right]
    \le\;&\hat{\mathcal{L}}_{\epsilon}(\tilde{\bm{w}})
    + \nabla \hat{\mathcal{L}}_{\epsilon}(\tilde{\bm{w}})^\top
    \mathbb{E}_{\bm{w}\sim\mathcal{Q}}[(\bm{w} - \tilde{\bm{w}})] \nonumber\\
    &+\frac{1}{2}\,\mathbb{E}_{\bm{w}\sim\mathcal{Q}}\!\left[(\bm{w}-\tilde{\bm{w}})^{\!\top}\hat{\mathbf{H}}_{\epsilon}(\tilde{\bm{w}})(\bm{w}-\tilde{\bm{w}})\right]
    + \mathbb{E}_{\bm{w}\sim\mathcal{Q}}[r_{\tilde{\bm{w}}}(\bm{w})].
\end{align}
Let $\Delta \bar{\bm{w}} := \bm{\mu}_{\mathcal{Q}} - \tilde{\bm{w}}$ denote the posterior mean shift. Since $\mathbb{E}_{\bm{w}\sim\mathcal{Q}}[\bm{w}] = \bm{\mu}_{\mathcal{Q}}$, we have
\begin{align}
\label{eq:linear term Lemma 4.2}
    \mathbb{E}_{\bm{w}\sim\mathcal{Q}}[(\bm{w} - \tilde{\bm{w}})] = \bm{\mu}_{\mathcal{Q}} - \tilde{\bm{w}} = \Delta \bar{\bm{w}}.
\end{align}
The remaining task is to simplify the quadratic term in \Eqref{eq:second order taylor expansion Lemma 4.2}. From the second-moment identity, we know
\begin{align}
\label{eq:basic quadratic term Lemma 4.2}
    \mathbb{E}_{\bm{w}\sim\mathcal{Q}}\!\left[(\bm{w}-\tilde{\bm{w}})(\bm{w}-\tilde{\bm{w}})^{\!\top}\right]
    =\mathbf{\Sigma}_{\mathcal{Q}}+\Delta \bar{\bm{w}}\,{\Delta \bar{\bm{w}}}^{\!\top},
\end{align}
which holds for any Gaussian with mean $\bm{\mu}_{\mathcal{Q}}$ and covariance $\mathbf{\Sigma}_{\mathcal{Q}}$. Hence, we have
\begin{align}
\label{eq:quadratic term Lemma 4.2}
    \nonumber \mathbb{E}_{\bm{w}\sim\mathcal{Q}}\!\left[(\bm{w}-\tilde{\bm{w}})^{\!\top}\hat{\mathbf{H}}_{\epsilon}(\tilde{\bm{w}})(\bm{w}-\tilde{\bm{w}})\right]
    &= \mathbb{E}_{\bm{w}\sim\mathcal{Q}} \left[  \mathrm{Tr}\!\left( (\bm{w}-\tilde{\bm{w}})^{\top} \hat{\mathbf{H}}_{\epsilon}(\tilde{\bm{w}}) (\bm{w}-\tilde{\bm{w}}) \right) \right] \\
    \nonumber &= \mathrm{Tr}\!\left(\hat{\mathbf{H}}_{\epsilon}(\tilde{\bm{w}})\,\mathbb{E}\!\left[(\bm{w}-\tilde{\bm{w}})(\bm{w}-\tilde{\bm{w}})^{\!\top}\right]\right) \\
    \nonumber &= \mathrm{Tr}\!\left(\hat{\mathbf{H}}_{\epsilon}(\tilde{\bm{w}})\mathbf{\Sigma}_{\mathcal{Q}}\right)
    + \mathrm{Tr}\!\left(\hat{\mathbf{H}}_{\epsilon}(\tilde{\bm{w}})\Delta \bar{\bm{w}}\,{\Delta \bar{\bm{w}}}^{\!\top}\right) \\
    &= \mathrm{Tr}\!\left(\hat{\mathbf{H}}_{\epsilon}(\tilde{\bm{w}})\mathbf{\Sigma}_{\mathcal{Q}}\right)
    + {\Delta \bar{\bm{w}}}^{\!\top}\hat{\mathbf{H}}_{\epsilon}(\tilde{\bm{w}})\Delta \bar{\bm{w}},
\end{align}
where the second and the last equalities hold because of the cyclic property of trace: $\mathrm{Tr}(\mathbf{A}\mathbf{u}\mathbf{u}^{\!\top}) = \mathrm{Tr}(\mathbf{u}^{\!\top}\mathbf{A}\mathbf{u})$ holds for any matrix $\mathbf{A}$ and vector $\bm{u}$, and the third equality is due to \Eqref{eq:basic quadratic term Lemma 4.2}.

Plugging \Eqref{eq:linear term Lemma 4.2} and \Eqref{eq:quadratic term Lemma 4.2} into \Eqref{eq:second order taylor expansion Lemma 4.2} yields
\[
\mathbb{E}_{\bm{w}\sim\mathcal{Q}} \big[ \hat{\mathcal{L}}_{\epsilon}(\bm{w}) \big]
\le  \hat{\mathcal{L}}_{\epsilon}(\tilde{\bm{w}})
+ \nabla \hat{\mathcal{L}}_{\epsilon}(\tilde{\bm{w}})^\top \Delta \bar{\bm{w}}
+ \frac{1}{2} {\Delta \bar{\bm{w}}}^{\top} \hat{\mathbf{H}}_{\epsilon}(\tilde{\bm{w}}) \Delta \bar{\bm{w}}
+ \frac{1}{2} \mathrm{Tr}(\hat{\mathbf{H}}_{\epsilon}(\tilde{\bm{w}}) \mathbf{\Sigma}_{\mathcal{Q}})
+ R_{\mathcal{Q}}(\tilde{\bm{w}}),
\]
where $R_{\mathcal{Q}}(\tilde{\bm{w}})=\mathbb{E}_{\bm{w}\sim\mathcal{Q}}[r_{\tilde{\bm{w}}}(\bm{w})]$.
This completes the proof of Lemma~\ref{lemma:expected empirical adversarial loss}.
\end{proof}

\subsection{Proof of Theorem~\ref{coro:compact_pac_bound}}
\label{append:proof_compact_pac_bound}
\begin{proof}[Proof of Theorem~\ref{coro:compact_pac_bound}]
We start from the generic PAC-Bayesian robust generalization bound in
Theorem~\ref{thm:generic advbound}:
\begin{align}
\label{eq:appendix_compact_bound_start}
        \mathbb{E}_{\bm{w} \sim \mathcal{Q}} \big[ \mathcal{L}_{\epsilon}(\bm{w}) \big]
        \le
        \mathbb{E}_{\bm{w} \sim \mathcal{Q}} \big[ \hat{\mathcal{L}}_{\epsilon}(\bm{w}) \big]
        + \frac{1}{\beta} \mathrm{KL}(\mathcal{Q} \,\|\, \mathcal{P})
        + \frac{\beta C_\ell^{2}}{8 |\mathcal{S}|}
        - \frac{1}{\beta} \ln{\alpha}.
\end{align}

\emph{Step 1: Expected empirical adversarial loss under a Gaussian posterior.}
Under the local quadratic upper bound of Lemma~\ref{lemma:expected empirical adversarial loss}, we have
\begin{align}
\label{eq:appendix_compact_empirical}
    \mathbb{E}_{\bm{w}\sim\mathcal{Q}} \big[\hat{\mathcal{L}}_{\epsilon}(\bm{w})\big]
    &\le \hat{\mathcal{L}}_{\epsilon}(\tilde{\bm{w}})
    + \big\langle \nabla \hat{\mathcal{L}}_{\epsilon}(\tilde{\bm{w}}), \Delta \bar{\bm{w}} \big\rangle
    + \frac{1}{2} {\Delta \bar{\bm{w}}}^{\top} \hat{\mathbf{H}}_{\epsilon} (\tilde{\bm{w}}) \Delta \bar{\bm{w}}
    + \frac{1}{2} \mathrm{Tr} \big( \hat{\mathbf{H}}_{\epsilon}(\tilde{\bm{w}}) \mathbf{\Sigma}_{\mathcal{Q}} \big)
    +R_{\mathcal{Q}}(\tilde{\bm{w}}),
\end{align}
where $\Delta \bar{\bm{w}}=\bm{\mu}_{\mathcal{Q}}-\tilde{\bm{w}}$.

\emph{Step 2: Closed-form KL divergence.}
For the spherical Gaussian prior $\mathcal{P}=\mathcal{N}(\bm{0},\sigma_{\mathcal{P}}^{2}\mathbf I)$ and
Gaussian posterior $\mathcal{Q}=\mathcal{N}(\bm{\mu}_{\mathcal{Q}},\mathbf{\Sigma}_{\mathcal{Q}})$,
Lemma~\ref{lemma:kl_divergence} yields
\begin{align}
\label{eq:appendix_compact_kl}
    \frac{1}{\beta}\mathrm{KL}(\mathcal{Q}\,\|\,\mathcal{P})
    &=
    \underbrace{
        \frac{1}{2\beta \sigma^{2}_{\mathcal{P}}}
        \Big(\| \bm{\mu}_{\mathcal{Q}} \|_{2}^{2} + \mathrm{Tr}(\mathbf{\Sigma}_{\mathcal{Q}})\Big)
        - \frac{1}{2\beta} \ln \det (\mathbf{\Sigma}_{\mathcal{Q}})
    }_{\text{terms depending on }\mathcal{Q}}
    + \underbrace{\frac{m}{2\beta}\ln(\sigma^2_{\mathcal{P}}) - \frac{m}{2\beta}}_{\text{constant in }\mathcal{Q}} .
\end{align}

\emph{Step 3: Combine terms.}
Substituting \eqref{eq:appendix_compact_empirical} and \eqref{eq:appendix_compact_kl} into
\eqref{eq:appendix_compact_bound_start}, and grouping all terms independent of the posterior
$\mathcal Q$ into
\begin{align}
\label{eq:Pac Bayes remainder term}
     C_{\mathrm{PB}} = C_{\mathrm{PB}}(\sigma^2_{\mathcal{P}}, \mathcal{S}, \alpha, \beta)
    =
    \frac{m}{2\beta}\ln\sigma^2_{\mathcal{P}}
    - \frac{m}{2\beta}
    + \frac{\beta C_\ell^2}{8|\mathcal{S}|}
    - \frac{1}{\beta}\ln\alpha,
\end{align}
where we obtain exactly the bound in \Eqref{eq:compact_pac_bound}, including the local Taylor upper remainder.
\end{proof}

\subsection{Robust Generalization Bounds under Gaussian-Mixture Posterior}
\label{append:proof of corollary mixture pac bound}

\begin{corollary}[Robust Generalization with Gaussian Mixtures]
\label{coro:mixture_pac_bound}
Let $\mathcal{P}=\mathcal{N}(0,\sigma_{\mathcal{P}}^{2}\mathbf{I})$ be the prior, and let the posterior be a mixture of Gaussians:
\begin{align}
\label{eq:mixture Gaussian posterior}
    \mathcal{Q}
    &= \sum_{\ell=1}^{L} \pi_{\ell}\, \mathcal{Q}_{\ell}, \quad \text{where } \:
    \mathcal{Q}_{\ell} = \mathcal{N}(\bm{\mu}_{\ell}, \mathbf{\Sigma}_{\ell}), \: \sum_{\ell=1}^{L}\pi_{\ell}=1, \text{ and } \pi_{\ell}\ge0.
\end{align}
Assume the generalized local quadratic upper bound (Lemma~\ref{lemma:expected empirical adversarial loss}) holds for each component $\ell$ around a reference point $\bm{w}_{\text{ref},\ell}$ with Hessian $\mathbf{H}_{\ell}$, gradient $\bm{g}_{\ell} := \nabla \hat{\mathcal{L}}_{\epsilon}(\bm{w}_{\text{ref},\ell})$, and remainder $R_{\mathcal{Q}_{\ell}}(\bm{w}_{\text{ref},\ell})$.
For any $\beta>0$ and $\alpha\in(0,1)$, with probability at least $1-\alpha$, we have
\begin{align}
\mathbb{E}_{\bm{w}\sim\mathcal{Q}}[\mathcal{L}_{\epsilon}(\bm{w})]
& \;\leq\;
\sum_{\ell=1}^{L} \pi_{\ell}
\big[
	    \hat{\mathcal{L}}_{\epsilon}(\bm{w}_{\text{ref},\ell})
	    + \underbrace{\big\langle
        \bm{g}_{\ell},\, \bm{\mu}_{\ell}-\bm{w}_{\text{ref},\ell}
        \big\rangle}_{\text{first-order bias}} + \underbrace{\frac{1}{2}
        (\bm{\mu}_{\ell}-\bm{w}_{\text{ref},\ell})^\top
        \mathbf{H}_{\ell}
        (\bm{\mu}_{\ell}-\bm{w}_{\text{ref},\ell})}_{\text{second-order bias}}
		    + \underbrace{\frac{1}{2}
        \mathrm{Tr}\!\left(\mathbf{H}_{\ell}\mathbf{\Sigma}_{\ell}\right)}_{\text{variance}} 
		    + R_{\mathcal{Q}_{\ell}}(\bm{w}_{\text{ref},\ell})
		\big] \nonumber \\
	&\qquad + 
	\sum_{\ell=1}^{L}
	\frac{\pi_{\ell}}{2\beta}
	\left(
	        \frac{\mathrm{Tr}(\mathbf{\Sigma}_{\ell})}{\sigma_{\mathcal{P}}^{2}}
	        + \frac{\|\bm{\mu}_{\ell}\|_{2}^{2}}{\sigma_{\mathcal{P}}^{2}}
            - m
	        + m\ln\sigma_{\mathcal{P}}^{2}
	        - \ln\det\mathbf{\Sigma}_{\ell}
	\right)
+\;
\frac{\beta C_\ell^{2}}{8|\mathcal{S}|}
-\frac{1}{\beta}\ln\alpha.
\end{align}
Here $C_\ell$ denotes the global scalar loss-bound constant, not the mixture-component index.
When $\mathbf{H}_{\ell}\succ 0$, the sum of the first- and second-order bias terms can be written by completing the square, and is minimized at the Newton-corrected mean $\bm{\mu}_{\ell}=\bm{w}_{\text{ref},\ell}-\mathbf{H}_{\ell}^{-1}\bm{g}_{\ell}$, recovering the familiar optimization gain $-\tfrac{1}{2}\bm{g}_{\ell}^\top\mathbf{H}_{\ell}^{-1}\bm{g}_{\ell}$.
\end{corollary}

\begin{proof}[Proof of Corollary~\ref{coro:mixture_pac_bound}]
We start with the most general PAC-Bayesian bound in Theorem~\ref{thm:generic advbound},
since it holds for any posterior~$\mathcal{Q}$.
For any $\beta>0$ and any $\alpha\in(0,1)$, with probability at least $1-\alpha$
over the finite sample set $\mathcal{S}$, we have
\begin{align}
\label{eq:appendix_mixture_catoni_start}
    \mathbb{E}_{\bm{w}\sim\mathcal{Q}}[\mathcal{L}_{\epsilon}(\bm{w})]
    \leq
    \mathbb{E}_{\bm{w}\sim\mathcal{Q}}
    \bigg[
        \frac{1}{|\mathcal{S}|}
        \sum_{(\bm{x},y)\in\mathcal{S}}
        \ell_{\mathrm{adv}}(\bm{w},\bm{x},y)
    \bigg]
    + \frac{1}{\beta}\mathrm{KL}(\mathcal{Q}\,\|\,\mathcal{P})
    + \frac{\beta C_\ell^{2}}{8|\mathcal{S}|}
    - \frac{1}{\beta}\ln\alpha .
\end{align}
We now instantiate this bound by choosing the posterior to be a mixture of Gaussian distributions specified in \Eqref{eq:mixture Gaussian posterior} and the prior $\mathcal{P}=\mathcal{N}(0,\sigma_{\mathcal{P}}^{2}\mathbf{I})$.

\shortsection{Step 1: Decomposition of the empirical adversarial loss}
By the definition of expectations under mixture distributions, for any measurable
function $f:\mathcal{W}\to\mathbb{R}$,
\begin{align}
\label{eq:appendix_mixture_expectation_identity}
\mathbb{E}_{\bm{w}\sim\mathcal{Q}}[f(\bm{w})]
=
\sum_{\ell=1}^{L}
\pi_{\ell}
\,
\mathbb{E}_{\bm{w}\sim\mathcal{Q}_{\ell}}[f(\bm{w})].
\end{align} 
Applying \Eqref{eq:appendix_mixture_expectation_identity} to
\( f(\bm{w}) = \hat{\mathcal{L}}_{\epsilon}(\bm{w}) \) gives
\begin{align}
\label{eq:appendix_mixture_empirical_decomp}
\mathbb{E}_{\bm{w}\sim\mathcal{Q}}
\big[\hat{\mathcal{L}}_{\epsilon}(\bm{w})\big]
=
\sum_{\ell=1}^{L}
\pi_{\ell}\,
\mathbb{E}_{\bm{w}\sim\mathcal{Q}_{\ell}}
\big[\hat{\mathcal{L}}_{\epsilon}(\bm{w})\big].
\end{align}
For each Gaussian component $\mathcal{Q}_{\ell}$, note that we assume $\hat{\mathcal{L}}_{\epsilon}(\bm{w})$ can be
locally upper bounded around the basin-specific reference point $\bm{w}_{\text{ref},\ell}$
by a quadratic form with Hessian $\mathbf{H}_{\ell}$ and gradient $\bm{g}_{\ell}:=\nabla \hat{\mathcal{L}}_{\epsilon}(\bm{w}_{\text{ref},\ell})$.  
Thus, applying Lemma~\ref{lemma:expected empirical adversarial loss} to each component, we obtain
\begin{align}
\label{eq:appendix_mixture_empirical_component}
\mathbb{E}_{\bm{w}\sim\mathcal{Q}_{\ell}}
\big[\hat{\mathcal{L}}_{\epsilon}(\bm{w})\big]
\le
\hat{\mathcal{L}}_{\epsilon}(\bm{w}_{\text{ref},\ell})
+
\bm{g}_{\ell}^\top(\bm{\mu}_{\ell}-\bm{w}_{\text{ref},\ell})
+
\frac{1}{2}
(\bm{\mu}_{\ell}-\bm{w}_{\text{ref},\ell})^\top
\mathbf{H}_{\ell}
(\bm{\mu}_{\ell}-\bm{w}_{\text{ref},\ell})
+
\frac{1}{2}
\mathrm{Tr}(\mathbf{H}_{\ell}\mathbf{\Sigma}_{\ell})
 + R_{\mathcal{Q}_{\ell}}(\bm{w}_{\text{ref},\ell}).
\end{align}

Substituting \Eqref{eq:appendix_mixture_empirical_component} into
\Eqref{eq:appendix_mixture_empirical_decomp} yields the mixture-expanded empirical loss:
\begin{align}
\label{eq:appendix_mixture_empirical_final}
\nonumber \mathbb{E}_{\bm{w}\sim\mathcal{Q}}
[\hat{\mathcal{L}}_{\epsilon}(\bm{w})]
&\le
\sum_{\ell=1}^{L}
\pi_{\ell}
\Big[
\hat{\mathcal{L}}_{\epsilon}(\bm{w}_{\text{ref},\ell})
+\bm{g}_{\ell}^\top(\bm{\mu}_{\ell}-\bm{w}_{\text{ref},\ell}) 
\\ 
&\qquad +\frac{1}{2}(\bm{\mu}_{\ell}-\bm{w}_{\text{ref},\ell})^\top
\mathbf{H}_{\ell}
(\bm{\mu}_{\ell}-\bm{w}_{\text{ref},\ell}) 
+\frac{1}{2}\mathrm{Tr}(\mathbf{H}_{\ell}\mathbf{\Sigma}_{\ell})
+R_{\mathcal{Q}_{\ell}}(\bm{w}_{\text{ref},\ell})
\Big].
\end{align}

\shortsection{Step 2: KL divergence upper bound for the mixture posterior}
Since KL divergence is convex in its first argument, the mixture posterior
$\mathcal{Q}=\sum_{\ell=1}^{L}\pi_{\ell}\mathcal{Q}_{\ell}$ satisfies
\begin{align}
\label{eq:appendix_mixture_kl_convexity}
\mathrm{KL}(\mathcal{Q}\,\|\,\mathcal{P})
=
\mathrm{KL}\Big(\sum_{\ell=1}^{L}\pi_{\ell}\mathcal{Q}_{\ell}\,\Big\|\,\mathcal{P}\Big)
\leq
\sum_{\ell=1}^{L}
\pi_{\ell}\,
\mathrm{KL}(\mathcal{Q}_{\ell}\,\|\,\mathcal{P}),
\end{align}
where the inequality follows from the definition KL divergence and the log sum inequality.
For each $\mathcal{Q}_{\ell}=\mathcal{N}(\bm{\mu}_{\ell},\mathbf{\Sigma}_{\ell})$,
Lemma~\ref{lemma:kl_divergence} provides the closed-form expression:
\begin{align}
\label{eq:appendix_mixture_kl_component}
\mathrm{KL}(\mathcal{Q}_{\ell}\,\|\,\mathcal{P})
=
\frac{\mathrm{Tr}(\mathbf{\Sigma}_{\ell})}{2\sigma_{\mathcal{P}}^{2}}
+\frac{\|\bm{\mu}_{\ell}\|_{2}^{2}}{2\sigma_{\mathcal{P}}^{2}}
-\frac{m}{2}
+\frac{m}{2}\ln\sigma_{\mathcal{P}}^{2}
-\frac{1}{2}\ln\det\mathbf{\Sigma}_{\ell}.
\end{align}

Multiplying \Eqref{eq:appendix_mixture_kl_convexity} by $1/\beta$ and substituting
\Eqref{eq:appendix_mixture_kl_component} yields
\begin{align}
\label{eq:appendix_mixture_kl_final}
\frac{1}{\beta}\mathrm{KL}(\mathcal{Q}\,\|\,\mathcal{P})
\leq
\sum_{\ell=1}^{L}
\frac{\pi_{\ell}}{2\beta}
\left(
\frac{\mathrm{Tr}(\mathbf{\Sigma}_{\ell})}{\sigma_{\mathcal{P}}^{2}}
+\frac{\|\bm{\mu}_{\ell}\|_{2}^{2}}{\sigma_{\mathcal{P}}^{2}}
-m
+m\ln\sigma_{\mathcal{P}}^{2}
-\ln\det\mathbf{\Sigma}_{\ell}
\right).
\end{align}

Finally, we substitute the empirical-loss expansion
(\Eqref{eq:appendix_mixture_empirical_final})
and the KL bound (\Eqref{eq:appendix_mixture_kl_final})
into \Eqref{eq:appendix_mixture_catoni_start}.
Noticing that the remaining terms, where $C_\ell$ denotes the global scalar loss-bound constant,
$\frac{\beta C_\ell^{2}}{8|\mathcal{S}|}$ and $-\frac{1}{\beta}\ln(\alpha)$ do not
depend on the mixture-component index~$\ell$, we obtain
\[
\mathbb{E}_{\bm{w}\sim\mathcal{Q}}[\mathcal{L}_{\epsilon}(\bm{w})]
\leq
\sum_{\ell=1}^{L}
\pi_{\ell}
\Big[
\hat{\mathcal{L}}_{\epsilon}(\bm{w}_{\text{ref},\ell})
+\bm{g}_{\ell}^\top(\bm{\mu}_{\ell}-\bm{w}_{\text{ref},\ell})
+\frac{1}{2}(\bm{\mu}_{\ell}-\bm{w}_{\text{ref},\ell})^\top
\mathbf{H}_{\ell}(\bm{\mu}_{\ell}-\bm{w}_{\text{ref},\ell})
+\frac{1}{2}\mathrm{Tr}(\mathbf{H}_{\ell}\mathbf{\Sigma}_{\ell})
+R_{\mathcal{Q}_{\ell}}(\bm{w}_{\text{ref},\ell})
\Big]
\]
\[
\hphantom{\mathbb{E}_{\bm{w}\sim\mathcal{Q}}[\mathcal{L}_{\epsilon}(\bm{w})]}
+\;
\sum_{\ell=1}^{L}
\frac{\pi_{\ell}}{2\beta}
\left(
\frac{\mathrm{Tr}(\mathbf{\Sigma}_{\ell})}{\sigma_{\mathcal{P}}^{2}}
+\frac{\|\bm{\mu}_{\ell}\|_{2}^{2}}{\sigma_{\mathcal{P}}^{2}}
-m
+m\ln\sigma_{\mathcal{P}}^{2}
-\ln\det\mathbf{\Sigma}_{\ell}
\right)
+\frac{\beta C_\ell^{2}}{8|\mathcal{S}|}
-\frac{1}{\beta}\ln\alpha.
\]

This expression gives the stated upper bound in
Corollary~\ref{coro:mixture_pac_bound}, completing the proof.
\end{proof}

\section{Proofs of Theoretical Results in Section~\ref{sec:posterior dynamics analysis}}

\subsection{Stationary Regime: Proof of Lemma~\ref{lemma:cov_param_projection_detailed}}
\label{append:proof_cov_param_projection_detailed}

\begin{proof}[Proof of Lemma~\ref{lemma:cov_param_projection_detailed}]
Under the generalized quadratic loss approximation, the stochastic gradient includes a constant drift $\bm{g}_{\text{ref}}:= \nabla \hat{\mathcal{L}}_{\epsilon}(\bm{w}_{\text{ref}})$.
For covariance calculations, we can omit this term without loss of generality, because covariance measures \emph{centered} fluctuations: letting $\tilde{\bm{u}}_t = \bm{u}_t - \mathbb{E}[\bm{u}_t]$ denote the centered state, the drift appears identically in both $\bm{u}_t$ and $\mathbb{E}[\bm{u}_t]$ and cancels upon subtraction.
This cancellation holds regardless of whether the system is stationary or transient (the multi-step transient recursion is handled in the proof of Lemma~\ref{lemma:windowed_mean_covariance_recursion}).

Throughout this proof, we treat the training set $\mathcal{S}$ (and hence the empirical loss $\hat{\mathcal{L}}_{\epsilon}$) as fixed, and take all expectations/covariances with respect to the algorithmic randomness conditional on $\mathcal{S}$.

\emph{State-space form.} To begin with, we prove Equation \ref{eq:posterior covariance iterates}.
From the updates in \Eqref{eq:sgd_update_momentum_prelim}, we have
\[
\begin{aligned}
\bm{v}_{t+1}
&= \kappa\,\bm{v}_{t} + \bm{g}_{\text{ref}} + \mathbf H_{\text{ref}}(\bm{w}_{t}-\bm{w}_{\text{ref}}) + \bm{\xi}_{t},\\
\bm{w}_{t+1}-\bm{w}_{\text{ref}}
&= (\bm{w}_{t}-\bm{w}_{\text{ref}}) - \eta \bm{v}_{t+1}
= (\mathbf I - \eta \mathbf H_{\text{ref}})(\bm{w}_{t}-\bm{w}_{\text{ref}}) - \eta \kappa\,\bm{v}_{t} - \eta\,\bm{g}_{\text{ref}} - \eta\,\bm{\xi}_{t}.
\end{aligned}
\]
Stacking the two lines with the joint state
\(\bm{u}_t:=\begin{bmatrix}\bm{w}_t-\bm{w}_{\text{ref}}\\ \bm{v}_t\end{bmatrix}\in\mathbb R^{2m}\), we obtain the affine state-space system
\[
\bm{u}_{t+1} \;=\; \mathbf A\,\bm{u}_{t} + \mathbf G\,\big[\bm{g}_{\text{ref}}+\bm{\xi}_{t}\big],
\quad
\mathbf A =
\begin{bmatrix}
\mathbf I - \eta \mathbf H_{\text{ref}} & -\eta\kappa\,\mathbf I \\
\mathbf H_{\text{ref}} & \kappa\,\mathbf I
\end{bmatrix},
\quad
\mathbf G=\begin{bmatrix}-\eta\mathbf I\\ \mathbf I\end{bmatrix},
\]
which is exactly \eqref{eq:posterior covariance iterates}.
\end{proof}

\shortersection{Stationary Mean \& Covariance} Next, we derive the stationary posterior mean and covariance as in Equations \ref{eq:stationary posterior mean} and \ref{eq:stationary posterior covariance}.

\begin{remark}[Commutative Structure and Spectral Decomposition]
\label{rem:commute diagonal alignment}
When $\mathbf{C}$ commutes with $\mathbf{H}_{\text{ref}}$, they share the same eigenbasis. Let $\{\lambda_1,\ldots,\lambda_m\}$ and $\{\gamma_1,\ldots,\gamma_m\}$ denote the eigenvalues of $\mathbf{H}_{\text{ref}}$ and $\mathbf{C}$, respectively. As shown in the derivation of \Eqref{eq:stationary posterior covariance}, the eigenvalues of the stationary covariance $\mathbf{\Sigma}$ are:
\begin{align}
\label{eq:eigenvalues stationary covariance}
    \forall i \in \{1,\ldots,m\}, \quad \sigma_i^2 = \frac{\eta}{1-\kappa}\cdot\frac{\gamma_i}{\lambda_i \left( 2-\frac{\eta}{1+\kappa}\lambda_i \right)}.
\end{align}
The stability condition requires $0<\lambda_i<\frac{2(1+\kappa)}{\eta}$ for all $i$. Since both $\mathbf{H}_{\text{ref}}$ and $\mathbf{C}$ depend on the perturbation strength $\epsilon$, altering $\epsilon$ correspondingly influences the stationary covariance structure.
\end{remark}

\begin{proof}[Derivation of \Eqref{eq:stationary posterior mean}]
	    Under the local quadratic approximation in Lemma~\ref{lemma:expected empirical adversarial loss}, the gradient of the empirical adversarial loss at any point $\bm{w}$ can be written as:
	    \[
	        \nabla_{\bm{w}} \hat{\mathcal{L}}_{\epsilon}(\bm{w}) = \nabla \hat{\mathcal{L}}_{\epsilon}(\bm{w}_{\text{ref}}) + \mathbf{H}_{\text{ref}} (\bm{w} - \bm{w}_{\text{ref}}),
	    \]
	    where $\nabla \hat{\mathcal{L}}_{\epsilon}(\bm{w}_{\text{ref}}) =: \bm{g}_{\text{ref}}$ is the (possibly non-zero) gradient at the reference point $\bm{w}_{\text{ref}}$.
	    
	    Taking expectation over the posterior $\mathcal{Q} = \mathcal{N}(\bm{\mu}_{\mathcal{Q}}, \mathbf{\Sigma}_{\mathcal{Q}})$, we obtain
	    \begin{align}
	    \label{eq:expectation over gradient}
	        \mathbb{E}_{\bm{w}\sim\mathcal{Q}} \big[ \nabla_{\bm{w}} \hat{\mathcal{L}}_{\epsilon}(\bm{w}) \big] 
	        = \bm{g}_{\text{ref}} + \mathbf{H}_{\text{ref}} (\bm{\mu}_{\mathcal{Q}} - \bm{w}_{\text{ref}}). 
	    \end{align}
	    
	    Since the posterior $\mathcal{Q}$ is stationary and obtained via SGD on $\hat{\mathcal{L}}_{\epsilon}(\bm{w})$, the expected gradient must vanish at stationarity. Otherwise, running another step of SGD would shift the mean and break the stationary assumption. Thus,
	    \[
	        \mathbb{E}_{\bm{w}\sim\mathcal{Q}} \big[ \nabla_{\bm{w}} \hat{\mathcal{L}}_{\epsilon}(\bm{w}) \big] = \bm{0}.
	    \]
    
    Combining with \Eqref{eq:expectation over gradient}, we have
    \[
        \bm{g}_{\text{ref}} + \mathbf{H}_{\text{ref}} (\bm{\mu}_{\mathcal{Q}} - \bm{w}_{\text{ref}}) = \bm{0}.
    \]
    
	    By the positive-definiteness assumption in Lemma~\ref{lemma:cov_param_projection_detailed}, $\mathbf{H}_{\text{ref}}\succ0$, and hence $\mathbf{H}_{\text{ref}}$ is invertible. Solving for $\bm{\mu}_{\mathcal{Q}}$ yields
	    \[
	        \bm{\mu}_{\mathcal{Q}} = \bm{w}_{\text{ref}} - \mathbf{H}_{\text{ref}}^{-1} \nabla \hat{\mathcal{L}}_{\epsilon}(\bm{w}_{\text{ref}}),
	    \]
	    which matches \Eqref{eq:stationary posterior mean} after identifying $\bm{w}_{\text{ref}}$ with $\tilde{\bm{w}}$ and $\mathbf{H}_{\text{ref}}$ with $\hat{\mathbf{H}}_{\epsilon}(\tilde{\bm{w}})$. This stationary mean relation has the form of a Newton-type correction around the reference point.
\end{proof}

\begin{proof}[Derivation of \Eqref{eq:stationary posterior covariance}]
\emph{Joint Lyapunov equation and projection.}
From Lemma~\ref{lemma:cov_param_projection_detailed}, the stacked state
\(
\bm{u}_t=\begin{bmatrix}\bm{w}_t-\bm{w}_{\text{ref}}\\ \bm{v}_t\end{bmatrix}\in\mathbb{R}^{2m}
\)
obeys the affine recursion
\[
\bm{u}_{t+1}=\mathbf A\,\bm{u}_{t}+\mathbf G\big(\bm{g}_{\text{ref}}+\bm{\xi}_{t}\big),
\qquad
\mathbf A=
\begin{bmatrix}
\mathbf I-\eta\mathbf H_{\text{ref}} & -\eta\kappa\,\mathbf I\\
\mathbf H_{\text{ref}} & \kappa\,\mathbf I
\end{bmatrix},\qquad
\mathbf G=\begin{bmatrix}-\eta\mathbf I\\ \mathbf I\end{bmatrix}.
\]
Since covariance is translation-invariant, the drift $\bm{g}_{\text{ref}}$ does not affect it: defining the centered state
\(\tilde{\bm{u}}_t:=\bm{u}_t-\mathbb{E}[\bm{u}_t\mid\mathcal S]\),
we have
\(
\tilde{\bm{u}}_{t+1}=\mathbf A\,\tilde{\bm{u}}_t+\mathbf G\,\bm{\xi}_t.
\)
Assume the mini-batch noises \(\{\bm{\xi}_t\}\) satisfy the martingale-difference (conditional unbiasedness) property
\(\mathbb E[\bm{\xi}_t\mid \mathcal F_t,\mathcal S]=\bm{0}\),
so that \(\Cov(\tilde{\bm{u}}_t,\bm{\xi}_t\mid\mathcal S)=\mathbf 0\). Then taking covariances at stationarity yields the discrete Lyapunov equation
\[
\boldsymbol{\Sigma}_{\mathrm{joint}}
=\mathbf A\,\boldsymbol{\Sigma}_{\mathrm{joint}}\,\mathbf A^\top+\mathbf G\,\mathbf C\,\mathbf G^\top,
\qquad
\mathbf C:=\Cov(\bm{\xi}_t\mid\mathcal S).
\]
Let \(\mathbf{\Pi}=[\,\mathbf I\;\;\mathbf 0\,]\in\mathbb R^{m\times 2m}\) be the projection onto the parameter component; then
\(
\boldsymbol{\Sigma}=\Cov(\bm{w}_t)=\mathbf{\Pi}\,\boldsymbol{\Sigma}_{\mathrm{joint}}\,\mathbf{\Pi}^\top,
\)
which yields the projected Lyapunov characterization of the stationary parameter covariance.

\emph{Closed form under commutativity.}
Assume \(\mathbf H_{\text{ref}}\succ0\), \(\mathbf C\) is finite, and \(\mathbf H_{\text{ref}}\mathbf C=\mathbf C\mathbf H_{\text{ref}}\). Then there exists an orthogonal \(\mathbf U\) such that
\(
\mathbf H_{\text{ref}}=\mathbf U\boldsymbol{\Lambda}\mathbf U^\top,\ 
\mathbf C=\mathbf U\boldsymbol{\Gamma}\mathbf U^\top
\)
with \(\boldsymbol{\Lambda}=\mathrm{diag}(\lambda_1,\ldots,\lambda_m)\) and \(\boldsymbol{\Gamma}=\mathrm{diag}(\gamma_1,\ldots,\gamma_m)\).
Define the block-orthogonal transform \(\mathbf T:=\mathrm{diag}(\mathbf U,\mathbf U)\) and rotated state/noise
\(
\bm{z}_t:=\mathbf T^\top\bm{u}_t,\ \bm{\zeta}_t:=\mathbf U^\top\bm{\xi}_t
\).
In this eigenbasis, the dynamics decouple:
\[
\bm{z}_t=\widetilde{\mathbf A}\,\bm{z}_{t-1}+\widetilde{\mathbf G}\,\bm{\zeta}_{t-1}, \text{ where }
\:\:
\widetilde{\mathbf A}=
\begin{bmatrix}\mathbf I-\eta\boldsymbol{\Lambda}&-\eta\kappa\,\mathbf I\\ \boldsymbol{\Lambda}&\kappa\,\mathbf I\end{bmatrix},\quad
\widetilde{\mathbf G}=\begin{bmatrix}-\eta\mathbf I\\ \mathbf I\end{bmatrix}, \quad 
\Cov(\bm{\zeta}_t)=\boldsymbol{\Gamma}.
\]
Hence each eigendirection $i$ follows a $2\times 2$ “heavy-ball’’ system with curvature $\lambda=\lambda_i$ and noise variance $\gamma=\gamma_i$:
\[
\mathbf A(\lambda)=
\begin{bmatrix}
1-\eta\lambda & -\eta\kappa \\
\lambda & \kappa
\end{bmatrix},\quad
\mathbf G=\begin{bmatrix}-\eta\\ 1\end{bmatrix},\quad
\mathbf Q(\gamma)=\mathbf G\,\gamma\,\mathbf G^\top
=\begin{bmatrix}\gamma\eta^2 & -\gamma\eta \\ -\gamma\eta & \gamma\end{bmatrix}.
\]

Let $\mathbf S=\begin{bmatrix}x&y\\ y&z\end{bmatrix}$ be the stationary joint covariance in this mode, solving
$\mathbf S=\mathbf A(\lambda)\,\mathbf S\,\mathbf A(\lambda)^\top+\mathbf Q(\gamma)$.
Solving the resulting linear system in $x,y,z$ (unique under stability) gives the parameter variance
\[
x=\frac{\gamma\,\eta(1+\kappa)}{\lambda(1-\kappa)\,\big(2(1+\kappa)-\eta\lambda\big)}.
\]
Therefore, in the eigenbasis the stationary parameter covariance is diagonal with entries
\[
\frac{\eta}{1-\kappa}\cdot\frac{\gamma_i}{\lambda_i\left(2-\frac{\eta}{1+\kappa}\lambda_i\right)}
\ =\ \Big(\lambda_i\big(2-\frac{\eta}{1+\kappa}\lambda_i\big)\Big)^{-1}\,\frac{\eta}{1-\kappa}\,\gamma_i,
\]
and conjugating back by \(\mathbf U\) yields
\[
\boldsymbol{\Sigma}
\;=\;
\left[\ \mathbf H_{\text{ref}}\left(2\mathbf I-\frac{\eta}{1+\kappa}\mathbf H_{\text{ref}}\right)\ \right]^{-1}\,
\frac{\eta}{1-\kappa}\,\mathbf C,
\]
which matches \Eqref{eq:stationary posterior covariance} after identifying $\mathbf H_{\text{ref}}$ with $\hat{\mathbf{H}}_{\epsilon}(\tilde{\bm{w}})$.

\emph{Stability condition.}
For the scalar mode, the characteristic polynomial of \(\mathbf A(\lambda)\) is
\(t^2-(1-\eta\lambda+\kappa)t+\kappa\).
All roots lie in the open unit disk iff \(|\kappa|<1\) and \(0<\eta\lambda<2(1+\kappa)\) (e.g., Jury criterion). 
Since \(\kappa\in[0,1)\), this equivalently requires \(0<\frac{\eta}{1+\kappa}\lambda_i<2\) for every \(i\).
Under this condition, the stationary covariance exists and is finite.
\end{proof}

\subsection{Non-Stationary Regime: Proof of Lemma~\ref{lemma:windowed_mean_covariance_recursion}}
\label{append:proof_windowed_mean_covariance_recursion}

\begin{proof}[Proof of Lemma~\ref{lemma:windowed_mean_covariance_recursion}]
Fix a window start time $t$ and let $t'=t+k$. Under the within-window invariance assumption, the linearized SGD dynamics (Lemma~\ref{lemma:cov_param_projection_detailed}) take the affine state-space form
\[
\bm{u}_{s+1}
=
\mathbf{A}_t \bm{u}_{s}
+\mathbf{G}\big[\nabla \hat{\mathcal{L}}_{\epsilon}(\tilde{\bm{w}}_t)+\bm{\xi}_s\big],
\qquad s=t,t+1,\ldots,t'-1,
\]
where $\bm{u}_s=[\bm{w}_s-\tilde{\bm{w}}_t;\bm{v}_s]$ and $\mathbb{E}[\bm{\xi}_s\mid \mathcal{F}_s,\mathcal{S}]=\bm{0}$.
The finite-noise assumption ensures that the fixed conditional covariance $\mathbf C_t=\mathbb E[\bm{\xi}_s\bm{\xi}_s^\top\mid\mathcal{F}_s,\mathcal{S}]$ exists in the window, and the mode-wise stability condition $0<\eta_t\lambda_{t,i}<2(1+\kappa)$ for the fixed learning rate used inside the window guarantees that the propagated covariance terms remain finite in the analyzed modes.

Iterating the recursion yields
\[
\bm{u}_{t'}
=
\mathbf{A}_t^{k}\bm{u}_{t}
+\sum_{j=0}^{k-1}\mathbf{A}_t^j\mathbf{G}\nabla \hat{\mathcal{L}}_{\epsilon}(\tilde{\bm{w}}_t)
+\sum_{j=0}^{k-1}\mathbf{A}_t^j\mathbf{G}\bm{\xi}_{t'-1-j}.
\]
Taking expectation conditional on $\mathcal{S}$ eliminates the noise terms and gives the mean recursion in \Eqref{eq:non-stationary mean iteration}.

For the covariance recursion, let $\bm{\Omega}_s=\Cov(\bm{u}_s\mid \mathcal{S})$.
Under the martingale-difference property of the mini-batch noise (as used in the proof of Lemma~\ref{lemma:cov_param_projection_detailed}), the cross-covariance terms vanish when propagating second moments across time, yielding
\[
\bm{\Omega}_{t'}
=
\mathbf{A}_t^{k}\bm{\Omega}_{t}(\mathbf{A}_t^{k})^\top
+\sum_{j=0}^{k-1}\mathbf{A}_t^{j}\mathbf{G}\mathbf{C}_t\mathbf{G}^\top(\mathbf{A}_t^{j})^\top,
\]
which is exactly \Eqref{eq:non-stationary cov iteration}.
\end{proof}

\shortsection{Corollary: Modal Recursions under Commutativity}
To connect the matrix recursion in Lemma~\ref{lemma:windowed_mean_covariance_recursion} to the scalar (per-direction) quantities used in Theorem~\ref{thm:final robust generalization bound combined} and in our empirical estimation procedure, we project the $2m$-dimensional augmented dynamics onto the eigenspace of the local Hessian.
Following the stationary analysis, we adopt the same (approximate) commutativity assumption within the window: $\hat{\mathbf{H}}_{\epsilon}(\tilde{\bm{w}}_t)$ and the finite covariance $\mathbf{C}_t$ are simultaneously diagonalizable, so each stable eigendirection evolves independently.

\paragraph{Modal covariance recursion (commuting approximation).}
Let $(\lambda_i,\bm{e}_i)$ be an eigenpair of $\hat{\mathbf{H}}_{\epsilon}(\tilde{\bm{w}}_t)$ and define the (modal) noise variance $\gamma_i := \bm{e}_i^\top \mathbf{C}_t\bm{e}_i$.
The corresponding $2 \times 2$ projected transition matrix is
\begin{equation}
\label{eq:projected_transition}
    \mathbf{A}_i = \begin{bmatrix} 1 - \eta \lambda_i & -\eta\kappa \\ \lambda_i & \kappa \end{bmatrix},
\end{equation}
with projected noise injection $\mathbf{Q}_i = \gamma_i\,\mathbf{G}_i\mathbf{G}_i^\top$ where $\mathbf{G}_i = [-\eta \;\; 1]^\top$.
Let $\bm{\Omega}_{t,i}\in\mathbb{R}^{2\times 2}$ denote the modal augmented covariance and define $\sigma_{t,i}^2 := [\bm{\Omega}_{t,i}]_{11}$.
Iterating the projected recursion yields
\begin{equation}
\label{eq:variance_iteration}
    \begin{aligned}
        \bm{\Omega}_{t+k,i}
        = \mathbf{A}_i^{k}\bm{\Omega}_{t,i}(\mathbf{A}_i^\top)^{k}
        + \sum_{j=0}^{k-1}\mathbf{A}_i^{j}\mathbf{Q}_i(\mathbf{A}_i^\top)^{j}, \quad
        \sigma_{t+k,i}^2 &= [\bm{\Omega}_{t+k,i}]_{11}.
    \end{aligned}
\end{equation}
Stability requires $0<\eta\lambda_i < 2(1+\kappa)$ for all analyzed modes $i$ and for each learning rate used inside the window; when satisfied, the propagated term contracts at rate $O(\rho(\mathbf{A}_i)^{2k})$.
If the window begins immediately after a learning-rate decay $\eta_1\!\to\!\eta$, then $\sigma_{t,i}^2$ can be initialized by the stationary formula under $\eta_1$ using \Eqref{eq:stationary posterior covariance} and evolved forward under the new learning rate $\eta$ using \Eqref{eq:variance_iteration}.

\paragraph{Non-stationary mean recursion (projected drift).}
Under the within-window invariance assumption in Lemma~\ref{lemma:windowed_mean_covariance_recursion}, the same affine dynamics yield an explicit short-window recursion for the mean drift.
Define the \emph{conditional} mean drift within the window
$\Delta \bar{\bm{w}}_{t+k \mid t} := \mathbb{E}[\bm{w}_{t+k} \mid \mathcal{F}_t, \mathcal{S}] - \tilde{\bm{w}}_t$.
Then
\begin{equation}
\label{eq:mean_iteration}
    \Delta \bar{\bm{w}}_{t+k \mid t}
    = \mathbf{\Pi} \mathbf{A}_t^k \bm{u}_t
    + \sum_{j=0}^{k-1} \mathbf{\Pi} \mathbf{A}_t^j \mathbf{G} \nabla \hat{\mathcal{L}}_{\epsilon}(\tilde{\bm{w}}_t),
\end{equation}
where $\mathbf{\Pi} = [\mathbf{I} \; \mathbf{0}]$ is the projection onto the parameter subspace and
$\bm{u}_t = [\bm{w}_t - \tilde{\bm{w}}_t; \bm{v}_t]$ is the augmented state.
Taking expectation conditional on $\mathcal{S}$ yields the \emph{posterior} mean drift
$\Delta \bar{\bm{w}}_{t+k} := \bm{\mu}_{t+k} - \tilde{\bm{w}}_t$ with $\bm{\mu}_{t+k}=\mathbb{E}[\bm{w}_{t+k}\mid \mathcal{S}]$:
\begin{equation}
\label{eq:mean_iteration_uncond}
    \Delta \bar{\bm{w}}_{t+k}
    = \mathbf{\Pi} \mathbf{A}_t^k \bar{\bm{u}}_t
    + \sum_{j=0}^{k-1} \mathbf{\Pi} \mathbf{A}_t^j \mathbf{G} \nabla \hat{\mathcal{L}}_{\epsilon}(\tilde{\bm{w}}_t),
\end{equation}
where $\bar{\bm{u}}_t=\mathbb{E}[\bm{u}_t\mid \mathcal{S}]=[\bm{\mu}_t-\tilde{\bm{w}}_t;\bar{\bm{v}}_t]$ with $\bm{\mu}_t=\mathbb{E}[\bm{w}_t\mid \mathcal{S}]$ and $\bar{\bm{v}}_t=\mathbb{E}[\bm{v}_t\mid \mathcal{S}]$.

\subsection{Closed-Form Bound: Proof of Theorem~\ref{thm:final robust generalization bound combined}}
\label{append:proof_transient_bound}

\begin{proof}[Proof of Theorem~\ref{thm:final robust generalization bound combined}]
We divide the proof of Theorem \ref{thm:final robust generalization bound combined} into three steps.
\shortsection{Step 1: Start from the compact PAC-Bayesian bound}
Fix a timestep $t$ and consider the posterior $\mathcal{Q}_{t}=\mathcal{N}(\bm{\mu}_{t},\mathbf{\Sigma}_{t})$ induced by momentum SGD.
Applying Theorem~\ref{coro:compact_pac_bound} with reference point $\tilde{\bm{w}}_t$ gives
\begin{align}
\label{eq:appendix_thm47_start}
    \nonumber \mathbb{E}_{\bm{w}\sim\mathcal{Q}_{t}} \big[\mathcal{L}_{\epsilon}(\bm{w})\big]
    \leq\;&
    \hat{\mathcal{L}}_{\epsilon}(\tilde{\bm{w}}_t) + \big\langle \nabla \hat{\mathcal{L}}_{\epsilon}(\tilde{\bm{w}}_t), \bm{\mu}_t - \tilde{\bm{w}}_{t} \big\rangle
    + \frac{1}{2} (\bm{\mu}_t - \tilde{\bm{w}}_{t})^{\top}  \hat{\mathbf{H}}_{\epsilon} (\tilde{\bm{w}}_t) (\bm{\mu}_t - \tilde{\bm{w}}_{t}) \\
	    &+ \frac{1}{2} \mathrm{Tr} \big( \hat{\mathbf{H}}_{\epsilon}(\tilde{\bm{w}}_t)\mathbf{\Sigma}_{t} \big)
	    + \frac{1}{2\beta \sigma^{2}_{\mathcal{P}}}\Big(\| \bm{\mu}_{t} \|_{2}^{2} + \mathrm{Tr}(\mathbf{\Sigma}_{t}) \Big)
	    - \frac{1}{2\beta} \ln \det (\mathbf{\Sigma}_{t}) + C_{\mathrm{PB}}
	    + R_t.
\end{align}

\shortsection{Step 2: Matrix form and spectral specialization}
Equation~\ref{eq:robust generalization bound combined} follows directly in matrix form from \Eqref{eq:appendix_thm47_start} after collecting the posterior-independent constants into $c_{\mathrm{PB}}$.
Under the additional simultaneous-diagonalization condition $\mathbf{H}_t\mathbf{\Sigma}_t=\mathbf{\Sigma}_t\mathbf{H}_t$, the trace terms admit the spectral representation
\[
\mathrm{Tr} \big( \hat{\mathbf{H}}_{\epsilon}(\tilde{\bm{w}}_t)\mathbf{\Sigma}_{t} \big)
=\sum_{i=1}^{m}\lambda_{t,i}\,\sigma_{t,i}^2,
\quad
\ln \det (\mathbf{\Sigma}_{t})
=\sum_{i=1}^{m}\ln \sigma_{t,i}^2,
\quad
\mathrm{Tr}(\mathbf{\Sigma}_{t})
=\sum_{i=1}^{m}\sigma_{t,i}^2.
\]
The covariance-trace KL contribution then combines with the curvature-weighted variance as:
\[
\frac{1}{2}\operatorname{Tr}(\mathbf{H}_t\mathbf{\Sigma}_t)+
\frac{1}{2\beta\sigma_{\mathcal P}^2}\operatorname{Tr}(\mathbf{\Sigma}_t)
=
\frac{1}{2}\sum_{i=1}^m\left(\lambda_{t,i}+\frac{1}{\beta\sigma_{\mathcal P}^2}\right)\sigma_{t,i}^2.
\]
This is the spectral specialization used for dominant-mode diagnostics, not an extra requirement for the matrix-form bound.

\shortsection{Step 3: Stationary vs.\ non-stationary posterior parameters}
Finally, the bound in \Eqref{eq:robust generalization bound combined} is made explicit by plugging in the posterior parameters.
In the stationary regime, Lemma~\ref{lemma:cov_param_projection_detailed} provides the closed-form mean and covariance
(Equations \ref{eq:stationary posterior mean} and \ref{eq:stationary posterior covariance}).
In particular, letting $\bm{g}_t := \nabla\hat{\mathcal{L}}_{\epsilon}(\tilde{\bm{w}}_t)$ and
$\mathbf H_t := \hat{\mathbf{H}}_{\epsilon}(\tilde{\bm{w}}_t)$, the stationary mean drift
$\Delta\bar{\bm{w}}_t=\bm{\mu}_{t}-\tilde{\bm{w}}_t=-\mathbf H_t^{-1}\bm{g}_t$
implies the familiar Newton-type optimization gain
\[
\big\langle \bm{g}_t, \Delta\bar{\bm{w}}_t \big\rangle
\;+\;\frac{1}{2}\Delta\bar{\bm{w}}_t^\top \mathbf H_t\,\Delta\bar{\bm{w}}_t
\;=\;
-\frac{1}{2}\bm{g}_t^\top \mathbf H_t^{-1}\bm{g}_t.
\]
In the non-stationary regime, Lemma~\ref{lemma:windowed_mean_covariance_recursion} provides the windowed matrix recursion for $(\bm{\mu}_{t+k},\bm{\Omega}_{t+k})$, and the modal recursion in Appendix~\ref{append:proof_windowed_mean_covariance_recursion} (Equations~\ref{eq:projected_transition} and \ref{eq:variance_iteration}) yields the corresponding scalar variances $\{\sigma_{t+k,i}^2\}_{i=1}^m$ under commutativity.
Substituting these quantities into \Eqref{eq:robust generalization bound combined} completes the proof.
\end{proof}

\section{Proofs of Theoretical Results in Section~\ref{sec:preliminaries} \& Section~\ref{sec:experiments}}

\subsection{Proof of Theorem~\ref{thm:generic advbound}}
\label{app:bounded_unbounded_losses}
\label{append:proof of theorem advbound}

Theorem~\ref{thm:generic advbound} is stated for bounded losses because Catoni's bounded-difference concentration is used in the generic PAC-Bayesian step. This condition is satisfied directly by bounded surrogate losses and clipped cross-entropy. For the cross-entropy losses used in our experiments, one can replace the bounded-loss concentration step by PAC-Bayes-Chernoff bounds for unbounded losses under light-tail or moment assumptions~\citep{casado2024pac}. This changes concentration constants or tail terms but not the Section~\ref{sec:pac-bayesian robust generalization} posterior-averaged loss and Gaussian-KL decomposition.

To prove Theorem~\ref{thm:generic advbound}, we first state a general PAC-Bayes inequality, 
also known as Catoni's bound~\citep{catoni2007pac}.

\begin{lemma}[PAC-Bayes Bound (Catoni's bound)]
\label{lemma:pacbound-generic}
Let $\alpha \in (0,1)$, $\beta > 0$, and $\mathcal{D}$ be any distribution over $\mathcal{X}\times\mathcal{Y}$. 
Let $\mathcal{W}$ be a parameter space and $\mathcal{P}$ be a data-independent prior on $\mathcal{W}$. 
Consider any measurable loss $\ell:\mathcal{W}\times\mathcal{X}\times\mathcal{Y}\to[0,C_\ell]$ bounded by the scalar $C_\ell>0$. 
Given an i.i.d.\ sample set $\mathcal{S}=\{(\bm{x}_i,y_i)\}_{i=1}^{|\mathcal{S}|}$ drawn from $\mathcal{D}$, 
for any posterior $\mathcal{Q}$ on $\mathcal{W}$, with probability at least $1-\alpha$ over the draw of $\mathcal{S}$, we have
\begin{align*}
\mathbb{E}_{\bm{w}\sim\mathcal{Q}}\mathbb{E}_{(\bm{x},y)\sim\mathcal{D}}\big[\ell(\bm{w};\bm{x},y)\big]
\;\le\;
\mathbb{E}_{\bm{w}\sim\mathcal{Q}}\bigg[\frac{1}{|\mathcal{S}|}\sum_{(\bm{x},y)\in\mathcal{S}}\ell(\bm{w};\bm{x},y)\bigg]
+ \frac{\beta C_\ell^2}{8|\mathcal{S}|}
+ \frac{1}{\beta} \mathrm{KL}(\mathcal{Q}\Vert\mathcal{P}) - \frac{1}{\beta} \ln\alpha.
\end{align*}
\end{lemma}

\begin{proof}[Proof of Theorem~\ref{thm:generic advbound}]
We instantiate Lemma~\ref{lemma:pacbound-generic} with the adversarial loss
\[
\tilde{\ell}(\bm{w};\bm{x},y) \;:=\; \ell_{\mathrm{adv}}(\bm{w},\bm{x},y)
= \max_{\bm{\delta}\in B_{\varepsilon}(0)} \ell\!\left(\bm{w};\, \bm{x}+\bm{\delta}, y\right),
\]
where the base loss $\ell$ is bounded by $C_\ell$ and the perturbation set $B_{\varepsilon}(0)$ is fixed. 
Since $\ell \in [0,C_\ell]$, the maximization preserves boundedness, so $\tilde{\ell} \in [0,C_\ell]$. 
Measurability follows directly from that of $\ell$ and the continuity of $(\bm{x},\bm{\delta}) \mapsto \ell(\bm{w};\bm{x}+\bm{\delta},y)$.
Applying Lemma~\ref{lemma:pacbound-generic} to $\tilde{\ell}$ yields
\[
\mathbb{E}_{\bm{w}\sim\mathcal{Q}}\mathbb{E}_{(\bm{x},y)\sim\mathcal{D}} \big[\ell_{\mathrm{adv}}(\bm{w},\bm{x},y)\big]
\;\le\;
\mathbb{E}_{\bm{w}\sim\mathcal{Q}}\bigg[\frac{1}{|\mathcal{S}|}\sum_{(\bm{x},y)\in\mathcal{S}} \ell_{\mathrm{adv}}(\bm{w},\bm{x},y)\bigg]
+\frac{\beta C_\ell^2}{8|\mathcal{S}|}
+ \frac{1}{\beta} \mathrm{KL}(\mathcal{Q}\Vert\mathcal{P}) - \frac{1}{\beta} \ln\alpha.
\]
Plugging the definitions of empirical and expected adversarial losses and
rearranging the terms completes the proof.
\end{proof}

\subsection{Proof of Proposition \ref{prop:robust optimization & curvature}}
\label{sec:proof proposition adversarial training -> Hessian increase}

\begin{proof}[Proof of Proposition \ref{prop:robust optimization & curvature}]
We prove the proposition in terms of the two commonly adopted perturbation metrics in the prior literature on adversarial training: $L_2$ norm and $L_{\infty}$ norm. In particular, our derivations rely on the assumption that the adversarial loss can be locally linearized for each training example (e.g., when the perturbation strength $\epsilon$ is small enough).

\shortsection{$\bm{L_2}$ Case}
We start with the simpler case of $L_2$ perturbations.
Consider the adversarial training objective $\hat{\mathcal{L}}_{\epsilon}$ in Equation~\ref{eq:def empirical adv loss} with $p=2$. Assuming the adversarial loss can be locally linearized at each $(\bm{x}, y)$, we can show that
\begin{align}
\label{eq:linearized empirical adv loss}
    \hat{\mathcal{L}}_{\epsilon}(\bm{w}; L_2) = \frac{1}{|\mathcal{S}|} \sum_{(\bm{x},y)\in\mathcal{S}} \bigg[ \ell(\bm{w}, \bm{x}, y) + \max_{\|\bm\delta\|_2 \leq \epsilon} \langle \bm{\delta}, \nabla_{\bm{x}} \ell(\bm{w}, \bm{x}, y) \rangle \bigg] = \hat{\mathcal{L}}(\bm{w}) + \frac{\epsilon}{|\mathcal{S}|} \sum_{(\bm{x},y)\in\mathcal{S}} \big\| \nabla_{\bm{x}} \ell(\bm{w}, \bm{x}, y) \big\|_2,
\end{align}
where $\hat{\mathcal{L}}(\bm{w}) = \frac{1}{|\mathcal{S}|} \sum_{(\bm{x},y)\in\mathcal{S}} \ell(\bm{w}, \bm{x}, y)$ denotes the clean loss, and the second equality holds due to Cauchy–Schwarz inequality. Equation \ref{eq:linearized empirical adv loss} suggests that achieving low empirical adversarial loss requires: (i) a good fit of the clean examples (i.e., low $\hat{\mathcal{L}}(\bm{w})$), and (ii) small input gradient sensitivity across the training dataset (i.e., low $\| \nabla_{x} \ell(\bm{w}, \bm{x}, y) \|_2$). 

Taking the derivative of Equation~\ref{eq:linearized empirical adv loss} with respect to the model weights gives the gradient of the adversarial loss:
\begin{align}
\label{eq:linearized gradient derivation}
    \nonumber \nabla_{\bm{w}} \hat{\mathcal{L}}_{\epsilon}(\bm{w}; L_2) &= \nabla_{{\bm{w}}} \hat{\mathcal{L}}(\bm{w}) + \frac{\epsilon}{|\mathcal{S}|} \sum_{(\bm{x},y)\in\mathcal{S}} \nabla_{{\bm{w}}} \bigg( \big\| \nabla_{\bm{x}} \ell(\bm{w}, \bm{x}, y) \big\|_2 \bigg) \\
    & = \nabla_{{\bm{w}}} \hat{\mathcal{L}}(\bm{w}) + \frac{\epsilon}{|\mathcal{S}|} \sum_{(\bm{x},y)\in\mathcal{S}} {\big\| \nabla_{\bm{x}} \ell(\bm{w}, \bm{x}, y) \big\|_2}^{-1} \bigg( \nabla_{\bm{w}} \nabla_{\bm{x}} \ell(\bm{w}, \bm{x}, y) \bigg)^{\top}  \nabla_{\bm{x}} \ell(\bm{w}, \bm{x}, y),
\end{align}
where the second equality follows from the chain rule. Taking another derivative with respect to the weight parameters of Equation~\ref{eq:linearized gradient derivation} gives us the connection between the Hessian in the parameter space and the input gradient sensitivity:
\begin{align}
\label{eq:linearized Hessian derivation}
    \nonumber \nabla^2_{\bm{w}} \hat{\mathcal{L}}_{\epsilon}(\bm{w}; L_2) &= \nabla^2_{{\bm{w}}} \hat{\mathcal{L}}(\bm{w}) + \frac{\epsilon}{|\mathcal{S}|} \sum_{(\bm{x},y)\in\mathcal{S}} \nabla_{{\bm{w}}} \bigg( {\big\| \nabla_{\bm{x}} \ell(\bm{w}, \bm{x}, y) \big\|_2}^{-1} \bigg( \nabla_{\bm{w}} \nabla_{\bm{x}} \ell(\bm{w}, \bm{x}, y) \bigg)^{\top}  \nabla_{\bm{x}} \ell(\bm{w}, \bm{x}, y) \bigg) \\
    &= \underbrace{\nabla^2_{{\bm{w}}} \hat{\mathcal{L}}(\bm{w})}_{\text{clean Hessian}} + \frac{\epsilon}{|\mathcal{S}|} \sum_{(\bm{x},y)\in\mathcal{S}} \underbrace{{\big\| \nabla_{\bm{x}} \ell(\bm{w}, \bm{x}, y) \big\|_2}^{-1} \cdot \bigg( \nabla_{\bm{w}} \nabla_{\bm{x}} \ell(\bm{w}, \bm{x}, y) \bigg)^{\top} \nabla_{\bm{w}} \nabla_{\bm{x}} \ell(\bm{w}, \bm{x}, y)}_{\text{PSD cross-Hessian term}} + \mathcal{R}(\bm{w}),
\end{align}
where $\mathcal{R}(\bm{w})$ stands for the remaining negligible terms, involving a third-order derivative term and negative rank-1 correction from norm normalization. One can expect the remaining terms to have a limited effect on the large Hessian eigenvalues compared to the two terms explicitly specified in Equation \ref{eq:linearized Hessian derivation}. Compared with the clean counterpart, the Hessian of the adversarial training loss includes a positive semi-definite (PSD) term averaged across the training dataset, involving the cross Hessian $\nabla_{\bm{w}} \nabla_{\bm{x}} \ell(\bm{w}, \bm{x}, y)$, which captures how the input gradient of adversarial loss changes with $\bm{w}$. 

Note that given any direction in parameter space $\bm{v}\in\mathbb{R}^m$, we have
\begin{align}
\label{eq:Hessian to input gradient sensitivity to parameters}
     \bm{v}^{\top} \bigg( {\big\| \nabla_{\bm{x}} \ell(\bm{w}, \bm{x}, y) \big\|_2}^{-1} \cdot \bigg( \nabla_{\bm{w}} \nabla_{\bm{x}} \ell(\bm{w}, \bm{x}, y) \bigg)^{\top} \nabla_{\bm{w}} \nabla_{\bm{x}} \ell(\bm{w}, \bm{x}, y) \bigg) \bm{v} = \frac{\big\| \nabla_{\bm{w}} \nabla_{\bm{x}} \ell(\bm{w}, \bm{x}, y) \bm{v} \big\|_2^2}{\big\| \nabla_{\bm{x}} \ell(\bm{w}, \bm{x}, y) \big\|_2} \geq 0.
\end{align}
Equation \ref{eq:Hessian to input gradient sensitivity to parameters} suggests that: (i) the curvature of adversarial loss is always expected to be larger than that of clean loss due to the extra PSD term, which aligns with our empirical observations that the top Hessian eigenvalues of adversarially trained models are larger than those of standard training, and (ii) if the top eigenvalues of the cross Hessian grow or the $\ell_2$ norm of the input gradient reduces, the curvature of the adversarial loss will accordingly grow.

\shortsection{$\bm{L_{\infty}}$ Case}
Now, we turn to prove Proposition \ref{prop:robust optimization & curvature} under $L_{\infty}$-norm bounded perturbations. Consider Equation \ref{eq:def empirical adv loss} with $p=\infty$. Under the same local loss linearization assumption, according to Hölder's inequality, we have 
\begin{align}
\label{eq:linearized empirical adv loss infty}
    \hat{\mathcal{L}_{\epsilon}}(\bm{w}; L_\infty) = \hat{\mathcal{L}}(\bm{w}) + \frac{\epsilon}{|\mathcal{S}|} \sum_{(x,y)\in\mathcal{S}} \big\| \nabla_{\bm{x}} \ell(\bm{w}, \bm{x}, y) \big\|_1.
\end{align}
Similar to the $L_2$ case, we next compute the first-order derivative of the adversarial loss with respect to $\bm{w}$:
\begin{align*}
    \nabla_{\bm{w}} \hat{\mathcal{L}_{\epsilon}}(\bm{w}; L_{\infty}) = \nabla_{\bm{w}} \hat{\mathcal{L}}(\bm{w}) + \frac{\epsilon}{|\mathcal{S}|} \sum_{(\bm{x},y)\in\mathcal{S}} \big( \nabla_{\bm{w}} \nabla_{\bm{x}} \ell(\bm{w}, \bm{x}, y) \big)^{\top} \mathrm{sgn}(\nabla_{\bm{x}}  \ell(\bm{w}, \bm{x}, y)),
\end{align*}
where $\mathrm{sgn}(\cdot)$ denotes the sign operator. For simplicity, denote by $g(\bm{w},\bm{z}) = \nabla_{\bm{x}} \ell(\bm{w},\bm{x},y)$ the input gradient at $\bm{z}=(\bm{x},y)$ in the following derivation. A key difference from the $L_2$ case is how to compute the second-order derivative given the sign operator. Instead of directly differentiating through the sign operator, one can make use of the following fact:
\begin{align*}
    \bm{\delta}^* = \epsilon \cdot \mathrm{sgn}\big(\nabla_{\bm{x}}  \ell(\bm{w}, \bm{x}, y)\big) = \argmax_{\|\bm\delta\|_{\infty}\leq \epsilon} \: \big\langle g(\bm{w}, \bm{z}), \bm\delta \big\rangle,
\end{align*}
Note that it is a linear programming (LP) in terms of $\bm\delta$ with coefficient $g(\bm{w},\bm{z})$ and a fixed $L_{\infty}$-norm ball. Taking the derivative with respect to $\bm{w}$ and using the chain rule of generalized derivatives, we obtain:
\begin{align*}
    \frac{\partial\bm\delta^*}{\partial \bm{w}} = \mathbf{D}(\bm{w}, \bm{z}) \cdot \nabla_{\bm{w}} g(\bm{w}, \bm{z}),
\end{align*}
where $\mathbf{D}(\bm{w}, \bm{z}) = \mathrm{diag}(d_1, d_2, \ldots, d_m)$ is a diagonal matrix that encodes the active set of coordinates: $d_i=1$ if the $i$-th coordinate $[g(\bm{w}, \bm{z})]_i = 0$ (sign-switching region, constraint is inactive), and $d_i=0$ if $[g(\bm{w}, \bm{z})]_i \neq 0$ (locally stable sign, saturated coordinate). The above equation results from the classical LP sensitivity theory~\citep{rockafellar1998variational}, where $\mathbf{D}$ is essentially an active-set selection matrix arising from the generalized derivative of the LP solution map. 

Now, we can derive the second-order derivative with respect to the adversarial loss under $L_{\infty}$ perturbations:
\begin{align}
\label{eq:Hessian to input gradient sensitivity to parameters infty}
    \nonumber \nabla_w^2 \hat{\mathcal{L}}_{\epsilon}(\bm{w}; L_\infty) &= \nabla_{\bm{w}}^2 \hat{\mathcal{L}}(\bm{w}) + \frac{\epsilon}{|\mathcal{S}|} \sum_{\bm{z}\in\mathcal{S}} \big( \nabla_{\bm{w}} g(\bm{w}, \bm{z}) \big)^{\top} \mathbf{D}(\bm{w}, \bm{z}) \big(\nabla_{\bm{w}} g(\bm{w}, \bm{z})\big) + \mathcal{R}(\bm{w}) \\
    &= \underbrace{\nabla^2_{{\bm{w}}} \hat{\mathcal{L}}(\bm{w})}_{\text{clean Hessian}} + \frac{\epsilon}{|\mathcal{S}|} \sum_{(\bm{x},y)\in\mathcal{S}} \underbrace{\bigg( \nabla_{\bm{w}} \nabla_{\bm{x}} \ell(\bm{w}, \bm{x}, y) \bigg)^{\top} \mathbf{D}(\bm{w}, \bm{x}, y) \bigg(\nabla_{\bm{w}} \nabla_{\bm{x}} \ell(\bm{w}, \bm{x}, y)\bigg)}_{\text{PSD cross-Hessian term}} + \mathcal{R}(\bm{w}),
\end{align}
where $\mathcal{R}(w)$ stands for the remainder terms, such as the higher-order derivatives. 
Comparing Equation \ref{eq:Hessian to input gradient sensitivity to parameters infty} to Equation \ref{eq:Hessian to input gradient sensitivity to parameters}, we note that the difference for $L_\infty$ is that the PSD cross-Hessian term is controlled by the diagonal active-set projector $\mathbf{D}(\bm{w}, \bm{x}, y)$, whereas the cross-Hessian term for $L_2$ case is scaled by a smooth projector $1/\|\nabla_{\bm{x}} \ell(\bm{w}, \bm{x}, y)\|_2$.
Finally, for any direction $\bm{v}\in\mathbb{R}^m$ in the parameter space, we have
\begin{align*}
    \bm{v}^{\top} \bigg( \nabla_{\bm{w}} \nabla_{\bm{x}} \ell(\bm{w}, \bm{x}, y) \bigg)^{\top} \mathbf{D}(\bm{w}, \bm{x}, y) \bigg(\nabla_{\bm{w}} \nabla_{\bm{x}} \ell(\bm{w}, \bm{x}, y)\bigg) \bm{v} = \bigg\| \mathbf{D}(\bm{w}, \bm{x}, y)^{1/2} \cdot \nabla_{\bm{w}}\nabla_{\bm{x}} \ell(\bm{w}, \bm{x}, y) \bm{v} \bigg\|_2^2 \geq 0,
\end{align*}
which immediately suggests that the cross-Hessian term is positive semi-definite (PSD).

\shortsection{Why adversarial training increases loss curvature}
Now we can explain why the adversarial training objective encourages an increase in the top Hessian eigenvalues. First, optimizing the empirical adversarial loss reduces the average input gradient sensitivity $\frac{1}{|\mathcal{S}|} \sum_{(\bm{x},y)\in\mathcal{S}} \| \nabla_{\bm{x}} \ell(\bm{w}, \bm{x}, y) \|_2$ by design (specifically for the $L_2$ case); otherwise, the second term in Equation~\ref{eq:linearized empirical adv loss} will not be well-controlled. 
Second, note that robust features are known to be highly anisotropic, whereas standard features are often noisier; effective adversarial training requires aligning input gradients across training examples within the robust feature subspace, which also suggests the same for the parameter gradients along the optimization trajectory. This implies that input and parameter gradients will become increasingly aligned as adversarial training advances. 

Note that the cross-Hessian term for each individual data point can be expressed as:
\begin{align}
\label{eq:rank-1 expression of cross Hessian}
    \nabla_{\bm{w}} \nabla_{\bm{x}} \ell(\bm{w}, \bm{x}, y) = \frac{\partial^2 \ell}{\partial f^2} (\nabla_{\bm{w}} f) (\nabla_{\bm{x}} f)^{\top} + \frac{\partial \ell}{\partial f} \nabla_{\bm{w}} \nabla_{\bm{x}} f, \: \text{ where } \: \ell(\bm{w}, \bm{x}, y) = \ell(f_{\bm{w}}(\bm{x}), y).
\end{align}
Near a well-fitted solution, we know $\frac{\partial \ell}{\partial f} \rightarrow 0$. This suggests that the second term in Equation \ref{eq:rank-1 expression of cross Hessian} is negligible compared to the first term, which is essentially the outer product between input and parameter gradient (rank 1 per sample). Enhanced alignment reinforces the outer-product growth and eventually boosts the spectral concentration of the cross Hessian. 
According to Equations~\ref{eq:Hessian to input gradient sensitivity to parameters} and~\ref{eq:Hessian to input gradient sensitivity to parameters infty}, increasing the perturbation size $\epsilon$ will naturally boost the Hessian eigenvalues, since the second term is proportional to $\epsilon$. A larger $\epsilon$ suggests a stricter alignment requirement for both input and parameter gradients, which can further enhance the spectral concentration with respect to the cross-Hessian term.
The above interpretations support our observations in Figure~\ref{fig:unified_spectra}: (i) ST can attain very low clean training loss and small Hessian eigenvalues simultaneously; (ii) AT lowers robust training loss in later stages when $\eta$ is small, aligning gradients within the robust feature subspace, which inevitably boosts the top Hessian eigenvalues; and (iii) AWP optimizes for worst-case weight perturbations and effectively controls curvature growth, but the sharpness penalization can weaken gradient alignment in the robust feature subspace (e.g., the $20$-th eigenvalue $\lambda_{20}$ remains constant), resulting in a smaller decrease in robust training loss. 
\end{proof}

\section{Detailed Experimental Setup}
\label{app:exp_settings}

\subsection{Datasets and Training Baselines}
We conduct experiments on CIFAR-10 \citep{krizhevsky2009learning} using the standard train/test split. Inputs are scaled to $[0,1]$ and normalized channel-wise. Data augmentation includes random cropping with $4$-pixel padding and random horizontal flipping. We also evaluate on CIFAR-100 \citep{krizhevsky2009learning} and SVHN \citep{netzer2011reading} to verify the generality of our findings. 
Unless stated otherwise, we use PreActResNet-18~\citep{he2016deep} and optimize with momentum SGD ($\kappa=0.9$) for $200$ epochs with batch size $128$ and weight decay $5\times 10^{-4}$.
For CIFAR-10 (and CIFAR-100), we use an initial learning rate of $0.1$ and a piecewise schedule decayed by a factor of $10$ at epochs $100$ and $150$.
For evaluating model robustness, we adopt PGD adversarial training (AT)~\citep{madry2017towards} as the default baseline. We generate $\ell_\infty$ adversarial perturbations with $\epsilon=8/255$, step size $2/255$, $10$ attack iterations, and $1$ random restart. 

For AWP~\citep{wu2020adversarial}, we follow the standard configuration with perturbation radius $\gamma_{\mathrm{AWP}}=0.01$ (no warmup) and use the same outer-loop optimizer and learning-rate schedule.
Unless stated otherwise, robust metrics are computed via PGD under the same $\epsilon$; we use $10$ steps for AT and $20$ steps for AWP.
All experiments were run on NVIDIA A100 GPUs, using a single GPU per training/estimation job.
Based on the wall-clock times recorded in our training logs and the end-to-end per-epoch estimation pipeline, reproducing all results and figures in this paper requires approximately $\sim 220$ GPU-hours in total (about $\sim 65$ GPU-hours for training and $\sim 155$ GPU-hours for spectral estimation).

\subsection{Spectral Estimation Protocol}
\label{app:spectral_details}

For computational efficiency, we approximate curvature and gradient-noise statistics using a small number of randomly sampled mini-batches. Specifically, unless stated otherwise, we estimate the top-$k$ Hessian eigenpairs with $k=20$ by averaging the loss over $M=128$ mini-batches, performing $30$ power iterations for each eigenpair, and computing the projected gradient-noise covariances from the same $M=128$ mini-batches. This stochastic approximation is sufficient to capture the dominant spectral structure while keeping runtime feasible.
Algorithm~\ref{alg:empirical_estimation} summarizes the empirical procedure used to estimate the spectral statistics and bound components reported in Section~\ref{sec:experiments}.
To obtain the Hessian spectrum, we approximate Hessian--vector products (HVPs) via the identity
\[
\hat{\mathbf{H}}_{\epsilon}(\bm{w}) \bm{v}
\;=\;
\nabla_{\bm{w}} \Big( \nabla_{\bm{w}} \hat{\mathcal{L}}_{\epsilon}(\bm{w})^\top \bm{v} \Big),
\]
which can be computed efficiently by automatic differentiation without explicitly forming $\hat{\mathbf{H}}_{\epsilon}$. 
We then apply power iteration with Gram--Schmidt orthogonalization to extract the top-$k$ eigenvectors $\{\bm{v}_i\}$, 
and estimate their associated eigenvalues using the Rayleigh quotient,
\[
\lambda_i \;=\; \frac{\bm{v}_i^\top (\hat{\mathbf{H}}_{\epsilon}(\bm{w}) \bm{v}_i)}{\bm{v}_i^\top \bm{v}_i}.
\]
This procedure is consistent with established practice for large-scale curvature estimation 
\citep{dangel2019backpack, yao2020pyhessian}.
For the posterior structure, we consider the stochastic gradients at the mini-batch level, 
$\bm{g}_b = \nabla_{\bm{w}} \mathcal{L}_b(\bm{w})$. Projecting these gradients onto the subspace spanned by the leading 
Hessian eigenvectors $\mathbf{V} = [\bm{v}_1,\dots,\bm{v}_k]$ yields the projected quantities 
$\bm{p}_b = \mathbf{V}^\top \bm{g}_b \in \mathbb{R}^k$. Their covariance matrix is
$
\Gamma = \mathrm{Cov}[\bm{p}_b],
$
whose diagonal entries $\gamma_i = \Gamma_{ii}$ quantify the variance of stochastic gradients along the principal curvature directions. 

By construction, this definition ensures that $\lambda_i$ characterizes curvature 
while $\gamma_i$ represents the corresponding noise magnitude in the same eigendirections. 
This approach follows recent empirical observations that gradient-noise covariance tends to align with the Hessian eigenspectrum in neural networks \citep{jastrzkebski2018relation,ziyin2025formation}, 
allowing for a direct analysis of curvature–noise interactions.
When instantiating our closed-form formulas (e.g., \Eqref{eq:stationary posterior covariance}), we further adopt a diagonal approximation in the Hessian eigenspace, using only the diagonal entries $\{\gamma_i\}$ of $\Gamma$.
This approximation is exact under the commutativity assumption (simultaneous diagonalizability of $\mathbf{H}$ and $\mathbf{C}$), and its accuracy in practice depends on the magnitude of the off-diagonal energy in $\Gamma$, which quantifies curvature--noise misalignment within the top-$k$ subspace.
We empirically verify that both commutativity and alignment hold to a good approximation throughout training under both AT and AWP (Appendix~\ref{app:comm_align}), justifying the use of $\{\lambda_i,\gamma_i\}$ as matched curvature--noise pairs in our empirical decomposition.

\shortsection{Negative eigenvalues}
The stationary covariance formula (\Eqref{eq:stationary posterior covariance}) requires local convexity ($\mathbf{H}\succ 0$); negative eigenvalues indicate saddle-point geometry where the stationary Gaussian approximation breaks down. In practice, when a few leading directions have negative curvature, we exclude them from the stationary instantiation and attribute their effect to the non-stationary drift analysis.

\subsection{Tracking Stationary and Transient Regimes}
When logging full-space stationary bias proxies, we estimate the damped Newton direction by conjugate gradient, solving $(\hat{\mathbf{H}}_\epsilon({\bm{w}})+\lambda_{\mathrm{damp}}\mathbf{I})\bm{m}=-\nabla \hat{\mathcal{L}}_\epsilon({\bm{w}})$ with $\lambda_{\mathrm{damp}}=10^{-3}$, a maximum of $200$ CG iterations, and stopping tolerance $10^{-6}$.
To instantiate the non-stationary transient dynamics, we treat each learning-rate change (epochs where the logged step size $\eta$ differs from the previous epoch) as the onset of a transient.
Starting from the stationary solution at the pre-change epoch, we iterate the per-mode $2\times2$ recursions for the number of SGD steps per epoch (computed from dataset size and batch size) and apply the resulting non-stationary overrides epoch-wise until they converge back to the stationary formulas.
Convergence is detected via the symmetric relative error between stationary and iterated per-mode quantities, using a threshold $\tau=0.05$ for $P=3$ consecutive epochs, after which we revert to the stationary estimates until the next learning-rate change.
To account for cross-epoch drift of the empirically estimated top-$k$ eigenspace, we align consecutive eigenbases by orthogonal Procrustes and transport diagonal mode quantities by the mixing map $x\leftarrow \mathrm{diag}(\mathbf{R}\mathrm{diag}(x)\mathbf{R}^\top)=(\mathbf{R}\odot\mathbf{R})x$ before applying each epoch-level transient update.

\subsection{Runtime and Scalability}
\label{app:runtime_scalability}

The spectral-estimation protocol is an offline checkpoint diagnostic that avoids materializing the full Hessian and the full noise covariance. For $T_{\rm eval}$ evaluated checkpoints, $k$ retained modes, $n_{\rm PI}$ power iterations, and $M$ sampled mini-batches, the leading HVP cost is $O(T_{\rm eval} \cdot kM  \cdot n_{\rm PI})$ backward-equivalent operations, plus $O(T_{\rm eval} \cdot M)$ mini-batch gradient evaluations for projected noise covariance.
The WRN-34-10 study (Figure~\ref{fig:extended_grid_wrn34_lr_schedules} in Appendix~\ref{app:generalizability}) uses the same protocol, but with a smaller retained eigenspace. Scaling to ViT-scale models would require memory-efficient HVPs, activation checkpointing, or sketched leading-mode estimators, but the diagnostic remains checkpoint-based rather than a training-time intervention.

\section{Theoretical Justification for Top-\texorpdfstring{$k$}{k} Proxy}
\label{app:topk_proxy}

This section provides a theoretical justification for the top-$k$ estimation used in our main experiments.
In Section \ref{sec:experiments}, note that we report top-$k$ proxies for the spectral terms (e.g., $-\sum_i \ln \sigma_i^2$) that appear in our PAC-Bayes decomposition.
These proxies are not intended to approximate the full $m$-dimensional quantities in absolute scale; rather, we use them to track the \emph{temporal trends} during training.
The following proposition formalizes this ``dominant-mode proxy'' viewpoint: the epoch-to-epoch \emph{change} of the full entropy term is well-approximated by a top-$k$ truncation whenever the tail modes are approximately stationary (as in a spiked-spectrum regime).

\begin{proposition}[Dominant-mode control of temporal trends]
\label{prop:dominant_mode_proxy}
	
Let $\{\sigma_i^2(t)\}_{i=1}^m$ denote the eigenvalues of the posterior covariance at epoch $t$, and define the full and truncated entropy contributions
	\[
	E(t) := -\sum_{i=1}^m \ln \sigma_i^2(t),
	\qquad
	E_k(t) := -\sum_{i=1}^k \ln \sigma_i^2(t).
	\]
	Then for any two epochs $t,t'$,
	\begin{align}
	\label{eq:topk_entropy_trend_error_identity}
	\Big|(E(t')-E(t)) - (E_k(t')-E_k(t))\Big|
	\;\le\;
	\sum_{i=k+1}^m \left|\ln\frac{\sigma_i^2(t')}{\sigma_i^2(t)}\right|.
	\end{align}
	Moreover, if the tail modes satisfy a small relative-change condition
	$
	\left|\frac{\sigma_i^2(t')}{\sigma_i^2(t)}-1\right|\le \varepsilon_i \le \frac{1}{2}
	$
	for all $i>k$, then
	\begin{align}
	\label{eq:topk_entropy_trend_error_small_change}
	\Big|(E(t')-E(t)) - (E_k(t')-E_k(t))\Big|
	\;\le\;
	2\sum_{i=k+1}^m \varepsilon_i.
	\end{align}
	Under the commuting stationary approximation in \Eqref{eq:stationary posterior covariance}, the per-mode variance satisfies
	\begin{align}
	\label{eq:stationary_mode_variance_app}
	\sigma_i^2(t)
	=
	\frac{\eta(t)}{1-\kappa}\cdot
	\frac{\gamma_i(t)}{\lambda_i(t)\left(2-\frac{\eta(t)}{1+\kappa}\lambda_i(t)\right)}.
	\end{align}
	Thus, the relative change in $\sigma_i^2$ (\Eqref{eq:topk_entropy_trend_error_identity}) is controlled by relative changes in the matched spectral statistics $(\lambda_i,\gamma_i)$.
	\end{proposition}

	\begin{proof}
	We start from the decomposition
	\[
	(E(t')-E(t)) - (E_k(t')-E_k(t))
	=
	-\sum_{i=k+1}^m \left(\ln \sigma_i^2(t')-\ln \sigma_i^2(t)\right)
	=
	-\sum_{i=k+1}^m \ln\frac{\sigma_i^2(t')}{\sigma_i^2(t)}.
	\]
	Applying the triangle inequality yields~\eqref{eq:topk_entropy_trend_error_identity}.
	For~\eqref{eq:topk_entropy_trend_error_small_change}, write $\sigma_i^2(t')/\sigma_i^2(t)=1+r_i$ with $|r_i|\le 1/2$.
	Using the bound $|\ln(1+r)|\le 2|r|$ valid for $|r|\le 1/2$, we obtain
	$
	\left|\ln\frac{\sigma_i^2(t')}{\sigma_i^2(t)}\right|
	\le 2\varepsilon_i
	$
	for any $i>k$, which completes the proof.
	Finally, \Eqref{eq:stationary_mode_variance_app} is the eigendirection-wise specialization of \Eqref{eq:stationary posterior covariance} under commutativity, showing explicitly that $\ln \sigma_i^2$ is an additive combination of $\ln \eta$, $\ln \gamma_i$, and functions of $\lambda_i$.
	\end{proof}

\begin{figure*}[t]
    \centering
        \includegraphics[width=0.95\linewidth]{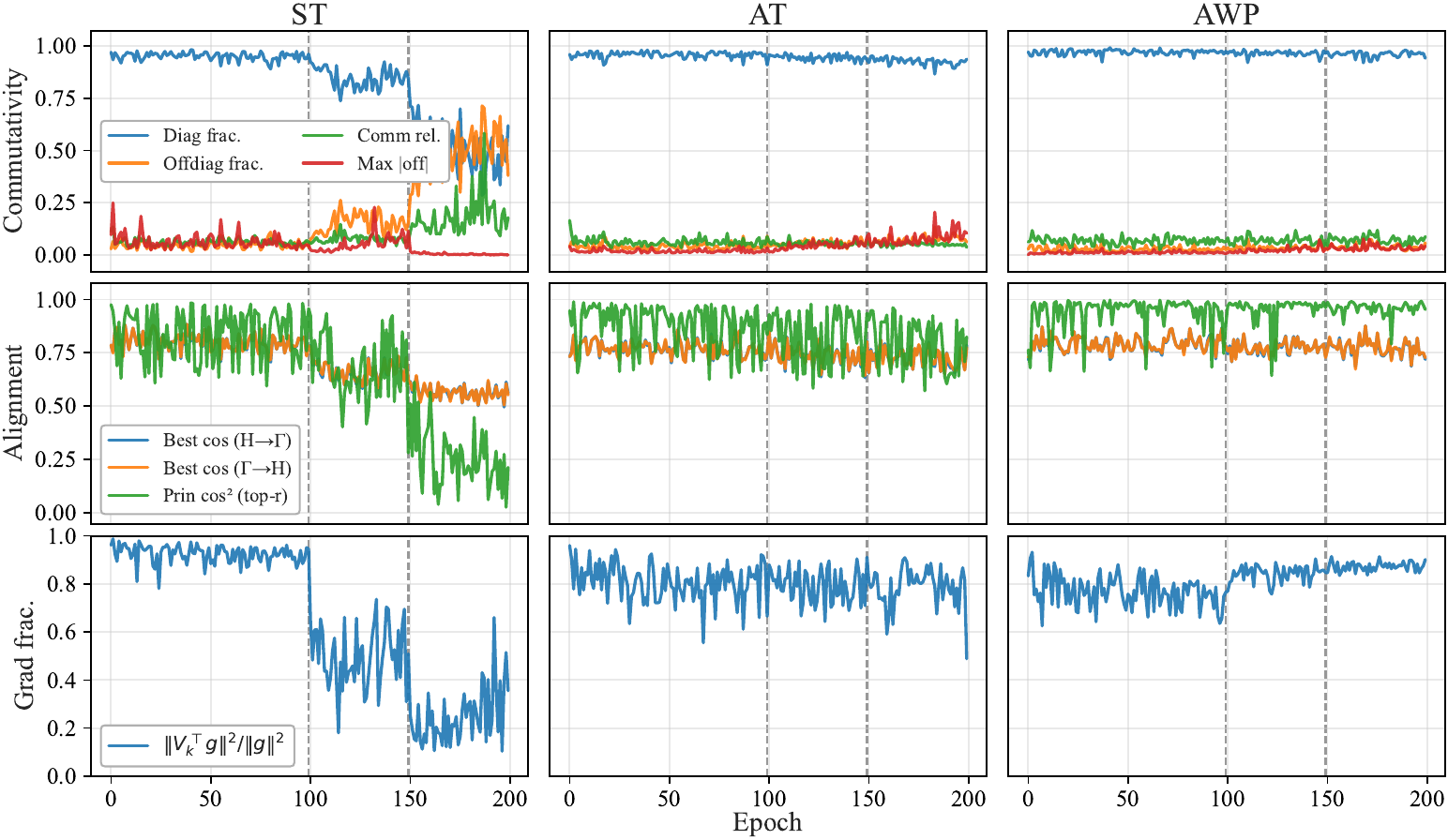}
        \caption{
        Commutativity, alignment, and top-$k$ representativeness diagnostics (CIFAR-10; $k=20$).
        Columns correspond to ST/AT/AWP, and vertical dashed lines indicate learning-rate drops.
        \textbf{Top row:} commutativity/diagonality of the projected covariance $\mathbf{\Gamma}_t$ in the Hessian basis.
        Legend abbreviations: \emph{Diag frac.} $=\|\mathrm{diag}(\mathbf{\Gamma}_t)\|_F^2/\|\mathbf{\Gamma}_t\|_F^2$,
        \emph{Offdiag frac.} $=1-\text{Diag frac.}$,
        \emph{Comm rel.} $=\|[\mathbf{\Lambda}_t,\mathbf{\Gamma}_t]\|_F/(\|\mathbf{\Lambda}_t\|_F\|\mathbf{\Gamma}_t\|_F)$, and
        \emph{Max $|$off$|$} $=\max_{i\neq j}|(\mathbf{\Gamma}_t)_{ij}|$, with $\mathbf{\Lambda}_t=\mathrm{diag}(\lambda_1(t),\dots,\lambda_k(t))$.
        \textbf{Middle row:} alignment between the dominant eigenspaces of $\mathbf{\Gamma}_t$ and the Hessian eigenbasis (cosine similarities and a principal-angle proxy on the top-$r$ subspace with $r=3$).
        \textbf{Bottom row:} gradient-energy fraction in the Hessian top-$k$ subspace, $\|\mathbf{V}_t^\top \bm{g}(t)\|^2/\|\bm{g}(t)\|^2$ (\emph{Grad frac.}).
        Overall, AT and AWP maintain strong commutativity and alignment throughout training and retain high gradient-energy ratios (top-$k$ remains representative for gradient-coupled quantities).
        ST is less stable, and its commutativity/alignment diagnostics can degrade in late training, alongside reduced gradient-energy ratios.
    }
\label{fig:commutativity_alignment}
\end{figure*}

\section{Additional Experiments}
\label{append:additional_experiments}

\subsection{Commutativity and Alignment Assumptions}
\label{app:comm_align}

Our theoretical instantiations rely on a \emph{commuting/diagonal} approximation, which treats the projected gradient-noise covariance as approximately diagonal in the Hessian eigenbasis.
We empirically probe this assumption---and the gradient-relevance of the Hessian top-$k$ subspace---on CIFAR-10 for ST, AT, and AWP. The results are presented in Figure \ref{fig:commutativity_alignment}.

\shortsection{Measurement protocol}
At each epoch $t$, we evaluate the empirical objective at the checkpoint ${\bm{w}}_t$ (clean loss for ST; robust loss for AT/AWP), estimate the top-$k$ Hessian eigenpairs
$\{(\lambda_i(t),\bm{v}_i(t))\}_{i=1}^k$, and form the matrix $\mathbf{V}_t=[\bm{v}_1(t),\dots,\bm{v}_k(t)]$.
Using mini-batch gradients $\bm{g}_b(t)$, we estimate the \emph{projected} gradient-noise covariance
$
\mathbf{\Gamma}_t := \mathrm{Cov}_b(\mathbf{V}_t^\top \bm{g}_b(t))\in\mathbb{R}^{k\times k},
$
whose diagonal entries are the per-mode variances $\gamma_i(t)$ used in our diagonal instantiation.
We then report:
(a) commutativity/diagonality metrics based on the off-diagonal energy of $\mathbf{\Gamma}_t$ in the Hessian basis and the relative commutator norm
$\|[\mathbf{\Lambda}_t,\mathbf{\Gamma}_t]\|_F / (\|\mathbf{\Lambda}_t\|_F\|\mathbf{\Gamma}_t\|_F)$ with $\mathbf{\Lambda}_t=\mathrm{diag}(\lambda_1(t),\dots,\lambda_k(t))$;
(b) alignment metrics based on cosine similarities and principal angles between the dominant eigenspaces of $\mathbf{\Gamma}_t$ and the Hessian basis; and
(c) the fraction of gradient energy captured by the Hessian top-$k$ subspace, $\|\mathbf{V}_t^\top \bm{g}(t)\|^2/\|\bm{g}(t)\|^2$ (\emph{Grad frac.}).

If the commutativity assumption breaks, then the two matrices cannot be simultaneously 
diagonalized, and the spectral quantities $\{\lambda_i,\gamma_i\}$ would no longer represent 
matched curvature–noise pairs. As a consequence, the posterior covariance could exhibit uncontrolled 
cross-terms, making the PAC-Bayesian bound less interpretable and potentially much looser. 
Similarly, if alignment were absent, the principal directions of stochastic gradient noise would 
not coincide with those of curvature. This mismatch would spread noise across directions with 
different curvatures, leading to inefficient exploration, less reliable stationary approximations, 
and weaker predictive power of our framework.

\shortsection{Top-$k$ representativeness}
In addition to commutativity/alignment, our use of top-$k$ truncation implicitly assumes that the dominant subspace remains \emph{gradient-relevant}.
We therefore report the gradient-energy fraction captured by the Hessian top-$k$ subspace, $\|\mathbf{V}_t^\top \bm{g}(t)\|^2/\|\bm{g}(t)\|^2$ (\emph{Grad frac.}; bottom row of Figure~\ref{fig:commutativity_alignment}).
AT/AWP retain high ratios throughout training, whereas ST exhibits a more pronounced drop after learning-rate changes, indicating reduced top-$k$ representativeness in late-stage ST.
Taken together, these diagnostics support the use of the diagonal/top-$k$ approximation as an informative model for AT and AWP, while also identifying regimes (notably late-stage ST) where commutativity/alignment and top-$k$ representativeness degrade, and the approximation should be interpreted with caution.

\begin{figure*}[t]
    \centering
    \includegraphics[width=\linewidth]{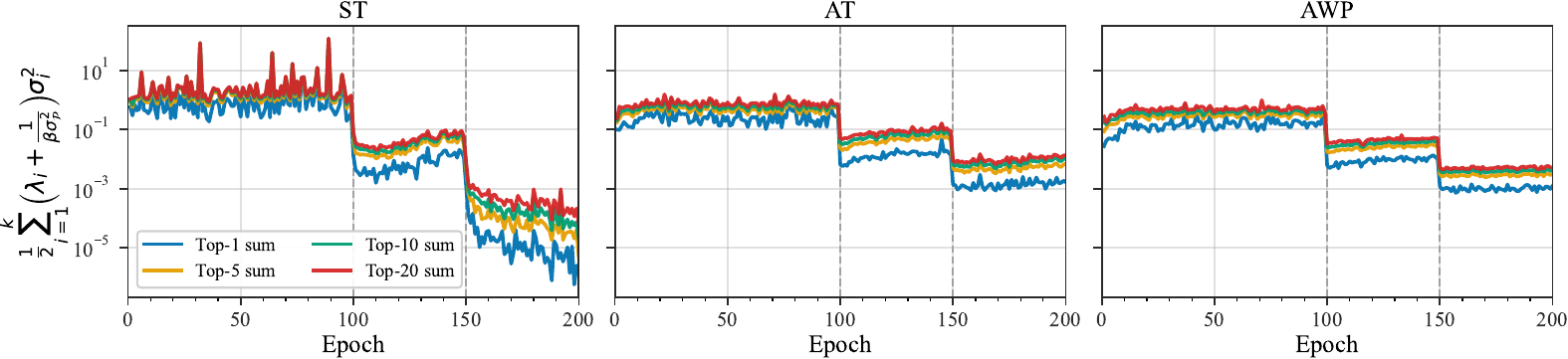}
    \caption{Full-KL shifted variance diagnostics for CIFAR-10 across ST, AT, and AWP ($\ell_\infty$, $\epsilon=8/255$). Curves report top-$k$ partial sums of $\frac12\sum_{i=1}^k(\lambda_{t,i}+1/(\beta\sigma_{\mathcal P}^2))\sigma_{t,i}^2$ for $k\in\{1,5,10,20\}$; dashed lines mark learning-rate drops. This figure verifies that the covariance-trace KL shift included in Theorem~\ref{thm:final robust generalization bound combined} does not change the qualitative temporal diagnostics.}
    \label{fig:full_kl_shifted_variance}
\end{figure*}

\begin{figure*}[t]
    \centering
    \includegraphics[width=\linewidth]{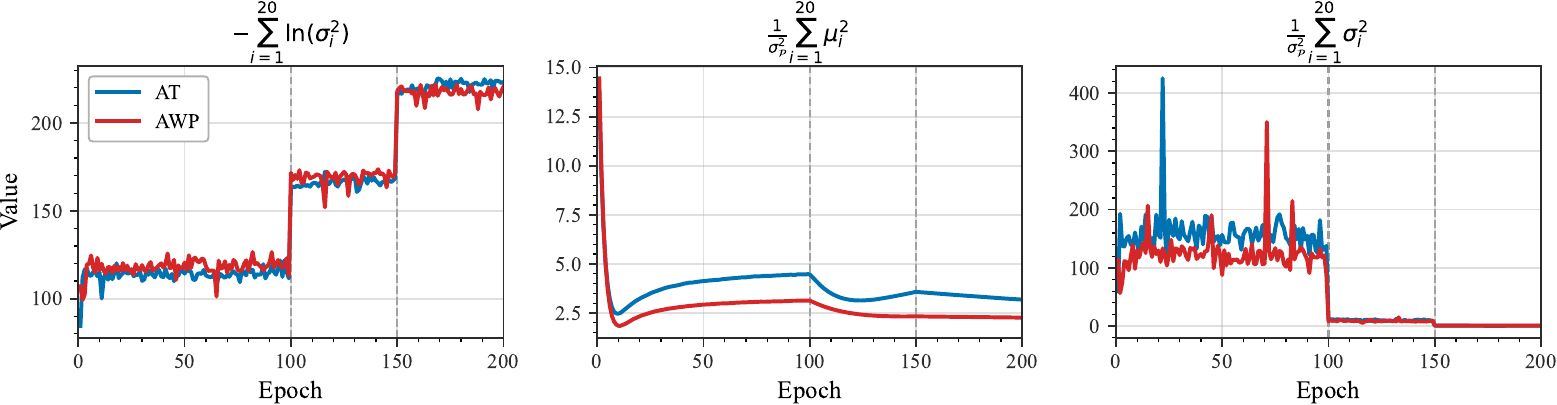}
    \caption{Posterior-dependent KL components for CIFAR-10 across AT and AWP ($\ell_\infty$, $\epsilon=8/255$), using the top-$20$ retained modes. From left to right: log-determinant component, mean-norm component, and covariance-trace component.}
    \label{fig:full_kl_components}
\end{figure*}

\subsection{Full KL and Omitted-Term Diagnostics}
\label{app:full_kl_diagnostics}

The strict bound in Theorem~\ref{thm:final robust generalization bound combined} contains KL terms beyond the log-determinant proxy emphasized in the main diagnostic grid. To ensure the rigor of our analysis and avoid overlooking bound components that may have a critical impact on adversarially robust generalization, we run additional experiments to analyze all KL-related terms and their relative importance. The results are reported in Figure~\ref{fig:full_kl_shifted_variance} and Figure~\ref{fig:full_kl_components}.

In particular, the covariance-trace part of the KL shifts the curvature-weighted variance:
\[
\text{from}\quad \frac12\sum_i\lambda_{t,i}\sigma_{t,i}^2
\quad\text{to}\quad
\frac12\sum_i\left(\lambda_{t,i}+\frac{1}{\beta\sigma_{\mathcal P}^2}\right)\sigma_{t,i}^2.
\]
For CIFAR-10 with $|\mathcal{S}|=50{,}000$, we use $\beta=\sqrt{|\mathcal{S}|}$ and the initialization-matched prior scale $\sigma_{\mathcal P}^2\approx4.93\times10^{-4}$, which gives $1/(\beta\sigma_{\mathcal P}^2)\approx9.07$. Figure \ref{fig:full_kl_shifted_variance} visualizes the diagnostic curves with consideration of this extra KL term. The results suggest that the qualitative temporal trends are not affected by including the KL covariance-trace term, which can be absorbed by the more dominant curvature-weighted variance term in our analytical framework. 

In Figure \ref{fig:full_kl_components}, we further visualize the diagnostic curves of the remaining KL-related terms (mean-norm component and covariance-trace component) for both AT and AWP. After the learning rate drops, both terms become much smaller than the log-determinant KL component, suggesting negligible influence on robust generalization/overfitting.

\subsection{Generalizability Study}
\label{app:generalizability}

To test the robustness of the diagnostic pattern, we repeat the unified diagnostic grids under variations of (i) dataset/learning algorithm, (ii) perturbation radius $\epsilon$, (iii) batch size $B$, and (iv) architecture, objective, dataset, and learning-rate schedule.
All plotted quantities and curve definitions match Figures~\ref{fig:unified_spectra} and~\ref{fig:unified_bound} in the main text:
\emph{loss/error} (top row), leading Hessian eigenvalues $\{\lambda_i\}$ (second row), projected gradient-noise variances $\{\gamma_i\}$ (third row), and the resulting bound decomposition terms (variance/entropy/bias; bottom rows).
Our goal is not only to document how these curves shift across settings, but also to connect the shifts back to the theory-motivated diagnostics: larger curvature is associated with smaller inferred posterior variances, stronger gradient noise is associated with larger $\gamma_i$, and robust-overfitting regimes coincide with learning-rate changes that upset this balance and produce posterior-collapse signatures.

\begin{figure}[t]
    \centering
    \includegraphics[width=\linewidth]{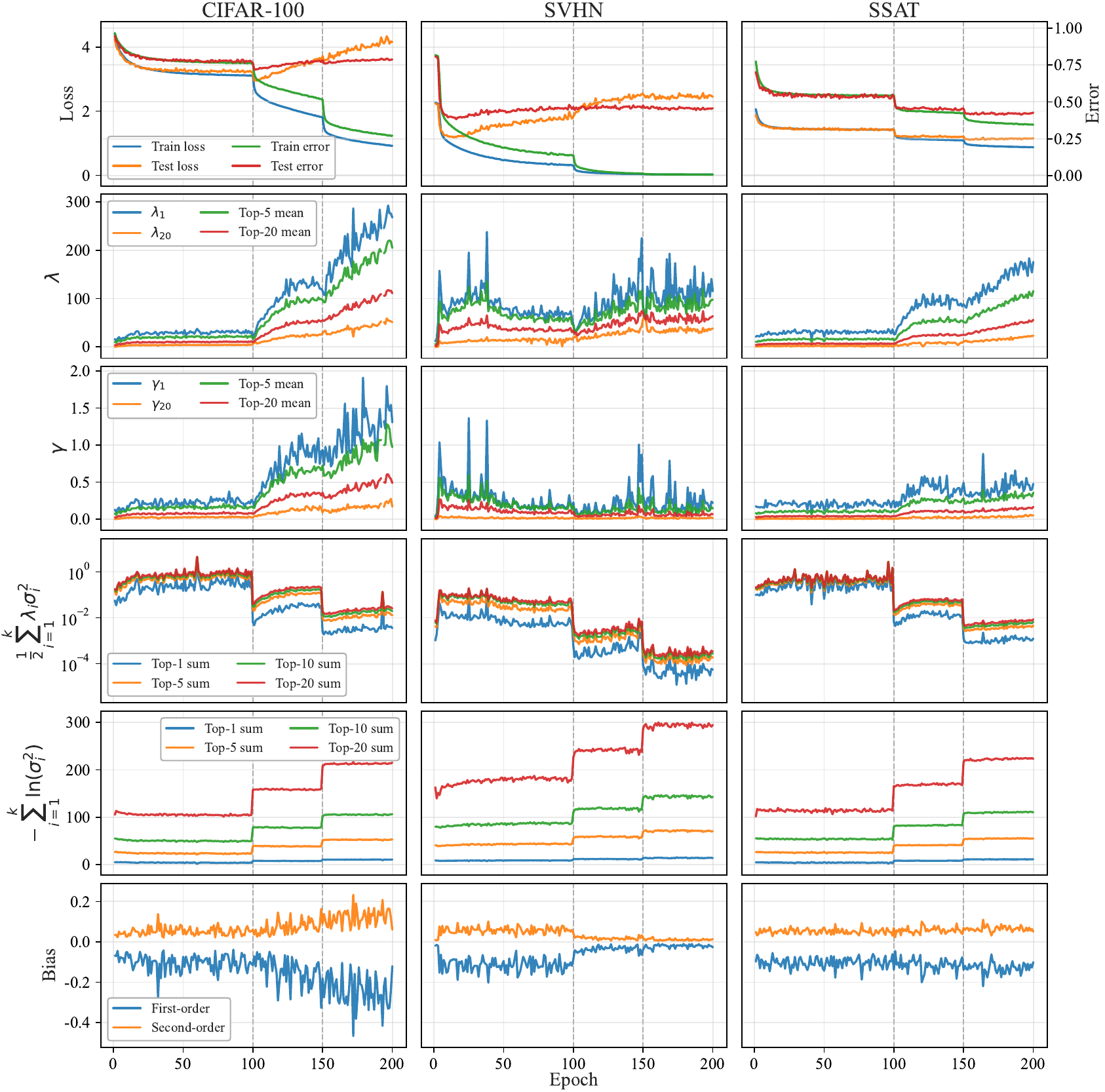}
    \caption{Diagnostic curves across datasets (CIFAR-100, SVHN) and learning algorithms (SSAT).}
    \label{fig:app_cifar100_svhn_semi_grid}
\end{figure}

\shortsection{Datasets and Learning Algorithms}
Figure~\ref{fig:app_cifar100_svhn_semi_grid} extends our analysis to CIFAR-100, SVHN, and semi-supervised adversarial training (SSAT) \citep{carmon2019unlabeled}.
For CIFAR-100 and SVHN, we use the same epoch budget and learning-rate decay \emph{strategy} as CIFAR-10: $200$ epochs with the same piecewise decay points (epochs $100$ and $150$).
We keep the optimizer (momentum SGD with $\kappa=0.9$ and weight decay $5\times10^{-4}$) and batch size ($128$) fixed.
We use an initial learning rate of $0.1$ for CIFAR-100 and $0.01$ for SVHN; other attack hyperparameters follow the CIFAR-10 setting unless stated otherwise (for SVHN, we use a smaller PGD step size $1/255$).

Relative to CIFAR-10, CIFAR-100 exhibits substantially larger leading curvature and projected noise in the dominant modes (larger $\lambda_1$ and $\gamma_1$), with correspondingly larger magnitudes in the PAC-Bayes proxy terms after learning-rate decays.
The post-decay regime shows sharper transients and stronger late-stage degradation in robust test performance, consistent with more pronounced robust overfitting.
In contrast, SVHN displays smaller $\lambda$ and $\gamma$ trajectories and a smoother post-decay evolution, and its bound components evolve more steadily with weaker evidence of abrupt contraction.
SSAT typically lowers risk curves relative to the purely supervised AT baseline, but its geometry does not exhibit the same systematic curvature suppression as AWP; accordingly, its spectra and bound terms fall between standard AT and AWP-like behavior.

These cross-dataset differences are consistent with a margin-distribution view of robust optimization.
Harder, fine-grained classification (CIFAR-100) keeps a larger fraction of training points near the adversarial decision boundary, so the inner maximization continues to surface high-sensitivity examples later into training.
This sustains sharp Gauss--Newton curvature modes in the robust loss and increases the variability of projected gradients, raising both $\lambda_i$ and $\gamma_i$ in the dominant subspace.
In our dynamical PAC-Bayes instantiation, larger curvature is associated with stronger posterior contraction and more pronounced posterior-collapse diagnostics, while larger noise inflates $\sigma_i^2$; after learning-rate decays reduce the step size, this balance can shift toward stronger posterior-contraction signatures.

SVHN is relatively easy for the same architecture: margins are larger, and the adversarial maximization is less likely to produce strongly boundary-adjacent updates, resulting in a smaller dominant spectrum and a more stable curvature--noise balance.
Finally, SSAT introduces additional constraints (via unlabeled examples) that can improve generalization, but it does not explicitly penalize weight-space sharpness as AWP does; hence, while risk curves improve, sharp directions need not be systematically suppressed, and the bound decomposition remains closer to AT than to AWP.

\begin{figure}[t]
    \centering
    \includegraphics[width=\linewidth]{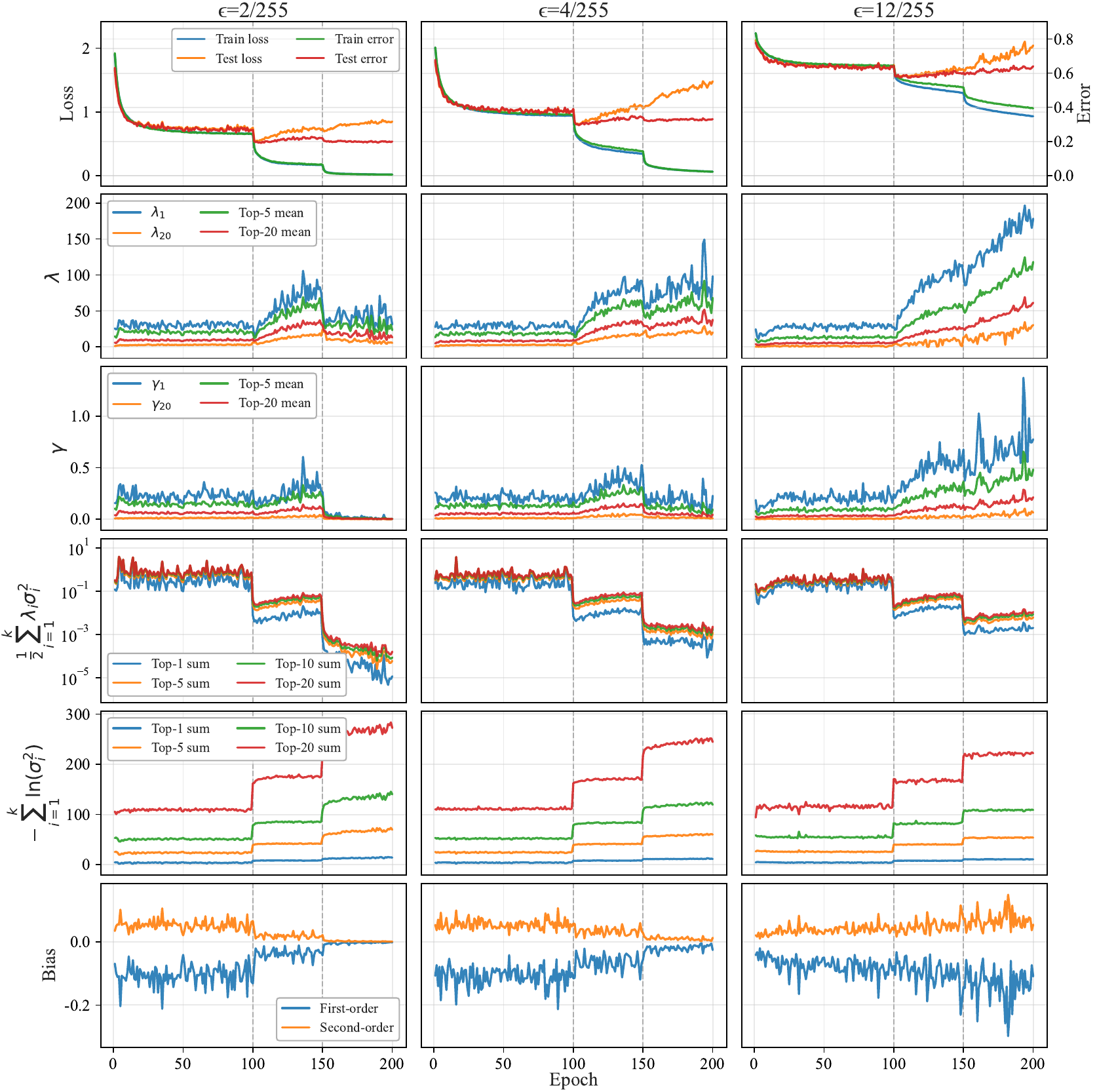}
    \caption{Diagnostic curves with varying perturbation radius $\epsilon\in\{2/255, 4/255, 12/255\}$.}
    \label{fig:app_eps2_eps4_eps12_grid}
\end{figure}

\shortsection{Perturbation Radius}
Perturbation radius $\epsilon$ indicates the strength of the inner maximization and thus affects the geometry of the robust objective.
Figure~\ref{fig:app_eps2_eps4_eps12_grid} reports AT runs with $\epsilon\in\{2/255,4/255,12/255\}$ (the baseline $\epsilon=8/255$ is shown in the AT column of Figures~\ref{fig:unified_spectra} and \ref{fig:unified_bound}).
As $\epsilon$ increases, the dominant curvature modes become systematically sharper (both $\lambda_1$ and the top-$k$ averages increase), and the robust test loss/error worsens.
Projected noise variances also increase, indicating that stronger attacks not only steepen the robust objective but also change the variability of mini-batch gradients in the sharp subspace.
These spectral shifts propagate into the bound decomposition: larger $\epsilon$ yields larger curvature-weighted variance proxies and typically larger entropy penalties, consistent with a tighter inferred posterior and weaker robust generalization.

This trend is expected because a larger $\epsilon$ expands the feasible set of adversarial perturbations and strengthens the inner maximization.
Training updates may then repeatedly accommodate harder, boundary-adjacent perturbations, which sustain sharp curvature modes in the robust loss for longer and amplify their magnitudes.
In the dynamical PAC-Bayes instantiation, posterior contraction associated with curvature (through $\lambda_i$) competes with noise-driven inflation (through $\gamma_i$).
Our ablations suggest that, as $\epsilon$ increases, the geometric tightening induced by sharper curvature is a dominant contributor to late-stage robust generalization degradation, even though the noise level also changes; this supports interpreting robust overfitting through posterior contraction when the curvature--noise balance becomes unfavorable after learning-rate drops.

\begin{figure}[t]
    \centering
    \includegraphics[width=\linewidth]{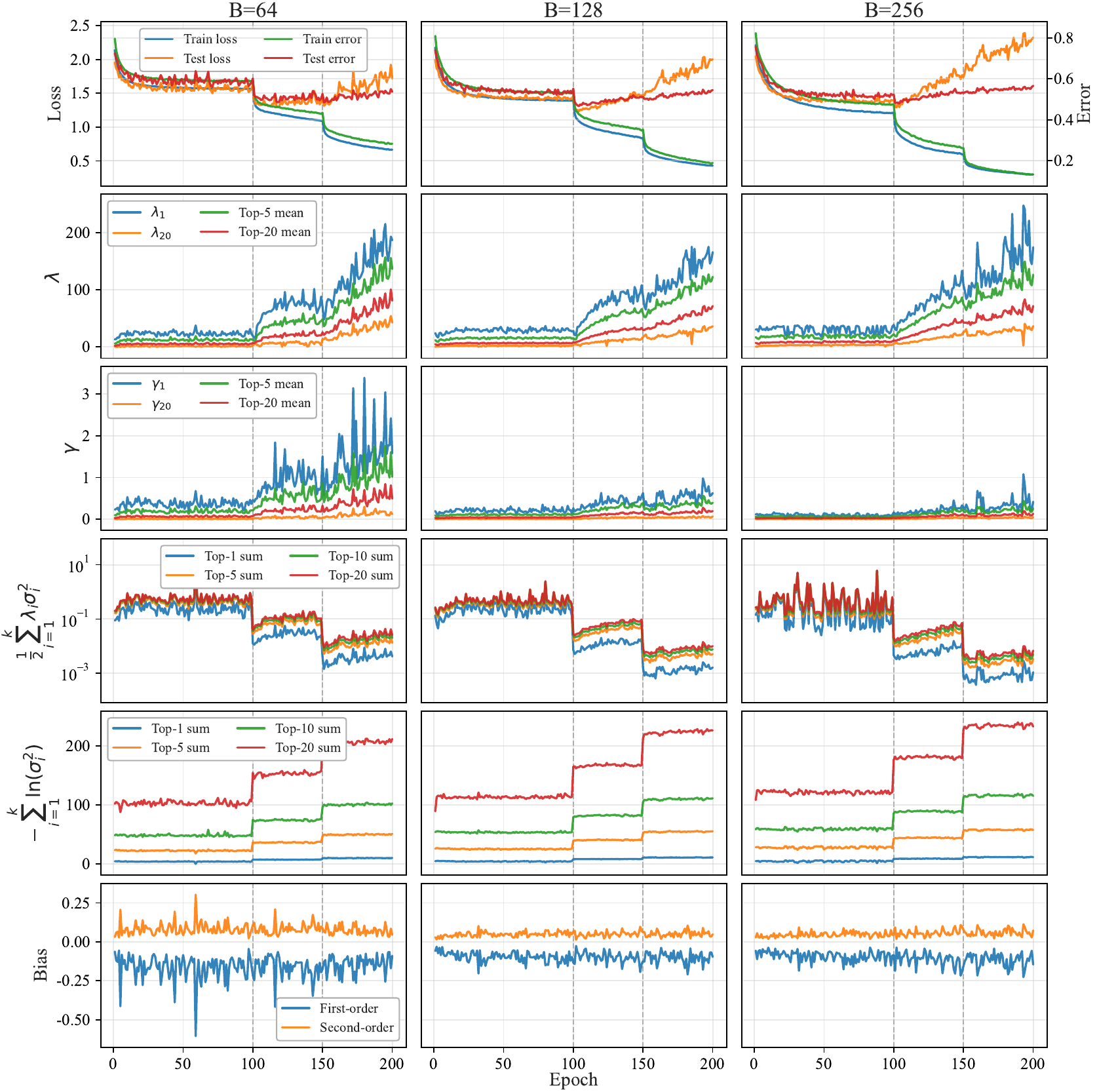}
    \caption{Diagnostic curves with varying batch size $B\in\{64, 128, 256\}$.}
    \label{fig:app_b64_b128_b256_grid}
\end{figure}

\shortsection{Batch Size}
In these experiments, batch size primarily changes the estimated gradient-noise level: larger batches reduce the stochasticity of SGD updates and shrink the projected noise variances.
Figure~\ref{fig:app_b64_b128_b256_grid} compares $B\in\{64,128,256\}$ (note that $B=128$ matches our default AT setting in Figures~\ref{fig:unified_spectra} and \ref{fig:unified_bound}).
Across batch sizes, the curvature trajectories ($\lambda$) remain comparatively similar, while the projected noise variances ($\gamma$) shift by orders of magnitude: smaller batches produce larger $\gamma_i$, and larger batches suppress them.
These shifts are reflected in the bound decomposition.
Larger batches correspond to smaller inferred posterior variances $\sigma_i^2$ and therefore larger entropy proxies $-\sum \ln\sigma_i^2$, and they are accompanied by stronger late-stage degradation in robust test performance.

Under our stationary (and transient-window) approximations, the per-mode posterior variance scales with the projected noise level, $\sigma_i^2 \propto \gamma_i$ (modulated by learning rate, momentum, and curvature).
Increasing $B$ reduces $\gamma_i$, which contracts $\sigma_i^2$ and inflates the log-determinant (entropy) contribution to the KL term in the PAC-Bayes bound.
This yields a dynamical view of a posterior-contraction diagnostic pattern: with insufficient noise injection (large $B$), the effective posterior volume shrinks rapidly, and the bound becomes dominated by KL/entropy penalties in sharp directions, especially after learning-rate decays.
Smaller batches inject more noise, maintain larger posterior variances, and can delay entry into this collapse-dominated regime, thereby mitigating robust overfitting in the same curvature landscape.

\begin{figure}[t]
    \centering
    \includegraphics[width=\linewidth]{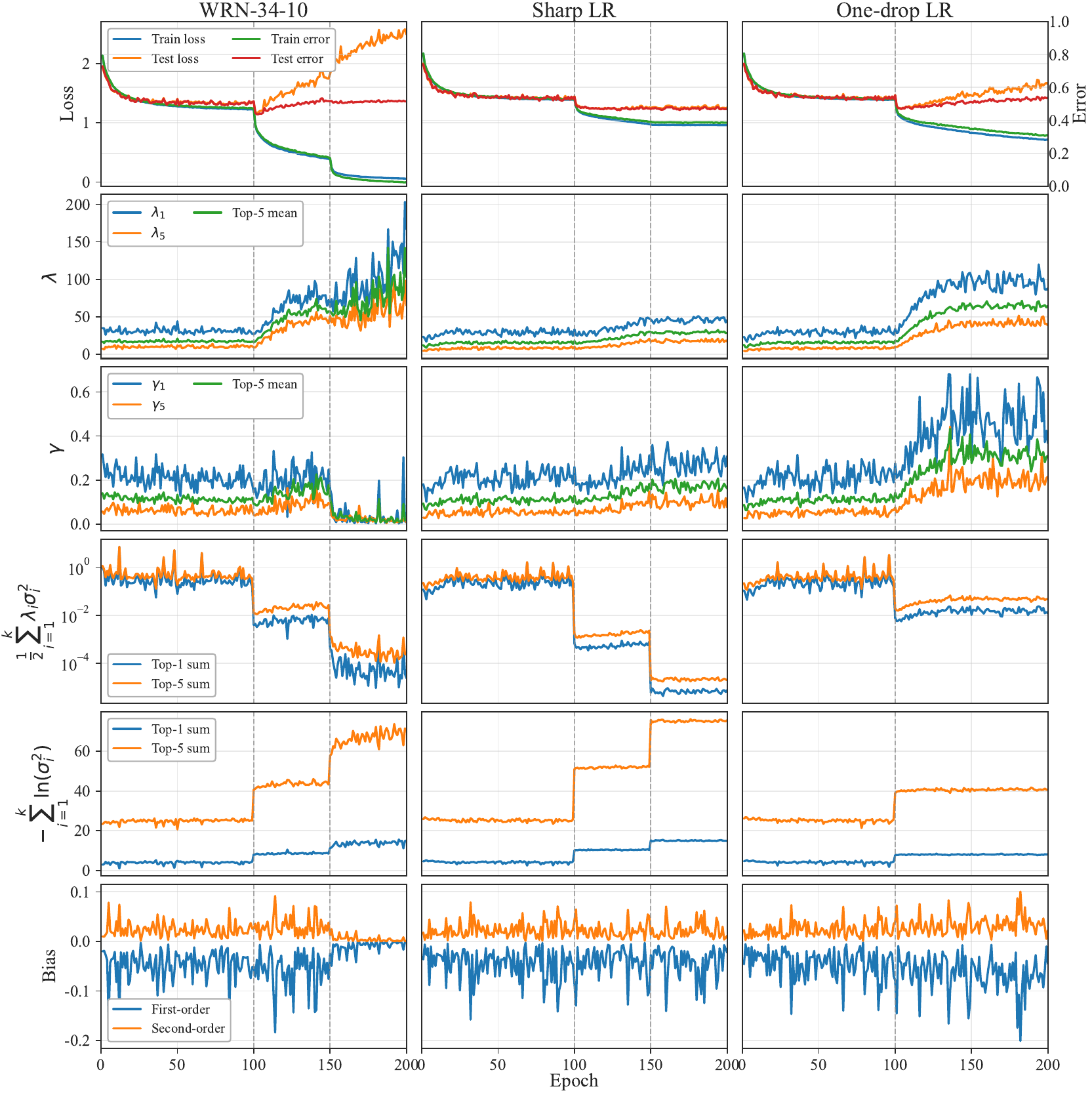}
    \caption{Diagnostic curves on WRN-34-10 and learning-rate schedules under $\ell_\infty$ perturbations with $\epsilon=8/255$. Columns (from left to right) present the results on WRN-34-10, the sharp two-drop LR schedule, and the one-drop LR schedule.}
    \label{fig:extended_grid_wrn34_lr_schedules}
\end{figure}

\begin{figure}[t]
    \centering
    \includegraphics[width=0.8\linewidth]{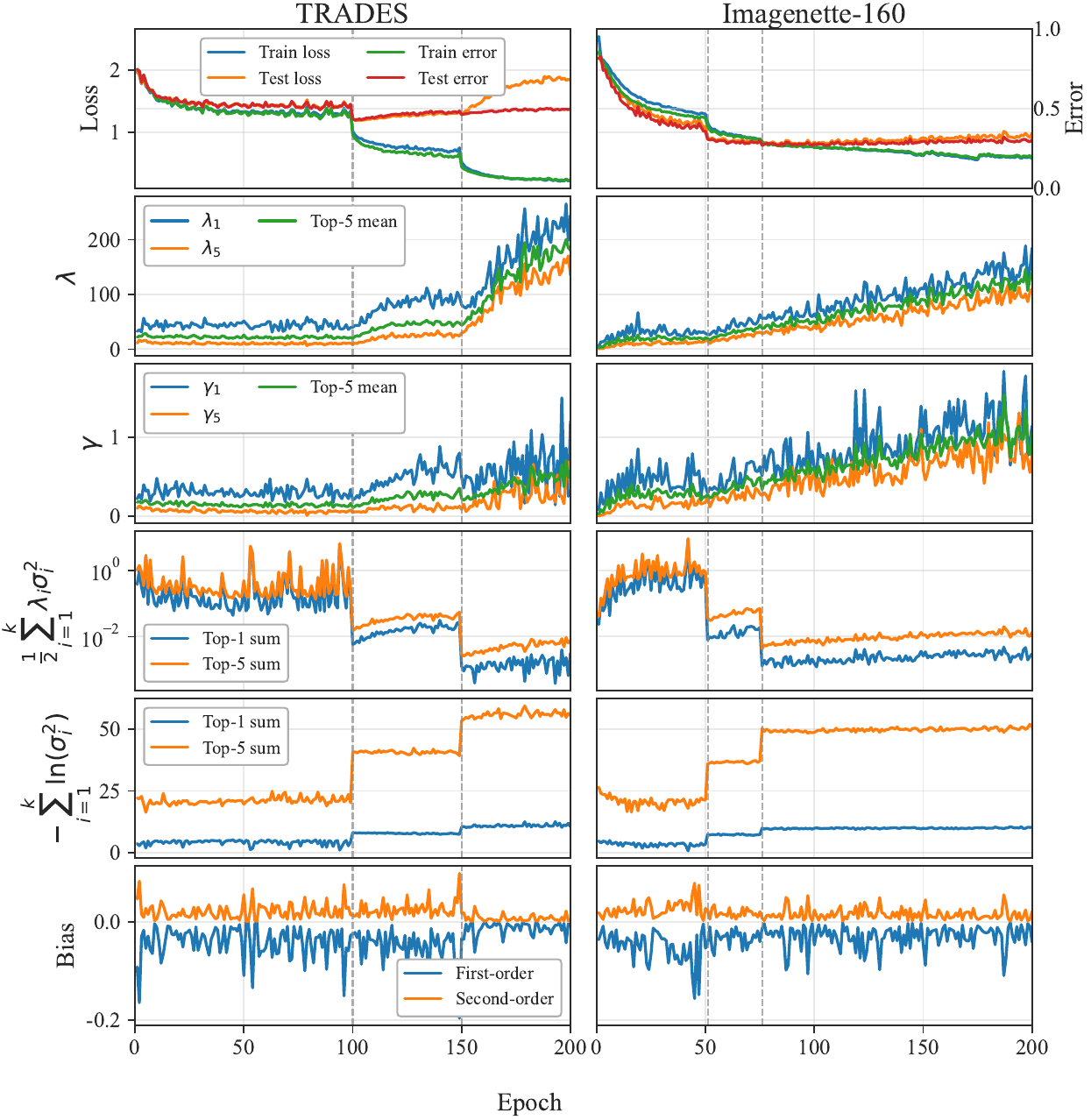}
    \caption{Extended diagnostics for objective-level and dataset-level variations. Columns refer to TRADES on CIFAR-10 under $\ell_\infty$ perturbations with $\epsilon=8/255$ and AT on Imagenette-160 with ResNet-18 under $\ell_\infty$ perturbations with $\epsilon=4/255$.}
    \label{fig:extended_grid_trades_imagenette}
\end{figure}

\shortsection{Other Variations}
Finally, we report extended diagnostics across architecture, learning rate schedules, objectives, and datasets, without altering the paper's central experimental design. Figures~\ref{fig:extended_grid_wrn34_lr_schedules} and~\ref{fig:extended_grid_trades_imagenette} use the same six-row grid format as the generalizability figures above, so each setting is shown with loss/error, curvature, noise, and bound-component diagnostics in one place. We use WRN-34-10~\citep{zagoruyko2016wide} to test a higher-capacity architecture than the default PreActResNet-18, TRADES~\citep{zhang2019theoretically} to test an objective-level change to adversarial training, Imagenette-160~\citep{howard2019imagenette} to test a higher-resolution natural-image setting beyond CIFAR-scale inputs, and schedule variants to isolate learning-rate transitions. Unless otherwise stated, CIFAR-10 settings use $\ell_\infty$ perturbations at $\epsilon=8/255$; Imagenette-160 uses $\ell_\infty$ perturbations at $\epsilon=4/255$. The sharp two-drop schedule uses $0.1$ for epochs $0$--$99$, $0.001$ for epochs $100$--$149$, and $10^{-5}$ afterward. The one-drop schedule uses $0.1$ before epoch $100$ and $0.01$ afterward.

These results show that the diagnostic pattern is not tied to the default PreActResNet-18/CIFAR-10/AT configuration. For WRN-34-10, the robust test loss increases after the learning-rate drop while the leading curvature and log-determinant diagnostics grow, consistent with the posterior-contraction pattern observed in the default AT run. The schedule variants isolate the role of learning-rate transitions: the sharp two-drop schedule produces abrupt changes in the variance and entropy diagnostics at the scheduled drops, whereas the one-drop schedule shows a milder post-drop increase in curvature and projected noise. The TRADES and Imagenette-160 runs preserve the same qualitative organization of the diagnostics: increases in leading curvature and projected noise are accompanied by larger curvature-weighted variance and log-determinant terms. These results support using the proposed diagnostic as a stress test across architecture, objective, dataset, and schedule choices, without changing the main claims or treating the top-$k$ proxies as calibrated full-dimensional bounds.

\end{document}